\documentclass{article}
\usepackage{arxiv}
\usepackage[T1]{fontenc}    
\usepackage{hyperref}
\usepackage{url}
\usepackage{booktabs}
\usepackage{nicefrac}
\usepackage{microtype}
\usepackage{graphicx}
\usepackage{doi}
\usepackage{amsfonts}
\usepackage{amsmath}
\usepackage{amsthm}
\usepackage{amssymb}
\usepackage{mathrsfs} 
\usepackage{algorithm} 
\usepackage{mathtools}
\usepackage{color} 
\usepackage{algpseudocode} 
\usepackage{derivative}
\usepackage{tikz-cd}
\usepackage{indentfirst}
\newcommand{\defeq}{\mathrel{\mathop:}=}
\newcommand\restr[2]{\ensuremath{\left.#1\right|_{#2}}}
\usepackage{algorithm,algcompatible,amsmath}
\algnewcommand\INPUT{\item[\textbf{Input:}]}%
\algnewcommand\OUTPUT{\item[\textbf{Output:}]}%
\algnewcommand\INITIALIZE{\item[\textbf{Initialize:}]}
\algnewcommand\PRETRAIN{\item[\textbf{Pretrain:}]}
\usepackage{xspace}

\title{Safe Reinforcement Learning in Tensor Reproducing Kernel Hilbert Space}

\author{ 
    {\hspace{1mm}Xiaoyuan Cheng}\\
	Department of Civil, Environmental\\
    \& Geomatic Engineering\\
	University College London\\
	London, WC1E 6BT, UK\\
	\And
	{\hspace{1mm}Boli Chen} \\
	Department of Electronic\\
    \& Electrical Engineering\\
	University College London\\
	London, WC1E 7JE, UK\\
    \And
    {\hspace{1mm}Liz Varga}\\
	Department of Civil, Environmental\\
    \& Geomatic Engineering\\
	University College London\\
	London, WC1E 6BT, UK\\
    \And
    \href{https://orcid.org/0000-0002-7480-4250} {\includegraphics[scale=0.06]{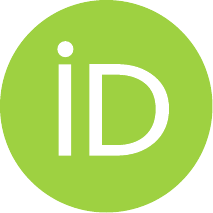}\hspace{1mm}Yukun Hu}\thanks{Corresponding author's Email: yukun.hu@ucl.ac.uk} \\
	Department of Civil, Environmental\\
    \& Geomatic Engineering\\
	University College London\\
	London, WC1E 6BT, UK\\
}

\begin{document}
\maketitle

\begin{abstract}
This paper delves into the problem of safe reinforcement learning (RL) in a partially observable environment with the aim of achieving safe-reachability objectives. In traditional partially observable Markov decision processes (POMDP), ensuring safety typically involves estimating the belief in latent states. However, accurately estimating an optimal Bayesian filter in POMDP to infer latent states from observations in a continuous state space poses a significant challenge, largely due to the intractable likelihood. To tackle this issue, we propose a stochastic model-based approach that guarantees RL safety almost surely in the face of unknown system dynamics and partial observation environments. We leveraged the Predictive State Representation (PSR) and Reproducing Kernel Hilbert Space (RKHS) to represent future multi-step observations analytically, and the results in this context are provable. Furthermore, we derived essential operators from the kernel Bayes' rule, enabling the recursive estimation of future observations using various operators. Under the assumption of \textit{undercompleness}, a polynomial sample complexity is established for the RL algorithm for the infinite size of observation and action spaces, ensuring an $\epsilon-$suboptimal safe policy guarantee.
\end{abstract}


\section{Introduction}
Reinforcement learning (RL) is a learning framework that handles sequential decision-making problems. Decision-makers are so-called RL agents who aim to learn an optimal policy to execute consecutive controls by achieving long-term rewards in interacting with the (unknown) environment. Although the RL has proved extensively successful in wide applications, it is still conservative on applications due to common concerns of modern RL algorithms, such as sample complexity, robustness, stability, and scalability \cite{sutton2018reinforcement}. Thus, traditional RL algorithms make it hard to guarantee a safe and adaptive policy with a provable result \cite{garcia2015comprehensive}. The challenge lies in learning a optimal control action selection strategy without known dynamics to sufficiently infer the true state of the system \cite{carr2022safe}. Many probabilistic methods estimate the optimal action by a belief latent state, but the probability space is intractable when it is infinitely large, and interpolating cannot sample all points in the target space.

In recent years, safe learning has addressed the above challenge by enforcing safety constraints when framing the learning task. We categorize and analyze two approaches to safe learning. The first is based on modifying the optimality criterion in the discounted finite/infinite horizon with a safety factor \cite{wachi2020safe, wen2018constrained, brunke2022safe, qin2021density}. The second is based on the modification of the exploration process through the incorporation of external knowledge or the guidance of a risk metric, such as teacher guidance \cite{torrey2012help}, adversarial training \cite{liang2022efficient}, Lyapunov function \cite{choi2020reinforcement} or control barrier functions \cite{taylor2020learning, tan2021high}. In the review of state-of-art safe learning research, high-level (hard and probabilistic) safety guarantee has been mostly proved in linear or affine-control dynamics. However, assumptions of the two former methods are almost based on the Markov Decision Process (MDP) \cite{laroche2019safe, kidambi2020morel}, and the latter methods are usually defined on linear equations \cite{liu2022robot, boffi2021learning}. It is still insufficient to design expressive models to increase the generalization of safe learning in nonlinear, continuous, and high-dimensional systems or stochastic dynamics \cite{levine2020offline}. It draws attention to developing an appropriate framework to characterize dynamical systems, quantify system uncertainties, and connect to RL optimization frameworks.

To characterize dynamical systems, our method can achieve sample complexity, robustness, stability, and scalability merely by using observations in a smart way and can give a generalized expression with analytical form through operator algebra theory in Reproducing Kernel Hilbert Space (RKHS). The method starts with Predictive State Representation (PSR), a more generic approach for modelling control system dynamics beyond many sequential models such as MDP, Partially Observable Markov Decision Process (POMDP), and Hidden Markov Model (HMM) \cite{thon2015links}, which can scarcely be reached in dynamical control systems. Furthermore, introducing the RKHS to the PSR framework can do mean embedding of the future states without inferring the original probability space. Even with the non-linear and high-dimension nature of the dynamics, the RKHS can still transfer it into an invariant feature space with an infinite basis.  Thus, we would like to answer the following question:

\begin{center}
    \textit{Can we propose a provably efficient safe RL framework in a partially observable environment?} 
\end{center}
\begin{center}
    \textit{The answer is 'yes', by combining the advantage of PSRs and Reproducing Kernel Hilbert Spaces.}
\end{center}

\textbf{Related work.} PSR was first proposed by Littman and Sutton \cite{littman2001predictive}, which is an extension of conventional sequential models. It has been illustrated that multi-step action-conditional predictions of future observations can represent states of a dynamical system. The state representations of a dynamical system grounded in data in this way may be easier to learn and less dependent on accurate prior models with better generalization than POMDP since it does not rely on its latent state. In addition, PSR can be linked with Baye's filter since the future states can be represented as $\mathbb{P}(o \mid a, h) = \frac{\mathbb{P}(o,a\mid h)}{\mathbb{P}(a \mid h)}$, where $o, a, h$ denotes the future observations, actions, and histories, respectively. The introduction of the kernel representation (Appendix \ref{Appendix: Basics of RKHS}) can improve the generalization of PSR in a stochastic dynamical system. \cite{fukumizu2011kernel} proposed the kernel Bayes' rule, which can derive Bayesian computation without a likelihood and filtering in a non-parametric state-space model. Based on this work, \cite{boots2013hilbert} expressed the controlled PSR with a canonical mean embedding in RKHS. \cite{thon2015links} compared various sequential decision-making models and emphasized the generalization of PSR in stochastic sequential problems. From a control theory perspective, the forward operator in PSR can have a natural connection with Model Predictive Control (MPC) since the mean embedding in RKHS can be regarded as an unbiased estimation of future states. To the best of the author's knowledge, no similar research focuses on this point.

In recent years, researchers in the learning theory field have started to pay attention to connecting RL with PSRs. For example, \cite{muandet2017kernel,lagoudakis2003least} indicated the potential application of RKHS on MDP, where a fixed point of Bellman optimality can be quickly obtained without a large data sample. \cite{zhan2022pac} developed a Probably-Approximately-Correct-RL for PSR to achieve a near-optimal policy in sample complexity scaling polynomially concerning all relevant parameters of systems. The PSR allows the compression of the sequential model by minimal core test, which can be more generalized than $m-$step revealing \cite{liu2022partially} and $m-$step decodable tabular POMDPs \cite{efroni2022provable}. However, research on constrained stochastic dynamic systems by PSRs is still lacking. To this end, we proposed a bilinear form to represent the dynamics of controlled stochastic systems in RKHS, where the simple regression was used in the actor-critic RL framework to constrain behaviours that satisfy criteria to achieve a goal. 

To quantify systems uncertainties, we introduce Kernel Mean Embedding into the safe learning framework for the first time by PSRs, a non-parametric method that enriches the expressiveness of traditional regression. The contributions of this study include the following:
\begin{itemize}
    \item proposed a generalized framework to represent the safe dynamical system without any assumption of Markovian property;
    \item represented the safe learning by kernel PSR with a bilinear form in tensor RKHS;
    \item  proposed some important operators in RKHS to estimate value/risk function with a polynomial sample complexity;
    \item provided proof of safe PSR kernel presentation RL with provably efficient properties.
\end{itemize}

\section{Preliminaries}
This section introduces the background of stochastic observable dynamical systems and the corresponding representation of Predictive State Representation (PSR) in such systems. Subsequently, the main properties of RKHS, kernels, and the connection with safe RL tasks are explained.

\subsection{Notion}
For any $n \in \mathbb{N}$, let $[n] = \{1, \cdots n \}$. $\lVert \cdot \rVert_p$ denotes the $l^p-$norm of a function and $\langle \cdot, \cdot \rangle$ the inner product. For any operator $R$,  $\lVert R \lVert$ denotes the norm of its matrix. For any discrete or continuous set $\mathcal{X}$, $L^p(\mathcal{X})$ is the space of \textit{p}-integrable functions over $\mathcal{X}$, and $\Delta(\mathcal{X})$ donates the space of probability density functions over $\mathcal{X}$ when $\mathcal{X}$ is continuous, or the space of probability mass functions when $\mathcal{X}$ is discrete. Also, the notion of $\int_{\mathcal{X}}$ in this paper is general for summation over $\mathcal{X}$ no matter whether $\mathcal{X}$ is discrete or continuous. For a sequence of variables $x_1, x_2, \cdots$, we use $x_{i:j}$ to denote the subsequence $x_i, x_{i+1}, \cdots x_{j}$, for $i \leq j$. Lastly, we use the notation $\lvert \mathcal{X} \rvert$ to denote the cardinality of arbitrary set $\mathcal{X}$, and $\oplus$ and $\otimes$ to denote the direct sum and tensor product defined over rings (see Appendix \ref{Appendix: Some Important Definitions and Properties of PSRs}), respectively.

\subsection{Stochastic observable dynamical and PSR}
\par
A canonical PSR \cite{littman2001predictive} notion is used to represent the dynamical system in this study.
Consider a controllable stochastic dynamical system of a single agent, where the action set $\mathcal{A}$ consists of all possible actions $a_1, a_2, \cdots, a_{\lvert \mathcal{A} \rvert} $, and all possible observation set $\mathcal{O}$ is generated after an action, containing all possible observations $o_1, o_2, \cdots, o_{\lvert \mathcal{O} \rvert} $. The symbol $\mathcal{S}$ denotes the latent state, and latent states cannot be observed in a generalized setting. In this context, we use $h_t$ to denote the \textit{history} before the time step $t$, $h_t$ is the action-observation pairs experienced by the agent, represented as $h_t = (o_1a_1, o_2a_2, \cdots,o_{t-2}a_{t-2},  o_{t-1}) $. $H =\{ h_t \in \mathcal{A}^{n-1} \times \mathcal{O}^{n} \mid \forall t \in \mathbb{N} \} \subset \bigcup_{n \in \mathbb{N}} \mathcal{A}^{n-1} \times \mathcal{O}^n $ is the set of all possible action-observation pairs over the entire horizon. Another important concept is the $test$ denoted as $ \mathcal{U}_{h} \subset \bigcup_{W \in \mathbb{N}} \mathcal{A}^{W} \times \mathcal{O}^W$, which is also a sequence of action-observation pairs with window length $W$ after an arbitrary history $h_t$. As such, an arbitrary $test$ is denoted as 
$t_{h} = a_{t-1 : t+W-2} \oplus o_{t : t+W-1}.$
For the sake of brevity, we use $t_h(a) = a_{t-1 : t+W-2}$ and $t_h(o) = o_{t : t+W-1}$ to denote actions and observations in a future episodic test, respectively. Thus, a prediction of a state is defined as the probability mass of seeing a test’s observation in sequence, given the actions of the tests are taken in sequence from a history, the basic properties of PSR will be given as follows.

\textbf{Test probability.} Let $t_h = (a_{t-1: t+W-2}, o_{t : t+W-1})$ with length $W \in \mathbb{N}$ represent a test. We define the probability of test $t_h$ being successfully conditioned on reachable history $h_t$ as:
\begin{equation}
\begin{split}
    \mathbb{P}(t_h \mid h_t) & \defeq \mathbb{P}(o_{t : t+W-1} \mid h_t, do(a_{t-1: t+W-2})) \\
    &= \prod_{t^{'} = t}^{t+W-1} \mathbb{P}(o_{t^{'}} \mid h_{t^{'}}, a_{t^{'}-1} ) \mathbb{P}(a_{t^{'}-1} \mid h_{t^{'}}) , \label{Equation: test probablity}
\end{split}
\end{equation}
where the $do$ operator means intervening with a sequence of action  $a_{t-1:t+w-2}$. The probability of the sequential observation $o_{t : t+W-1}$ can be revealed after executing actions $a_{t-1: t+W-2}$ conditioning on history $h_t$. Obviously, if the history $h_t$ is not reachable, the probability of $\mathbb{P}(t_h \mid h_t)$ is measure-zero.

According to Baye's rule, the observation conditional probability under action intervention is:
\begin{equation}
    \mathbb{P}(t_h(o) \mid h_t, t_h(a)) = \frac{\mathbb{P}(t_h(o), t_h(a) \mid h_t)}{\mathbb{P}(t_h(a) \mid h_t)} \label{Equation: observation conditional probability}.
\end{equation}
The core idea behind the PSR is that if we know the probability distribution of all possible outcomes of executing all possible tests, then we can obtain optimal expectations by taking a proper action sequence. The idea can have many connections with optimal control problems without any assumptions of the complexity and linearity of a dynamical system, e.g. MPC and stochastic control problems. 

\textbf{Forward dynamics of predictive states.}  One-step dynamics $(o_t, a_{t-1}) \in \mathcal{O} \times \mathcal{A}$ after one arbitrary history $h_t \in H$, the shift test probability $t_{h+1} \defeq (a_{t: t+W-1}, o_{t+1: t+W})$ can be measured as:
\begin{equation}
    \mathbb{P} (t_{h+1} \mid h_{t+1} ) = \frac{\mathbb{P}(t_{h+1}, o_t,  a_{t-1} \mid h_{t})}{\mathbb{P}(o_t,a_{t-1} \mid h_t)}.
\end{equation}
After the one-step shift of dynamics, the new history will be recursively updated as $h_t \oplus (o_t, a_{t-1})$. By the Baye's rule, the filtered  probability $\mathbb{P} (t_{h+1} \mid h_{t+1} )$ can be measured. Similar to Eq. \eqref{Equation: observation conditional probability}, the conditional probability of shifted observation can be: 
\begin{equation}
    \mathbb{P} (t_{h+1}(o) \mid h_{t+1}, t_{h+1}(a)) = \frac{\mathbb{P}(t_{h+1}(o), o_t \mid h_{t}, t_{h+1}(a), a_{t-1} )}{\mathbb{P}(o_t \mid h_t, a_{t-1})} \label{Equation: shifted observation probablity}.
\end{equation}
Here, the core idea of updating future observation probability is analogous to the effect of the Kalman Filter \cite{sun2023adaptive} or the Bayesian filter in POMDP \cite{song2016kernel}. For simplicity, we denote $(t_{h+1}(o), o_t)$ and $( t_{h+1}(a), a_{t-1})$ as $\tilde{t}_{h+1}(o)$ and $\tilde{t}_{h+1}(a)$, repsectively. 

\subsection{Safe Reinforcement Learning Under Probabilistic PSR} \label{Subsection: Safe Reinforcement Learning Under Probabilistic PSR}
It is generally believed that an agent's trajectory is considered safe if and only if all states in the trajectory are in the target safe states \cite{fisac2019bridging}. Instead of adopting belief-based POMDP \cite{araya2010pomdp, fischer2020information}, our framework utilizes finite horizon PSRs as the fundamental control framework for model-based RL tasks. The objective of RL is to maximize the accumulated reward by determining the optimal sequence of actions $ \mathcal{A}^{W}$, and the policy is defined as $\pi: [H] \rightarrow \Delta(\mathcal{A}^W) \subset L^1(\mathcal{A}^W)$. First, the agent takes one step action. Then, the corresponding one-step observation is obtained. In the conventional POMDP problem, the belief state of latent states should be estimated, and the unsafe belief state of latent states should be restricted to a sufficiently small value. In this context, there exists an observable operator $\mathbb{O}: [H] \times L^1(\mathcal{S}) \rightarrow L^1(\mathcal{O})$ as
\begin{equation}
     \mathbb{O} \circ f( \mathcal{S}_t \mid h_t) = \int_{s_t \in \mathcal{S}_t} \mathbb{P}(o_t \mid s_t, h_t) d f(s_t \mid h_t)  \label{Equation: Observable operators} 
\end{equation}
where $ f( \mathcal{S}_t \mid h_t)$ measures the belief state, the observable operator can infer the observation probability. The observable operator can use PSR to partially observable RL problems, and safety can be guaranteed after introducing risk functions. Following the definition of PSR, we consider a finite-horizon safe problem as
\begin{equation}
\begin{split}
    &  \max_{\pi_t} \quad \mathbb{E} [\sum_{t^{'}=t}^{t+W} r(s_{t^{'}}) \mid h_t, a_{t-1: t+ W -1} \sim \pi(h_t)], \qquad \forall h_t \in [H]
    \\ & s.t. \quad \mathbb{E}[\sum_{t^{'}=t}^{t+W} c_i(s_{t^{'}}) \mid h_t, a_{t-1: t+ W -1} \sim \pi(h_t)] \leq \Bar{C}_i, \qquad i \in \{1, \cdots, N \}
    \label{RL objective}
\end{split}
\end{equation}
where $r$ is the reward function defined on the latent state $\mathcal{S}$, and  $c_i$ is the risk function defined as $c_i: \mathcal{S} \rightarrow \mathbb{R}$. In this setting, future observations under the PSR setting can be estimated by Eq. \eqref{Equation: observation conditional probability}, where 
\begin{align*}
    \mathbb{P} ( \tilde{t}_h(o) \mid h_t, \pi(h_t)) = \frac{\mathbb{P}(\tilde{t}_h \mid h_t)}{\pi(h_t)} 
\end{align*}
According to the properties of the observable operator in the incompleteness setting (definition of the incompleteness setting in Appendix \ref{Appendix: Some Important Definitions and Properties of PSRs}), the belief state can be measured by the inverse of observable operator $\mathbb{O}^{\dag}: [H] \times L^1(\mathcal{O}) \rightarrow L^1(\mathcal{S})$. The inverse of the observable operator $\mathbb{O}^{\dag}$ is essential in representing a partially observable environment since changes from latent state $\mathcal{S}$ to observation space $\mathcal{O}$ measurements are more common in the real world. The random matrix $\mathbb{O}^{\dag}$ will converge to a unique barycenter in the convex probability simplex \cite{berg1984harmonic}, and then we can obtain the result as:
\begin{equation}
\begin{split}
    \mathbb{E}_{{\mathbb{O}^{\dag} \sim \mu(\mathbb{O}^{\dag})}}[\int_{o_t \in \mathcal{O}} ( r \circ \mathbb{O}^{\dag}) (o_t) d \mathbb{P}(o_t \mid h_t, a_{t-1} )]  = \int_{s_t \in \mathcal{S}} r(s_t) f(s_t \mid h_t, a_{t-1}) ds_t
    \label{Equation: pull-back value}
\end{split}
\end{equation}
and 
\begin{equation}
\begin{split}
    \mathbb{E}_{{\mathbb{O}^{\dag} \sim \mu(\mathbb{O}^{\dag})}}[\int_{o_t \in \mathcal{O}} ( c \circ \mathbb{O}^{\dag}) (o_t) d \mathbb{P}(o_t \mid h_t, a_{t-1} )]  = \int_{s_t \in \mathcal{S}} c(s_t) f(s_t \mid h_t, a_{t-1}) ds_t
    \label{Equation: pull-back risk}
\end{split}
\end{equation}
where $\mu(\mathbb{O}^{\dag})$ is the distribution of random matrix $\mathbb{O}^{\dag}$. The push-forward operator $\mathbb{O}$ and pull-back operator $\mathbb{O}^{\dag}$ preserve the invariant of expected reward, where we avoid bad states \cite{ni2023learning}. In this situation, PSR allows estimating the future finite-horizon accumulated reward/risk without inferring the latent states according to Eq. \eqref{Equation: pull-back value} and \eqref{Equation: pull-back risk}, such that 
\begin{equation}
\begin{split}
    V^\pi(h_t) & \defeq \mathbb{E} [\sum_{t^{'}=t}^{t+W} r(s_{t^{'}}) \mid h_t, a_{t-1: t+ W -1} \sim \pi(h_t)]  \\
    & = \mathbb{E}[\sum_{t^{'}=t}^{t+W} (r \circ \mathbb{O}^{\dag}) (o_{t^{'}}) \mid h_t, a_{t-1: t+ W -1} \sim \pi(h_t)] \\
    & = \int_{\tilde{t}_{h}(a) \sim \pi(h_t)} (r \circ \mathbb{O}^{\dag}) (\tilde{t}_h(o)) \mathbb{P} ( \tilde{t}_h(o) \mid h_t, \tilde{t}_h(a)) d \mathbb{P}( \tilde{t}_{h}(a) \mid h_t) \label{Equation: integral form value function}
\end{split}
\end{equation}
and the corresponding risk function $c$ is 
\begin{equation}
\begin{split}
    C_{i}^\pi(h_t) &  \defeq \mathbb{E} [\sum_{t^{'}=t}^{t+W} c_{i}(s_{t^{'}}) \mid h_t, a_{t-1: t+ W -1} \sim \pi(h_t)]  \\
    & = \int_{\tilde{t}_{h}(a) \sim \pi(h_t)} (c_{i} \circ \mathbb{O}^{\dag}) (\tilde{t}_h(o)) \mathbb{P} ( \tilde{t}_h(o) \mid h_t, \tilde{t}_h(a)) d \mathbb{P}( \tilde{t}_{h}(a) \mid h_t)  \label{Equation: integral form risk function}
\end{split}
\end{equation}
In this context, instead of using an infinite horizon of rewards and risks, we exploit the accumulated rewards and risks within a horizon of $(W+1)$-step in the future as the value function $V^\pi(h_t)$ and the risk function $C_i^{\pi}(h_t)$, as shown in Eq. \eqref{Equation: integral form value function} and \eqref{Equation: integral form risk function}, respectively. This formulation clarifies the central question of this paper: can we ensure the safety of PSR with an $\epsilon-$suboptimal policy almost surely? 
\begin{equation}
    \mathbb{P} \biggl( \tilde{V}^{\tilde{\pi}}(h_t) \geq V^{\pi^*}(h_t) - \epsilon \quad \text{and} \quad   \tilde{C}_{i}^{\tilde{\pi}}(h_t) \leq \Bar{C} + \epsilon, \ \forall h_t \in [H], \ i \in [N] \biggr) \geq 1 - \delta    
    \label{Equation: objective of this paper}
\end{equation}
However, it is difficult to directly estimate the probability distribution of PSR due to the singularity of function approximation \cite{xu2022importance, watanabe2009algebraic}. To solve the problem, we construct a kernel representation of PSR instead of approximating the probability distribution of PSR to guarantee the condition in Eq. \eqref{Equation: objective of this paper}. 

\section{Kernel Mean Embedding of PSRs} 
PSRs have a generalized form that can be represented with an inner product. After introducing the corresponding properties of PSR in the inner product, we can use RKHS to perform kernel mean embedding of PSRs, which can cover a wide range of dynamical systems. Furthermore, kernel mean embedding PSRs have more compact forms to represent value and risk functions without any probabilistic inference. 
\subsection{PSR in Inner Product Space} \label{Subsection: PSR in Inner Product Space}
According to the definition of PSR, there can be such a test set denoted as $ \mathcal{U} \subset \bigcup_{W \in \mathbb{N}} \mathcal{A}^{W} \times \mathcal{O}^W \times H$. We can assert for any given reachable history $h_t \in H$, there exists a linear function such that $m_{t_h, h_t} \in \mathbb{R}^{\lvert \mathcal{U}_h \rvert}$ satisfying
\begin{equation}
    \mathbb{P}(t_h \mid h_t) = \langle m_{t_h, h_t}, [\mathbb{P}(t_h  \mid h_t)]_{u_h \in \mathcal{U}_h}  \rangle \label{Equation: inner product test in PSR} 
\end{equation}
where the functional $ [\mathbb{P}(t_h  \mid h_t)]_{u_h \in \mathcal{U}_h}$ is the probability simplex, which can be similar to weighted spectral. Under the formulation in Eq. \eqref{Equation: inner product test in PSR}, the conditional observation probability can be represented with a similar form such that
\begin{equation}
       \mathbb{P}(t_h(o) \mid h_t,t_h(a)) = \langle m_{t_h(o), t_h(a), h_t},  [\mathbb{P}(t_h(o) \mid h_t, t_h(a))]_{t_h(o) \in \mathcal{U}_{\mathcal{O},h}, t_h(a) \in \mathcal{U}_{\mathcal{A},h}}  \rangle;  \label{Equation: cond t_h(o) inner product}
\end{equation}
where $ [\mathbb{P}(t_h(o) \mid h_t, t_h(a))]_{t_h(o) \in \mathcal{U}_{\mathcal{O},h}, t_h(a) \in \mathcal{U}_{\mathcal{A},h}}$ is the conditional observation probability simplex under the space of $\mathbb{I}_{h_t} \otimes \ \mathcal{U}_{\mathcal{O}} \otimes  \ 
 \mathcal{U}_{\mathcal{A}}$ \footnote{$\mathbb{I}_{h_t}$ is the indicator function of $h_t \in H$}. Here, the tensor product $\otimes$ among $H$, $\mathcal{O}$ and $\mathcal{A}$ indicates the dimension of $[\mathbb{P}(t_h(o) \mid h_t, t_h(a))]_{t_h(o) \in \mathcal{U}_{\mathcal{O},h}, t_h(a) \in \mathcal{U}_{\mathcal{A},h}}$ is $\lvert \mathcal{U}_{\mathcal{O},h} \rvert \times \lvert \mathcal{U}_{\mathcal{A},h} \rvert$ when $h_t$ is given. The Eq. \eqref{Equation: inner product test in PSR} and \eqref{Equation: cond t_h(o) inner product} revealed the probability $t_h$ and observation probability $t_h(o)$ can be represented symmetrically in the inner product space as linear forms. However, when the $\lvert \mathcal{U}_{\mathcal{O},h} \rvert \times \lvert \mathcal{U}_{\mathcal{A},h} \rvert$ is infinitely large, the tabular (i.e., conventional) RL becomes insufficient to measure the corresponding probability. Under the inspiration of PSR, we can further infer the mean embedding result of observations under the given actions and historical information. This result will be discussed in the next subsection. 

\subsection{PSR in RKHS}
The kernel method as a universal function approximator has been widely used to express complex structures of learning problems  \cite{fukumizu2011kernel, muandet2017kernel, micchelli2006universal}. The basics of RKHS and related properties have been listed in Appendix \ref{Appendix: Basics of RKHS}.  The relationships among histories, action states, latent states, and observations in PSR can naturally come together through RKHS because they can be represented non-parametric within Hilbert spaces. This representation is accomplished through kernel regression, enabling the mean embedding of future information to be updated through diverse operators. This section will demonstrate how to incorporate PSR connections into RKHS.

To maintain consistency in our definitions (see Appendix \ref{Appendix: Basics of RKHS}), we will use identical symbols for both definite measurable spaces and their corresponding RKHSs within the context of PSRs. For the histories $H$, tests $\mathcal{U}_{\mathcal{O}}$ and $\mathcal{U}_{\mathcal{A}}$ in PSRs, we use the $(H, \mathcal{B}_{H})$, $(\mathcal{U}_{\mathcal{O}}, \mathcal{B}_{\mathcal{U}_{\mathcal{O}}})$ and $(\mathcal{U}_{\mathcal{A}}, \mathcal{B}_{\mathcal{U}_{\mathcal{A}}})$ to denote the measurable functions, and $(\mathcal{H}_{H},k_{H})$, $(\mathcal{H}_{\mathcal{U}_{\mathcal{O}}}, k_{\mathcal{U}_{\mathcal{O}}})$ and $(\mathcal{H}_{\mathcal{U}_{\mathcal{A}}}, k_{\mathcal{U}_{\mathcal{A}}})$ denote the corresponding RKHSs. We will continue to use the symbols such that $ t_h(o) \mapsto \phi^{\mathcal{O}}(t_h(o))$, $ s \mapsto \phi^{\mathcal{S}}(s)$, $ t_h(a) \mapsto \phi^{\mathcal{A}}(t_h(a))$ and $ h \mapsto \phi^{H}(h)$ represent the future observation features, future latent state features, future action features, and history features, respectively. The corresponding feature maps $\phi^\mathcal{O}, \phi^\mathcal{S}, \phi^\mathcal{A}, \phi^{H}$ are their kernel functions in their own RKHS as $\mathcal{H}_{\mathcal{U}_{\mathcal{O}}}$, $\mathcal{H}_{\mathcal{U}_{\mathcal{S}}}$, $\mathcal{H}_{\mathcal{U}_{\mathcal{A}}}$, $\mathcal{H}_{H}$. \\ 
\newline
\textbf{Lemma 1. (Kernel Baye’s Rule (KBR) for multiple elements)}. Introducing a new measurable space $(\mathcal{Z}, \mathcal{B}_{\mathcal{Z}})$ with corresponding RKHS $(H_{\mathcal{Z}}, k_{\mathcal{Z}})$. Extending the embedding theorem to multiple elements $(X, Y, Z)$ ($X$ and $Y$ keep consistency with the basic definition in RKHS), we have the following properties:   
\begin{itemize}
    \item The uncentred cross-variance of multiple elements can be represented as
    \begin{align*}
        \Sigma_{X, Y, Z} = \mathbb{E}[\phi^{X}(x) \otimes \phi^{Y}(y) \otimes \phi^{Z}(z)]
    \end{align*}
    \item The conditional operator $\Sigma_{XY\mid Z}: H_{\mathcal{Z}} \rightarrow H_{\mathcal{X}} \otimes H_{\mathcal{Y}}$ can be represented as
    \begin{align*}
        \Sigma_{XY\mid Z} = \Sigma_{X, Y, Z} \Sigma_{ZZ}^{-1} 
    \end{align*}
    \item The other conditional operator $\Sigma_{X \mid Y, Z}: H_{\mathcal{Z}} \otimes H_{\mathcal{Y}} \rightarrow H_{\mathcal{X}}$ can be 
    \begin{align*}
        \Sigma_{X \mid Y, Z} = \Sigma_{X, Y, Z} [\Sigma_{ZY} \otimes \Sigma_{ZY}^{*}]^{\dag}
    \end{align*}
\end{itemize}
See the detailed description and proof in Appendix \ref{Appendix: proof of lemma 1}. 

\vspace{12 pt}

\textbf{Proposition 1. (Relationships of tests and histories of PSR in RKHSs)} By the Lemma 1 in Appendix \ref{Appendix: Basics of RKHS}, various operators can interpret the relationship between tests and histories. By mapping PSR elements into RKHS, 
\begin{itemize}
    \item we can translate the connection in the covariance matrix into mean embedding in the following kernel descriptions:
    \begin{equation}
        \Sigma_{\mathcal{O},\mathcal{A}, H}  = \mathbb{E} [\phi^{\mathcal{O}}(t_h(o)) \otimes \phi^{\mathcal{A}}(t_h(a))  \otimes \phi^{H}(h)] \label{cov_OAH}
    \end{equation}
    The cross-variance is induced by the Borel set $\mathcal{B}_{ \mathcal{U}_{\mathcal{O}}, \mathcal{U}_{\mathcal{A}}, H} = \mathcal{B}_{\mathcal{U}_{\mathcal{O}}} \otimes \mathcal{B}_{\mathcal{U}_{\mathcal{A}}} \otimes \mathcal{B}_{H}$, see Lemma 1 and Eq. \eqref{Equation: cond t_h(o) inner product}. 
    \item the conditional operator of $\Sigma_{\mathcal{O, A} \mid H}$ indicates the conditional expectation of $\phi(t_h) \defeq \phi^{\mathcal{O}}(t_h(o)) \otimes \phi^{\mathcal{A}}(t_h(a)) $ under arbitrary history $h_t$ as 
    \begin{equation}
    \begin{split}
        \Sigma_{\mathcal{O},\mathcal{A} \mid H} & =  \Sigma_{\mathcal{O},\mathcal{A}, H} \Sigma_{HH}^{\dag} \label{Equation: conditional operator 1 in PSR}
    \end{split}
    \end{equation}
    where according to inner product property in Section \ref{Subsection: PSR in Inner Product Space} and Lemma 1, the linear operator $\Sigma_{\mathcal{O},\mathcal{A} \mid H} \in L(\mathcal{H}_{H},  H_{\mathcal{U}_{\mathcal{O}}} \otimes  H_{\mathcal{U}_{\mathcal{A}}})$. 
    \item  Recursively using the properties in Eq. \eqref{Equation: conditional operator 1 in PSR}, we have 
    \begin{equation}
    \begin{split}
        \Sigma_{\mathcal{O} \mid\mathcal{A}, H} & =  \Sigma_{\mathcal{O},\mathcal{A}, H} \Sigma_{HH}^{\dag} \Sigma_{\mathcal{A,A} \mid H} \\
        & = \Sigma_{\mathcal{O},\mathcal{A}, H} [\Sigma_{H \mathcal{A}} \otimes \Sigma_{H \mathcal{A}}^{*}]^{-1}
        \label{Equation: conditional operator 2 in PSR}        
    \end{split}
    \end{equation}
    where the last line of the equation is due to the tensor product on RKHS, see Appendix \ref{Appendix: Some Important Definitions and Properties of PSRs} and \cite{jakobsen1979tensor};  the linear operator $\Sigma_{\mathcal{O},\mathcal{A} \mid H} \in L(\mathcal{H}_{H}  \otimes  H_{\mathcal{U}_{\mathcal{A}}},   H_{\mathcal{U}_{\mathcal{O}}})$. 
    \item  Introducing the latent test state set $\mathcal{U}_{\mathcal{S}}$ with RKHS as $\mathcal{H}_{\mathcal{U}_{\mathcal{S}}}$ and
    recursively using the properties in Eq. \eqref{Equation: conditional operator 1 in PSR}, we have a similar cross-variance operator $\Sigma_{\mathcal{S, O }, H}$:
    \begin{equation}
        \Sigma_{\mathcal{S}, \mathcal{O}, H} = \mathbb{E} [\phi^{\mathcal{S}}(t_h(s)) \otimes \phi^{\mathcal{O}}(t_h(o))  \otimes \phi^{H}(h)].
        \label{Cov_SOH}        
    \end{equation}
    and conditional operator $\Sigma_{\mathcal{S \mid O }, H}$ can be analogous to a Bayesian filter or Kalman filter to estimate the true state (as observable operators in Eq. \eqref{Equation: pull-back value}), which can be represented as 
    \begin{equation}
        \Sigma_{\mathcal{S \mid O }, H} = \Sigma_{\mathcal{S},\mathcal{O}, H} [\Sigma_{H \mathcal{O}} \otimes \Sigma_{H \mathcal{O}}^{*}]^{-1} 
        \label{Equation: adaptive Kalman Filter}
    \end{equation}
    The operator $\Sigma_{\mathcal{S \mid O }, H} \in L(\mathcal{H}_{H} \otimes \mathcal{H}_{\mathcal{U}_{\mathcal{O}}}, \mathcal{H}_{\mathcal{U}_{\mathcal{S}}})$
\end{itemize}

 The proof of Proposition 1 is highly similar to the proof of Lemma 1 in Appendix \ref{Appendix: Basics of RKHS}, so we omitted the proof details.  

\vspace{12 pt}
\textbf{Remark.}
\begin{itemize}
    \item Eq. \eqref{cov_OAH} is an observation-action-history cross-variance tensor since the joint probability of triples induced a tensor form of Borel sets. The tensor of weighting Hilbert basis of triples reveals the multi-linearity (see definition of the tensor ring in Appendix \ref{Appendix: Some Important Definitions and Properties of PSRs}), on the other hand, the Fubini theorem also indicates the multi-linearity. 
    Similar to Lemma 1, the conditional operator in Eq. \eqref{Equation: conditional operator 1 in PSR} can help to estimate the mean embedding of $\mathcal{H}_{\mathcal{U}_{\mathcal{O}}} \otimes  \mathcal{H}_{\mathcal{U}_{\mathcal{A}}}$. This core idea is no different between Eq. \eqref{Equation: observation conditional probability} and \eqref{Equation: inner product test in PSR}, it directly obtains the conditional expectation instead of measuring the likelihood. By the KBR, the conditional expectation of the test observations can be derived by Eq. \eqref{Equation: conditional operator 2 in PSR}, which is similar to Eq. \eqref{Equation: observation conditional probability} and \eqref{Equation: cond t_h(o) inner product}. 
    
    \item Eq. \eqref{Cov_SOH} and Eq. \eqref{Equation: adaptive Kalman Filter} links the latent state $\mathcal{S}$ and observation $\mathcal{O}$ under the histories. The idea behind the conditional operator can be referred to the Eq. \eqref{Equation: Observable operators}, where the inverse of observable operators $\mathbb{O}^{\dag}: [H] \times L^1(\mathcal{O}) \rightarrow L^1(\mathcal{S})$ is actually filtering the adaptive conditional probability of latent state. $\Sigma_{\mathcal{S \mid O}, H}$ as a mean embedding operator (pull-back the information from $\mathcal{H}_{\mathcal{U}_{O}}$ to $\mathcal{H}_{\mathcal{U}_{S}}$), calculates the expectation of latent states under given observations and histories information. 
\end{itemize}
According to the definition of the conditional operator in Eq. \eqref{uncentred covariance} in Appendix \ref{Appendix: Basics of RKHS}, The expectation of future test observation can be computed under a given history and test actions, the operator $\Sigma_{\mathcal{O} \mid \mathcal{A}, h_t}: \mathcal{H}_{\mathcal{U}_{\mathcal{A}}} \rightarrow  \mathcal{H}_{\mathcal{U}_{\mathcal{O}}}$ is defined on the history $h_t \in H$, intervening by any test actions $ t_h(a) \mapsto \phi^{\mathcal{A}}(t_h(a)), \forall t_h(a) \in \mathcal{A}^{W}$, we have
\begin{equation}
\begin{split}
    \mathbb{E} [\phi^{\mathcal{O}}(t_h(o)) \mid \phi^{\mathcal{A}}(t_h(a)), h_t] = \Sigma_{\mathcal{O} \mid \mathcal{A}, h_t} \phi^{\mathcal{A}}(t_h(a))
    \label{Equation: projection operator}
\end{split}
\end{equation}
The form is the integral form of Eq. \eqref{Equation: observation conditional probability}. For arbitrary functional $f$ defined $t_h(o)$ can be calculated such that
\begin{equation}
\begin{split}
        \mathbb{E}[f(t^h(o)) \mid \phi^{\mathcal{A}}(t_h(a)), h_t] & = \langle f, \mathbb{E}[\phi^{\mathcal{O}}(t_h(o)) \mid \phi^{\mathcal{A}}(t_h(a)), h_t] \rangle \\ 
        & = \langle f, \Sigma_{\mathcal{O} \mid \mathcal{A}, h_t} \phi^{\mathcal{A}}(t_h(a) \rangle
\end{split}
\end{equation}
For the practical calculation, the operator $\tilde{\Sigma}_{\mathcal{O} \mid \mathcal{A}, h_t}$ can be estimated by KBR in Eq. \eqref{condition on xy|z}, \eqref{conditional covariance 1} and \eqref{estimation matrix on x|yz} as
\begin{equation}
\begin{split}
    \tilde{\Sigma}_{\mathcal{O} \mid \mathcal{A}, h_t} &= \Sigma_{\mathcal{O,A} \mid h_t} \Sigma_{\mathcal{A,A}\mid h_t}^{-1}  \\
    &= \Phi_{\mathcal{O}} (\Psi_{h_t}\Phi_{\mathcal{A}}\Phi_{\mathcal{A}}^{*} + \lambda I)^{-1}\Psi_{h_t} \Phi_{\mathcal{A}} \label{practical kernel bayes rule}
\end{split}
\end{equation}
where $\Phi_{\mathcal{O}}, \Phi_{\mathcal{A}}, \Phi_{H}, \Psi_{h}$ follow the definition:
\begin{center}
    $\Phi_{\mathcal{O}} = (\phi^{\mathcal{O}}(t_h(o)_{1}), \phi^{\mathcal{O}}(t_h(o)_{2}), \cdots, \phi^{\mathcal{O}}(t_h(o)_{\lvert K \rvert}))$, \\
    $\Phi_{\mathcal{A}} = (\phi^{\mathcal{A}}(t_h(a)_{1}), \phi^{\mathcal{A}}(t_h(a)_{2}), \cdots, \phi^{\mathcal{A}}(t_h(a)_{\lvert K \rvert}))$, \\
    $\Phi_{H} = (\phi^{H}(h_{1}), \phi^{H}(h_{2}), \cdots$, $\phi^{H}(h_{\lvert K \rvert}))$,\\
    $\Psi_{h} = diag ((\Phi_{H}\Phi_{H}^{*} + \lambda I)^{-1} \Phi_{H}^{*}\phi^H(h))$.
\end{center}
Here, the dataset is denoted $ K $, the operator error $\lVert  \tilde{\Sigma}_{\mathcal{O} \mid \mathcal{A}, h_t} -  \Sigma_{\mathcal{O} \mid \mathcal{A}, h_t} \rVert$ relies on the sample size of the dataset, and the error bound will be given a rigorous analysis in the next sections.

Another operator $\Sigma_{\mathcal{S \mid O }, H}: \mathcal{H}_{H} \otimes \mathcal{H}_{\mathcal{U}_{\mathcal{O}}} \rightarrow \mathcal{H}_{\mathcal{U}_{\mathcal{S}}}$ for filtering the true state will not be explicit calculation in the context, since the true state sometimes is agnostic. On the other hand, we have indicated the random matrix $\Sigma_{\mathcal{S \mid O }, H}$ has a barycenter, which preserves the same unique value with the pull-back integral forms see Eq. \eqref{Equation: pull-back value} and \eqref{Equation: pull-back risk}. In this situation, the latent states are not necessary to be estimated. The "ghost" operator $\Sigma_{\mathcal{S \mid O }, H}$ will be used as an indeterminate to derive some relationship under the framework. It is worth mentioning that the kernel mean embedding PSR can cover a wide range of controlled dynamical systems, two examples (LQR and POMDP) are given in Appendix \ref{Appendix. Generalization of Kernel Mean Embedding PSRs}. 

\section{Forward Operators in PSRs and Function Approximation}
We have introduced a variety of conditional operators within the context of kernel mean embedding PSRs. The utilization of these operators offers a concise formulation for both filtration and function approximation. In this section, we initially present five crucial operators that extend from the PSRs in RKHS. Building upon these operators, we establish a universal representation for the value/risk function and Bellman optimality. Subsequently, we address three primary questions:

\vspace{12pt}

\textbf{(a)} How can we construct operators to facilitate the learning of PSRs?

\textbf{(b)} How can we create operator-based value and risk functions?

\textbf{(c)} What is the analysis of Bellman optimality under these forward operators?

\subsection{Operators}
The benefits of using operator-assisted control in dynamical systems are multifaceted and can be summarized into three key points: \textbf{a) Infinite Size and Discretization-Invariance:} Unlike the traditional Reinforcement Learning (RL) approach that discretizes large spaces into smaller grids,  our method enables direct representation on a continuum. It avoids the limitations of tabular methods. \textbf{b) Output as Function:} This approach holds the potential for representing complex dynamical systems, such as fluid flow and plasma flow, where the output is inherently functional. \textbf{c) Universal Approximation:} The method demonstrates sample efficiency in modeling and controlling large-scale nonlinear systems, serving as a universal approximator for a wide range of dynamics.

Following the definition of conditional operator, we will construct five operators under PSR in the RKHS context. The five operators have internal connections with each other, and the definitions are as follows:
\begin{itemize}
    \item \textbf{One-step forward Operator.} One-step forward prediction involves predicting a single-step observation, taking into account an arbitrary action $a_{t-1} \in \mathcal{A}$ within the context of a given history $h_t \in H$. The one-step forward operator $\Sigma_{o \mid a, H} \in L(\mathcal{H}_{H} \otimes \mathcal{H}_{a}, \mathcal{H}_{o})$, we can obtain the conditional expectation of $\phi^{o}(o_t) \in \mathcal{H}_{o}$ is 
    \begin{equation}
    \begin{split}
        \mathbb{E}[\phi^{o}(o_t) \mid \phi^{a}(a_{t-1}), h_t] = \Sigma_{o \mid a, h_t}  \phi^{a}(a_{t-1})
        \label{Equation: one-step dynamics based on operator}
    \end{split}
    \end{equation}
    \item \textbf{Forward operator.} The forward operator $\Sigma_{\mathcal{O \mid A}, H} \in L( \mathcal{H}_{H} \otimes \mathcal{H}_{\mathcal{U}_\mathcal{A}},  \mathcal{H}_{\mathcal{U}_\mathcal{O}})$ is for prediction of future observation with $W-$step based on the history $h_t$ and $W-$step action $t_h(a)$, it has the same definition as the conditional operator described in Eq. \eqref{Equation: projection operator}. Here, we repeat it again as follows:
    \begin{equation}
        \mathbb{E} [\phi^{\mathcal{O}}(t_h(o)) \mid \phi^{\mathcal{A}}(t_h(a)), h_t] = \Sigma_{\mathcal{O} \mid \mathcal{A}, h_t} \phi^{\mathcal{A}}(t_h(a))
    \end{equation}
    \item \textbf{Shifted forward operator.} The shifted forward operator $\Sigma_{\mathcal{O \mid A}, h_{t+1}}$ is defined on the history $h_{t+1}$. After one-step dynamics $(o_t, a_{t-1})$ is revealed, the shifted operator will predict the future $W-$step observation after $h_{t+1} \defeq h_t \oplus (o_t, a_{t-1})$. The core idea behind this operator is similar to Eq. \eqref{Equation: shifted observation probablity}, using an adaptive way to filter future observation. It is worth mentioning that the shifted forward operator lies in the same category as the forward operator. We have the shifted observation prediction as: 
    \begin{equation}
        \mathbb{E} [\phi^{\mathcal{O}}(t_{h+1}(o)) \mid \phi^{\mathcal{A}}(t_{h+1}(a)), \underbrace{h_t, o_t, a_{t-1}}_{h_{t+1}}] = \Sigma_{\mathcal{O} \mid \mathcal{A}, h_{t+1}} \phi^{\mathcal{A}}(t_{h+1}(a))
    \end{equation}
    where $t_{h+1}(o) = o_{t+1:t+W}$ and $t_{h+1}(a) = a_{t:t+W-1}$.
    \item \textbf{Shifted operator.} The shifted operator is actually a lifted operator (or hom-functor \cite{banaschewski1976tensor}) defined on the forward operator as $\mathcal{P}\in Hom(L( \mathcal{H}_{H} \otimes \mathcal{H}_{\mathcal{U}_\mathcal{A}},  \mathcal{H}_{\mathcal{U}_\mathcal{O}}), L( \mathcal{H}_{H} \otimes \mathcal{H}_{\mathcal{U}_\mathcal{A}},  \mathcal{H}_{\mathcal{U}_\mathcal{O}}))$. Under the shifted operator, the forward operator will be mapped to the shifted forward operator. For a given one-step dynamics $(o_t, a_{t-1})$ and history $h_t$, we have 
    \begin{center}
        $\mathbb{E}[\Sigma_{\mathcal{O}\mid \mathcal{A}, h_{t+1} }] = \mathcal{P}_{o_{t}, a_{t-1}} \Sigma_{\mathcal{O}\mid \mathcal{A}, h_t }$
    \end{center}
    and 
    \begin{equation}
    \begin{split}
        \mathbb{E}[\phi^{\mathcal{O}}(t_{h+1}(o)) \mid \phi^{\mathcal{A}}(t_{h+1}(a)), \underbrace{h_t, o_t, a_{t-1}}_{h_{t+1}}] & = \underbrace{(\mathcal{P}_{o_{t}, a_{t-1}} \Sigma_{\mathcal{O}\mid \mathcal{A}, h_t } )}_{\text{shift forward}} \phi^{\mathcal{A}}(t_{h+1}(a)) \label{Equation: shifted W-step observation}
    \end{split}
    \end{equation}
    \item \textbf{Extended forward operator.} The extended forward operator $\Sigma_{\mathcal{O}, o \mid \mathcal{A}, a, H }$ is for the prediction of extended observation as $\tilde{t}_{h}(o) \defeq o_{t: t+W} $, the category of extended forward operator is $L(\mathcal{H}_{H} \otimes \mathcal{H}_{a} \otimes \mathcal{H}_{\mathcal{A}}, \mathcal{H}_{\mathcal{O}} \otimes \mathcal{H}_{o})$. Under a given history $h_t$, we have the prediction as 
    \begin{equation}
    \begin{split}
        & \ \quad \mathbb{E}[\phi^{\mathcal{O}}(t_{h+1}(o)) \otimes \phi^{o}(o_t) \mid \phi^{\mathcal{A}}(t_{h+1}(a)), \phi^{a}(a_{t-1}), h_t] \\ 
        & = \Sigma_{\mathcal{O}, o \mid \mathcal{A}, a, h_t } [\phi^{\mathcal{A}}(t_{h+1}(a)) \otimes \phi^{a}(a_{t-1})] 
        \label{Equation: extended observation under extended operator}
    \end{split}
    \end{equation}
\end{itemize}

\vspace{12pt}

\textbf{Remark.}
In this context, we give a mild assumption of the ergodicity of the dynamical system, which will be provable for its error bound. These operators naturally arise from the conditional operators in RKHS and exhibit interconnections. Each operator can be represented by the others, underscoring their inherent relationships. The one-step forward operator can be employed to derive the other forward operators, given that each step must adhere to the local relationships associated with one-step dynamics. The forward operator and shifted forward operator are connected by the shifted operator $\mathcal{P}$. It is interesting to mention when the inverse shifted operator $\mathcal{P}^{\dag}$ is defined on the left side of the shifted forward operator, it will become the backward dynamics, which can be analogous to the Fokker-Planck equation. The operator $\mathcal{P}$ can be represented by two forward operators as $\mathcal{P}_{o_t,a_{t-1}} = \Sigma_{\mathcal{O \mid A}, h_t, (o_t, a_{t-1})} \Sigma_{\mathcal{O \mid A}, h_t}^{-1}$. Furthermore, the one-step forward and shifted forward operators can represent the extended forward operators. For the given $h_t$, The prediction of extended prediction is $\phi^{\mathcal{O}}(t_{h+1}(o)) \otimes \phi^{o}(o_t)$ under the intervening of $\phi^{\mathcal{A}}(t_{h+1}(a)) \otimes \phi^{a}(a_{t-1})$. Based on this fact, we have
\begin{align*}
    \Rightarrow & \quad \Sigma_{\mathcal{O}, o_t \mid \mathcal{A}, a_t, h_t } \\
    &= \Sigma_{o_t \mid h_t, a_{t-1}} \otimes \Sigma_{\mathcal{O} \mid \mathcal{A}, h_{t+1} } \\
    &=\Sigma_{o_t \mid h_t, a_{t-1}} \otimes (\mathcal{P}_{o_{t}, a_{t-1}} \circ \Sigma_{\mathcal{O}\mid \mathcal{A}, h_t } )
\end{align*}
The results show the bijection relationship $\Sigma_{\mathcal{O}, o_t \mid \mathcal{A}, a_t, h_t }\simeq \Sigma_{o_t \mid h_t, a_t} \otimes \Sigma_{\mathcal{O} \mid \mathcal{A}, h_{t+1} } \simeq \Sigma_{o_t \mid h_t, a_t} \otimes \mathcal{P}_{a_{t}, o_{t+1}} \circ \Sigma_{\mathcal{O}\mid \mathcal{A}, h_t }$. The operator will be essential in the analysis of Bellman optimality in the following subsections, a brief introduction in Eq. \eqref{Equation: pull-back value} and \eqref{Equation: pull-back risk} have shown both value and risk functions are defined on the $(W+1)-$step states, which have natural connections with the prediction of the extended forward operator. The rollout of each step action will shift the dynamics to the next $W-$step prediction, in this way, it allows to decomposition of the $(W+1)-$step as one-step dynamics and shift $W-$step dynamics. After estimating those well-defined operators on the RKHS,  the following subsection will leverage the property of \textit{bilinearity} of Hilbert space and use the bilinear form to represent value/risk functions. Notably, in contrast to the probabilistic representation, the operator $\mathcal{P}_{o_{t}, a_{t-1},}$ is directly defined on the mean embedding results without the need for likelihood inference. In the following section, we will introduce the concept of "link functions," allowing us to provide bilinear value/risk functions in a partially observable environment by combining these operators in RKHS. 

\subsection{Value and Risk Link Functions}
Following the probabilistic version of value/risk functions in Eq. \eqref{Equation: integral form value function} and \eqref{Equation: integral form risk function} in Section \ref{Subsection: Safe Reinforcement Learning Under Probabilistic PSR}, we define a value function for a policy $\pi$ at step $t$ as the expected accumulated reward under the policy starting from $h_t \in H$ and $s \in \mathcal{S}$ such that $V_{t}^{\pi}: H \rightarrow \mathbb{R} $ and $C_{t}^{\pi}: H \rightarrow \mathbb{R} $, where:
\begin{align*}
   & V^\pi(h_t) = \mathbb{E} [ \sum_{t^{'} = t}^{t+W} r(s_{t^{'} }) \mid h_t = h, a_{t-1:t+W-1} \sim \pi ] \\
   & C_{i}^\pi(h_t) = \mathbb{E} [ \sum_{t^{'} = t}^{t+W} c_{i}(s_{t^{'} }) \mid h_t = h, a_{t-1:t+W-1} \sim \pi ], \quad i \in [N]
\end{align*}
The above value function describes the conditional accumulated reward in horizon $t$ to $t+W$ with the given policy and histories. Compared with the standard MDP, the reward expectation is conditional not only on $s_t$ but also on $h_t$. Then, the corresponding value function can be decomposed as: 
\begin{equation}
    V^{\pi} (h_t) = \mathbb{E}[r(s_{t}) + V^{\pi} (h_{t+1}) \mid h_t, s_t,  a_{t-1} \sim \pi] \label{Equation: conventional value function}
\end{equation}
Where $h_{t+1} = h_t \oplus (s_t, a_{t-1})$. Similarly, the accumulated risk function becomes: 
\begin{equation}
     C_{i}^{\pi} (h_t) = \mathbb{E}[c(s_{t}) + C_{i}^{\pi} (h_{t+1}) \mid h_t, s_t,  a_{t-1} \sim \pi], \quad i \in [N]    
     \label{Equation: conventional risk function}
\end{equation}

In contrast to conventional MDPs, working directly with value functions $V_{t}^{\pi}(s)$ (or $Q$ functions) in POMDPs is not straightforward, as they lack memory of the state history before time $t$. Furthermore, the value/risk function defined in the previous equations is based on states rather than state histories, making it agnostic to the true state in general cases. To address these challenges, we introduce a novel operator-driven framework representing the value/risk function in partially observable environments. Given the presence of unknown latent states, building the value/risk functions directly as in Eq. \eqref{Equation: conventional value function} and \eqref{Equation: conventional risk function} is not feasible. Therefore, we introduce a concept known as the "link function" in advance. Ultimately, synthesising the forward operators will provide a bilinear form for the value/risk function in RKHS.

\textbf{Definition. (1-step value/risk link functions).} Suppose there is a fixed set of policies $\pi = \{\pi_i\}_{i=1}^{W+1}$ where $\pi_i: [H] \times \mathcal{O} \rightarrow \Delta(\mathcal{A})$. The 1-step value link function $g: [H] \times \underbrace{\mathcal{O} \times \mathcal{A}}_{\text{one step}} \rightarrow \mathbb{R}$ at step $t$ for a policy $\pi$ is defined as the solution to the following integral equation:

\begin{equation}
\begin{split}
   & V^{\pi} (h_t) \defeq \mathbb{E}[g^{\pi}(h_t, o_t) \mid h_t, s_t, a_{t-1} \sim \pi_1]  \quad  \forall h_t \in H, s_t \in \mathcal{S}
    \label{Equation: 1-step measure_v}
\end{split}
 \end{equation}
where the expectation is taken under the policy $\pi$. Similarly, the risk link function is $m: [H] \times \mathcal{O} \times \mathcal{A} \rightarrow \mathbb{R}$ with the integral formulation:
\begin{equation}
\begin{split}
   &C^{\pi} (h_t) \defeq  \mathbb{E}[m^{\pi}(h_t, o_t) \mid h_t, s_t, a_{t-1} \sim \pi_1]  \quad \forall h_t \in H, s_t \in \mathcal{S}
    \label{Equation: 1-step measure_c}
\end{split}
\end{equation}
 The value/risk link functions demonstrate our ability to derive the mean embedding of value/risk functions onto the observation space. It's noteworthy that we have established an intuitive symmetry in the link functions, as evidenced by Eq. \eqref{Equation: 1-step measure_v} and \eqref{Equation: 1-step measure_c}, which transition to the one-step expectations as shown in Eq. \eqref{Equation: pull-back value} and \eqref{Equation: pull-back risk}. To be more specific, the one-step link function $(h_t, o_t, a_{t-1}) \mapsto g(h_t, o_t, a_{t-1})$ can be considered as the inverse of the observable operator $(h_t, o_t, a_{t-1}) \mapsto r\circ \mathbb{O}^{\dag}(h_t, o_t, a_{t-1})$.

On the other hand, it's important to note that both the value and risk link functions need not be unique, as link functions are implicitly expressed, allowing us to embed latent states into the observation space. However, it should be emphasized that even though the link functions may not be unique, the induced value and risk functions are unique. This fact will be demonstrated later. Before that, we need to establish that link functions are sufficient to transfer observations to the latent state value/risk function (see Lemma 2 Appendix \ref{Appendix: Bilinear Form and Function Approximation}. Here, we extend the one-hot encoding result \cite{uehara2022provably} to a general case in RKHS. 

\textbf{Extension to $(W+1)-$step value/risk link functions.} Since the 1-step link functions are well-defined, the $(W+1)$-step value/risk function $g: [H] \times \underbrace{\mathcal{O}^{W+1} \times \mathcal{A}^{W+1}}_{(W+1)-\text{step}} \rightarrow \mathbb{R}$ can be represented as:
\begin{equation}
\begin{split}
   &V^{\pi} (h_t) \defeq \mathbb{E}[g^{\pi}(h_t, o_{t:t+W}) \mid h_t, s_{t:t+W}, a_{t-1:t+W-1} \sim \pi]  ; 
    \\& \forall h_t \in H, s_{t:t+W} \in \mathcal{S}^{W+1}.
\end{split}
\label{multi value}
\end{equation}
The corresponding risk function is:
\begin{equation}
\begin{split}
   &C^{\pi} (h_t) \defeq \mathbb{E}[m^{\pi}(h_t, o_{t:t+W-1}) \mid h_t, s_{t:t+W}, a_{t-1:t+W-1} \sim \pi]  ; 
    \\& \forall h_t \in H, s_{t:t+W} \in \mathcal{S}^{W+1}.
\end{split}
\label{multi risk}
\end{equation}
The core idea behind the extension of one-step link functions to the $(W+1)-$step functions is similar to the transition from Eq. \eqref{Equation: pull-back value} and \eqref{Equation: pull-back risk} to Eq. \eqref{Equation: integral form value function} and \eqref{Equation: integral form risk function}. We employ a fixed-length horizon value/risk function in this scenario to guide the dynamical system. In the following subsections, we will illustrate how the defined operators can represent the value/risk functions using bilinear forms with link functions within RKHS.

\subsection{Function Approximation with Bilinear Form in RKHS}

Before connecting the link functions and RKHS, an important step is to prove the existence of value link functions such as $g, m$ for any $\pi$ and $h$. Previous work of \cite{uehara2022provably} demonstrated the formulation in the one-hot encoding scenario, and we generalize the transformation for any $(h_t,o_{t}, a_{t-1})$ into a generalized form in RKHS.  

\textbf{Lemma 2. (the existence of link functions)} For any separable functions $V$ and $C$ lie in Hilbert spaces, corresponding link functions always exist to represent $V$ and $C$ for any policy $\pi$. (see detailed Proof in Appendix \ref{Appendix: Bilinear Form and Function Approximation})

Hence, for any function $f$ defined on the tensor product of $\phi^H(h_t) \otimes \phi^{o}(o_t) \otimes \phi^{a}(a_{t-1})$ \footnote{It is an equivalent representation for $f^{a_{t-1}}(h_t, o_t) \equiv f(h_t, o_t, a_{t-1}) $. For simplicity, if a function is parameterized by policy $\pi$, we directly write as $f^{\pi}(\cdot, \cdot)$ (or say $dom(f^{\pi}) \subset [H] \times \mathcal{A}$)}, according to the reproducing property, we have:
\begin{equation}
    f(h_t,o_t,a_{t-1}) = \langle f, \phi^H(h_t) \otimes \phi^{o}(o_t) \otimes \phi^{a}(a_{t-1})\rangle.
\end{equation}
where $f$ can be indicated arbitrarily $g, m$ in RKHS. Previously, we have indicated the connections between the $\Sigma_{\mathcal{S} \mid \mathcal{O}, H}$ and $\mathbb{O}^{\dag}$. According to the incompleteness setting in this context (the extension of incompleteness has been listed in Appendix \ref{Appendix: Some Important Definitions and Properties of PSRs}), the embedding operator  $\Sigma_{\mathcal{S} \mid \mathcal{O}, H}$ satisfying the property as 
\begin{align*}
    \Sigma_{\mathcal{S} \mid \mathcal{O}, h_t} \phi^{o}(o_t) = \phi^{\mathcal{S}}(s_t)
\end{align*}
It is due to the fact that $\Sigma_{\mathcal{S} \mid \mathcal{O}, H}$ is the left inverse of $\Sigma_{\mathcal{O} \mid \mathcal{S}, H}$. Using the property, the pull-back process from $\phi^{\mathcal{S}}(s_t)$ to $\Sigma_{\mathcal{S} \mid \mathcal{O}, h_t} \phi^{o}(o_t)$ keeps invariant for any functional defined on its dual space. By the reproducing property in RKHS, it can derive the following result
\begin{equation}
\begin{split}
    f(h,s) &= \langle f, \phi^H(h) \otimes \phi^\mathcal{S}(s) \rangle \\
    &= \langle f, \phi^H(h) \otimes \Sigma_{\mathcal{S} \mid \mathcal{O}, h }\phi^\mathcal{O}(o)   \rangle , 
\end{split}
\label{link function in rkhs}
\end{equation}
where the pull-back information from latent state space to observation space keeps the equivariant of the $f(h,s)$. Since $f$ indicates arbitrary functionals in RKHS, both value/risk functions inherit this property. The existence of the mean embedding operator $\Sigma_{\mathcal{S} \mid \mathcal{O}, H}$ guarantees the observation information can be pulled back to the state space without loss of any information. From this perspective, the core idea of link functions can help to measure value/risk directly on observation space, without inferring any distribution on latent state. In this situation, the well-defined link function can now plug our case, define the value/risk link functions can be represented for arbitrary policy $\pi$, as 
\begin{equation}
    g^{\pi}(h_t, \Tilde{t}_{h}(o)) =  \langle g^{\pi}, \phi^H(h_t) \otimes \phi^{\mathcal{O}}(\Tilde{t}_{h}(o)) \rangle ,
\end{equation}
and
\begin{equation}
    m^{\pi}(h_t, \Tilde{t}_{h}(o)) =  \langle m^{\pi}, \phi^H(h_t) \otimes \phi^{\mathcal{O}}(\Tilde{t}_{h}(o)) \rangle ,
\end{equation}
Then the value/risk function can be expressed as
\begin{equation}
\begin{split}
        V^{\pi} (h_t)
       & = \mathbb{E}[\langle g^{\pi}, \phi^H(h_t) \otimes \phi^{\mathcal{O}}(\Tilde{t}_{h}(o)) \rangle \mid h_t, \Tilde{t}_{h}(a) \sim \pi] \label{Equation: value function bilinear form}
\end{split}
\end{equation}
\begin{equation}
\begin{split}
        C_{i}^{\pi}(h_t) = \mathbb{E} [\langle m_{i}^{\pi}, \phi^H(h_t) \otimes \phi^{\mathcal{O}}(\Tilde{t}_{h}(o)) \rangle  \mid h_t, \Tilde{t}_{h}(a) \sim \pi] , \quad i \in [N] \label{Equation: risk function bilinear form}
\end{split}
\end{equation}
After obtaining the bilinear form link functions in RKHS to represent the value/risk functions, we can give an explicit form to represent the value/risk function by combining the forward operators. Decompose the Eq. \eqref{Equation: value function bilinear form} and \eqref{Equation: risk function bilinear form} as follows: 
\begin{equation}
\begin{split}
       V^{\pi} (h_t)
       & = \mathbb{E}[ g^{\pi} (h_t, \Tilde{t}_{h}(o) ) \mid h_t, \Tilde{t}_{h}(a) \sim \pi] \\
       & = \underbrace{\mathbb{E}[g^{\pi}(h_t, o_t) \mid h_t, a_{t-1} \sim \pi]}_{\text{one-step value}} \\ 
       & \qquad + \underbrace{\mathbb{E}[ g^{\pi} (h_{t+1}, t_{h+1}(o) ) \mid \underbrace{h_t, o_{t}, a_{t-1} \sim \pi}_{h_{t+1}} , t_{h+1}(a) \sim \pi]}_{\text{shifted $W-$step value}} \\ 
       & = \mathbb{E}[\langle g^{\pi}, \phi^{H}(h_t) \otimes \phi^{o}(o_t) \rangle \mid h_t, a_{t-1} \sim \pi]  \\
       & \qquad + \mathbb{E}[\langle g^{\pi}, \phi^H(h_{t+1}) \otimes \phi^{\mathcal{O}}(t_{h+1}(o)) \rangle \mid h_{t+1}, t_{h+1}(a) \sim \pi] \\
       & = \mathbb{E} [ \langle g^{\pi}_t, \Sigma_{{o \mid a, h_t}} \phi^{a} (a_{t-1}) \rangle \mid h_t, a_{t-1} \sim \pi ] \\
       & \qquad +  \underbrace{\mathbb{E} [\mathbb{E}}_{\text{telescope property}} [\langle g^{\pi}_{t+1}, \Sigma_{\mathcal{O \mid A} , h_{t+1}} \phi^{\mathcal{O}}(t_{h+1}(o)) \rangle \mid t_{h+1}(a) \sim \pi] \mid h_{t},\underbrace{ o_t, a_{t-1} \sim \pi]}_\text{one-step} \\
       & = \mathbb{E} [ \langle g^{\pi}_t, \Sigma_{{o \mid a, h_t}} \phi^{a} (a_{t-1}) \rangle \mid h_t, a_{t-1} \sim \pi ] \\
       & \qquad + \mathbb{E} [ \mathbb{E}[\langle g^{\pi}_{t+1}, \mathcal{P}_{o_t, a_{t-1}}\Sigma_{\mathcal{O \mid A} , h_{t}} \phi^{\mathcal{O}}(t_{h+1}(o)) \rangle \mid t_{h+1}(a) \sim \pi]\mid h_{t}, o_t, a_{t-1} \sim \pi] \\
       & = \mathbb{E}[ \langle g^{\pi}_{t+1}  \otimes g^{\pi}_{t}, \Sigma_{\mathcal{O},o \mid \mathcal{A}, a, h_t} (\phi^{\mathcal{A}}(t_{h+1}(a)) \otimes \phi^{a}(a_{t-1})) \rangle \mid h_t, \tilde{t}_{h}(a) \sim \pi]
    \label{Equation: operator-driven value function}
\end{split}
\end{equation}
Similarly,
\begin{equation}
\begin{split}
    C_{i}^{\pi}(h_t) 
    & = \mathbb{E} [\langle m_{i}^{\pi}, \phi^H(h_t) \otimes \phi^{\mathcal{O}}(\Tilde{t}_{h}(o)) \rangle \mid h_t, \Tilde{t}_{h}(a) \sim \pi] , \quad i \in [N] \\
    & = \mathbb{E} [\langle m^{\pi}_{i,t}, \Sigma_{{o \mid a, h_t}} \phi^{a} (a_{t-1}) \rangle \mid h_t, a_{t-1} \sim \pi ] \\
    & \qquad + \mathbb{E} [ \mathbb{E}[\langle m^{\pi}_{i, t+1}, \mathcal{P}_{o_t, a_{t-1}}\Sigma_{\mathcal{O \mid A} , h_{t}} \phi^{\mathcal{O}}(t_{h+1}(o)) \rangle \mid t_{h+1}(a) \sim \pi]\mid h_{t}, o_t, a_{t-1} \sim \pi] \\
    & = \mathbb{E}[ \langle m^{\pi}_{i, t+1}  \otimes m^{\pi}_{i, t}, \Sigma_{\mathcal{O},o \mid \mathcal{A}, a, h_t} (\phi^{\mathcal{A}}(t_{h+1}(a)) \otimes \phi^{a}(a_{t-1})) \rangle \mid h_t, \tilde{t}_{h}(a) \sim \pi]
    \label{Equation: operator-driven risk function}
\end{split}
\end{equation}

Here, we denote the $g^{\pi}_{t} (\cdot) \equiv g^{\pi}(h_t, \cdot)$ for simplicity, thus the domain of $g^{\pi}_{t} $ becomes $\mathcal{H}_{\mathcal{O}}$. Under the definition of link functions, the $(W+1)-$step value function can be decomposed as two parts: one-step value function represented by the one-step link function and shifted $W-$step value function defined on the history $h_{t+1} = (h_t, o_{t}, a_{t-1})$, the one-step dynamics $(o_{t}, a_{t-1})$ is determined by the policy $\pi$. Due to the reproducing property in RKHS, the link function can be written as the inner product form in the second line of the equation. For the one-step value function, the $\langle g^{\pi}, \phi^{H}(h_t) \otimes \phi^{o}(o_t) \rangle$ is determined by one-step dynamics. By the definition of operators in Eq. \eqref{Equation: one-step dynamics based on operator}, it can derive the one-step value is just $\mathbb{E} [ \langle g^{\pi}_t, \Sigma_{{o \mid a, h_t}} \phi^{a} (a_{t-1}) \rangle \mid h_t, a_{t-1} \sim \pi ]$. After estimating the one-step value under the policy $\pi$, the $W-$step value function is conditioned on the result of one-step dynamics. In such a situation, the $\mathbb{E}[\langle g^{\pi}_{t},\phi^{\mathcal{O}}(t_{h+1}(o)) \rangle \mid h_{t+1}, t_{h+1}(a) \sim \pi]$ is conditioned as $\mathbb{E} [\mathbb{E}[\langle g^{\pi}, \phi^H(h_{t+1}) \otimes \phi^{\mathcal{O}}(t_{h+1}(o)) \rangle \mid t_{h+1}(a) \sim \pi] \mid h_{t}, o_t, a_{t-1}]$, since shifted result relies on $h_{t+1}$. Inspired by this fact, various forward operators can be plugged into the formula to give a more expressive and explicit result.

For the one-step operator-based value function in the fourth line, the $g^{\pi}_{t}$ controls two variables $\pi$ and $h_t$, it determines the conditional variables on $\Sigma_{o \mid a, h}$, then the one-step value can be $\mathbb{E}[ \langle g^{\pi}_{t}, \Sigma_{o \mid a, h} \phi^{a}(a_{t-1}) \rangle \mid h_t, a_{t-1} \sim \pi]$, the ($a_{t-1}, h_t$) conditions the one-step forward operator as $\Sigma_{o \mid a, h}$. Similarly, the shifted $W-$step value function can be $\mathbb{E}[\langle g^{\pi}_{t+1}, \Sigma_{\mathcal{O \mid A} , h_{t+1}} \phi^{\mathcal{O}}(t_{h+1}(o)) \rangle \mid h_{t+1}, t_{h+1}(a) \sim \pi] $, where $g^{\pi}_{t+1}$ controls the variables of shifted forward operator $\Sigma_{\mathcal{O \mid A}, h_{t+1}}$. However, directly obtaining the $h_{t+1}$ is impossible since $h_{t+1}$ relies on the last one-step dynamics, therefore, the $W-$step value function can be rewritten as $\mathbb{E} [\mathbb{E} [\langle g^{\pi}_{t+1}, \Sigma_{\mathcal{O \mid A} , h_{t+1}} \phi^{\mathcal{O}}(t_{h+1}(o)) \rangle \mid t_{h+1}(a) \sim \pi] \mid h_{t}, o_t, a_{t-1} \sim \pi]$ by telescope property of conditional expectation. The first condition determines the $h_{t+1}$, and the second expectation determines the dynamics $t_{h+1}$ under policy $\pi$. By introducing the shifted operator $\mathcal{P}$, it can be derived that $\mathbb{E} [ \mathbb{E}[\langle g^{\pi}_{t+1}, \mathcal{P}_{o_t, a_{t-1}}\Sigma_{\mathcal{O \mid A}, h_{t}} \phi^{\mathcal{O}}(t_{h+1}(o)) \rangle \mid t_{h+1}(a) \sim \pi]\mid h_{t}, o_t, a_{t-1} \sim \pi]$, where the variable of $\mathcal{P}_{o_{t}, a_{t-1}}$ conditioned on the one-step dynamics, so we can give a more compact form as
\begin{align*}
    & \ \quad \mathbb{E} [ \mathbb{E}[\langle g^{\pi}_{t+1}, \mathcal{P}_{o_t, a_{t-1}}\Sigma_{\mathcal{O \mid A} , h_{t+1}} \phi^{\mathcal{O}}(t_{h+1}(o)) \rangle \mid t_{h+1}(a) \sim \pi]\mid h_{t}, o_t, a_{t-1} \sim \pi] \\
    & = \mathbb{E}[\langle g^{\pi}_{t+1}, \mathcal{P}^{\pi} \Sigma_{\mathcal{O \mid A} , h_{t+1}} \phi^{\mathcal{O}}(t_{h+1}(o)) \rangle \mid h_t, a_{t} \sim \pi , t_{h+1}(a) \sim \pi]
\end{align*}
Since the 
\begin{align*}
    \mathbb{E}[\phi^{o}(o_t) \mid a_{t-1} \sim \pi ,  h_t] = \mathbb{E}_{a_{t-1} \sim \pi}[\Sigma_{o \mid a, h_{t-1}} \phi^{a}(a_{t-1})] 
\end{align*}
This form will be essential in analysing the convergence of Bellman loss in the following section. When one-step policy rollout, future $W-$step value is estimated under shifted operator $\mathcal{P}^{\pi}$ and shifted dynamics $\Sigma_{\mathcal{O \mid A}, h_{t+1}}$. For the total value of the $(W+1)-$step, it can symmetrically be represented as a tensor form under the extended forward operator as
\begin{align*}
     V^{\pi} (h_t)
       & = \mathbb{E}[ \langle g^{\pi}, \phi^H(h_t) \otimes \phi^{\mathcal{O}}(\Tilde{t}_{h}(o)) \rangle \mid h_t, \Tilde{t}_{h}(a) \sim \pi] \\
       & = \mathbb{E}[ \langle g^{\pi}_{t} \otimes g^{\pi}_{t+1}, \Sigma_{\mathcal{O},o \mid \mathcal{A}, a, h_t} (\phi^{\mathcal{A}}(t_{h+1}(a)) \otimes \phi^{a}(a_{t-1}) )  \rangle \mid h_t, a_{t-1}\sim \pi ,t_{h+1}(a) \sim \pi] \\
       & = \mathbb{E}[ \langle \underbrace{g^{\pi}_{t+1} \otimes g^{\pi}_{t}}_{\text{ shifted $W-step$ $\otimes$ $1-step$}},  \underbrace{\phi^{\mathcal{O}}(t_{h+1}(o)) \otimes \phi^{o}(o_{t-1}}_\text{shifted $W-step$ $\otimes$ $1-step$})\mid h_t, a_{t-1}\sim \pi ,t_{h+1}(a) \sim \pi] 
\end{align*}
The final line of the equation gives a dual representation of Eq. \eqref{Equation: extended observation under extended operator}, where $g^{\pi}_{t+1} \otimes g^{\pi}_{t} \in \mathcal{H}_{\mathcal{O}} \otimes \mathcal{H}_{o}$ is the tensor of shifted $W-$step value function $g^{\pi}_{t+1}$ one-step value functional $g^{\pi}_{t}$. Please note that the reward $r$ defined on the latent state is not necessary to be known since all reward information has been embedded into the functional $g$. This property also holds in the risk functions. By levering the information of operators, instead of merely using the histories, the shifted observations will be conditioned to the policy $\underbrace{t_{h+1}(o)}_{\text{see Eq. \eqref{Equation: shifted W-step observation}}} \mapsto \pi(a_t) \in \Delta(\mathcal{A})$, since the one-step optimal action will be constrained by the shifted $W-$step value/risk functions.


\subsection{Bellman Loss} \label{Subsection: Bellman Loss}
\textbf{Proposition 2. (Uniqueness of value/risk functions represented by link functions in RKHS)} When a class of link functions $f: \mathcal{H}_{H} \otimes \mathcal{H}_{\mathcal{A}} \otimes \mathcal{H}_{\mathcal{A}} \rightarrow \mathbb{R}$ is well-defined in RKHS, there exists a unique value/risk function represented by link functions concerning arbitrary policy $\pi \in \Pi$.

The proof of the proposition is direct by combining the properties of forward operators since all those operators exist uniquely in each RKHS due to the characteristic kernels. Followed by the Reisz representation theory in functional analysis \cite{brezis2011functional}, the induced value/risk functions defined in Eq. \eqref{Equation: operator-driven value function} and \eqref{Equation: operator-driven risk function} are all unique once the kernel functions are well defined. Even though, when different characteristic kernel functions parameterize the link functions, the induced functions should remain equivariant, it can be proved from the locally compact group perspective and operators in ergodic theory \cite{drewnik2021reproducing, de2000topics}, detailed proof omitted in this paper.
 
Based on the uniqueness of the value/risk functions represented by link functions, we can define the corresponding Bellman loss based on derived Eq. \eqref{Equation: operator-driven value function} and \eqref{Equation: operator-driven risk function}. Consider two policies $g^{\pi_1}$ and $g^{\pi_2}$, the Bellman loss is denoted as $\text{BL}$, for given arbitrary history $h_{t}$, $\text{BL}_{t}$ is 
\begin{equation}
\begin{split}
        &\text{BL}_t(\pi_{1}, g, \pi_{2}) \defeq \mathbb{E} [ g^{\pi_1}_{t} - g^{\pi_2}_{t} ] \label{Equation: bellman loss}
\end{split}
\end{equation}
If $\text{BL}_t(\pi_{1}, g, \pi_{2}) \equiv 0, \forall h_t \in [H]$, we will say $\pi^{1} \equiv \pi^{2}$.  The Bellman operator $B$ can be defined as 
\begin{equation}
\begin{split}
    B^{\pi} V (h_t)
       & = \mathbb{E} [ \langle g^{\pi}_t, \Sigma_{{o \mid a, h_t}} \phi^{a} (a_{t-1}) \rangle \mid h_t, a_{t-1} \sim \pi] \quad (\text{one-step rollout under $\pi$})\\
       & \qquad + \mathbb{E} [ \mathbb{E}[\langle g^{\pi}_{t+1}, \mathcal{P}_{o_t, a_{t-1}}\Sigma_{\mathcal{O \mid A} , h_{t+1}} \phi^{\mathcal{O}}(t_{h+1}(o)) \rangle \mid t_{h+1}(a) \sim \pi] \mid h_{t}, o_t, a_{t-1} \sim \pi]  
\end{split}
\end{equation}
Therefore, according to Eq. \eqref{Equation: bellman loss}, the updating of the value function based on
\begin{equation}
\begin{split}
    & \text{BL}_t(\pi, g, \pi) = 0 \\ 
    & \Rightarrow \limsup_{n \rightarrow \infty } \lVert \tilde{V}^{\pi, (n)}(h_t) - V^{\pi}(h_t) \rVert = \limsup_{n \rightarrow \infty } \lVert \mathbb{E} [ \tilde{g}^{\pi, (n)}_{t} - g^{\pi}_{t} - g^\pi_{t+1} ] \rVert_{\infty}  \rightarrow 0 
    \label{Equation: TD in PSR}
\end{split}
\end{equation}
Where $\tilde{V}^{(n)}$ is the $n-th$ update. The uniqueness property indicates that the Bellman loss will decay to zero when the two policies are equivalent. The Eq. \eqref{Equation: TD in PSR} can be analogues to the TD-difference algorithm, but the link functions and forward operators represent it, we can see 
then we can see $\limsup_{n \rightarrow \infty }  \lVert \mathbb{E} [ \tilde{g}^{\pi, (n)}_{t} - g^{a_1}_{t} - g^{\pi^*}_{t+1} ] \rVert_{\infty} \rightarrow 0 $. 
It is similar for the risk function $C^{\pi}_{i}(h_t)$ when $m^{\pi}_{i}$ is the risk link function with $\text{BL}_t (\pi, m_{i}, \pi) = 0$. Although it has indicated the unique representation of value/risk functions represented by link functions, proving the convergence of $V^{\pi} (h_t)$ and $C^{\pi}_{i}(h_t)$ relies on the error bound of forward operators, see Eq. \eqref{Equation: TD in PSR}. The accuracy of $g^{a_1}_{t}$ and $g^{\pi^*}_{t+1}$ is controlled by the one-step dynamics, shifted dynamics, and shifted operators. In the following section, we will provide the error bound and sample complexity to estimate such as system.

\section{Main Algorithm and Theoretical Theorem} 
In this part, we first introduce the algorithms and how to estimate the operators, followed by providing the algorithm's error bound and sample complexity. 

\subsection{Estimation of Operators} \label{Section: Estimation of Operators}
\begin{itemize}
    \item Estimation of the forward operator
    \begin{equation}
    \begin{split}
        \Sigma_{\mathcal{O} \mid \mathcal{A}, H}  =  \Sigma_{\mathcal{O},\mathcal{A}, H} [\Sigma_{H \mathcal{A}} \otimes \Sigma_{H \mathcal{A}}^{*}]^{-1}
    \end{split} 
    \end{equation}
    \item Loss function of the forward operator
    \begin{equation}
    \begin{split}
        & \arg \min_{\Tilde{\Sigma}} \frac{1}{N} \sum_{i =1}^{N} \lVert  \Tilde{\Sigma}_{\mathcal{O} \mid \mathcal{A}, H} \times (\phi^{H}(h_i) \otimes \phi^{\mathcal{A}}(t_h(a)_i) ) 
          - \phi^{\mathcal{O}}(t_h(o)_i)\rVert_{L^2}   + \lambda \lVert \tilde{\Sigma}_{\mathcal{O} \mid \mathcal{A}, H}
   \rVert_{HS}
    \end{split}
    \end{equation}
    \item Estimation of the shifted operator
    \begin{equation}
        \mathcal{P}_{o_{t}, a_{t-1}} = \Sigma_{\mathcal{O,A} \mid h_{t}, o_{t}, a_{t-1}} \Sigma_{\mathcal{O,A} \mid h_{t}}^{-1} 
    \end{equation}
    \item The loss function of the shifted operator
    \begin{equation}
    \begin{split}
        & \arg min_{\Tilde{\mathcal{P}}} \frac{1}{N-1} \sum_{i =1}^{N-1} \lVert  \Tilde{\mathcal{P}}_{o_{t}, a_{t-1}} \circ  \Tilde{\Sigma}_{\mathcal{O} \mid \mathcal{A}, H} \times (\phi^{H}(h_i)
         \otimes \phi^{\mathcal{A}}(t_h(a)_i) )  \\
        & \qquad \qquad - \phi^{\mathcal{O}}(t_{h+1}(o)_i)\rVert_{L^2}   + \lambda \lVert \tilde{\mathcal{P}}
   \rVert_{HS} \label{loss function of shifted operator}
   \end{split}
    \end{equation}
    \item Representation of shifted forward operator
    \begin{equation}
        \tilde{\Sigma}_{\mathcal{O}\mid \mathcal{A}, h_{t+1} } = \tilde{\mathcal{P}}_{o_t, a_{t-1}} \circ \tilde{\Sigma}_{\mathcal{O}\mid \mathcal{A}, h_t }
    \end{equation}
    It does not need to construct a new loss function to calculate the shifted forward operator, since it only relies on the shifted operator and forward operator. The estimation of the extended operator is just the same as the estimation of the forward operator. 
\end{itemize}

\subsection{Error Bound and Sample Complexity}
The convergence of $V^{\pi}$ and $C^{\pi}$ relies on the estimation of operators, operators are spanned by the rank-one random matrix. The detailed proof has been listed in Appendix \ref{Appendix: Main Theorem Proof}. Here, only the main theorems are listed in the main text.

\textbf{Theorem 1. (Error bound of the forward operator)} Under the regularity assumption, the error bound of the forward operator $\Sigma_{\mathcal{O \mid A,} H}$ has the empirical estimation as:
\begin{equation}
\begin{split}
    \tilde{\Sigma}_{\mathcal{O \mid A,} H} & = \tilde{\Sigma}_{\mathcal{O,A,}H} (\tilde{\Sigma}_{HH} \otimes  \tilde{\Sigma}_{\mathcal{AA}} + \lambda I )^{-1} \\
    & = \tilde{\Sigma}_{\mathcal{O,A,}H} (\tilde{\Sigma}_{H,\mathcal{A}} ^2  + \lambda I)^{-1} 
    \label{Error bound of forward operator}
\end{split}
\end{equation}
we have the probability with at least $1 - \delta $, $\forall \delta \in (0,1)$ satisfying the  
\begin{equation}
    \mathbb{P}(\lVert \Sigma_{\mathcal{O \mid A,} H} -  \tilde{\Sigma}_{\mathcal{O \mid A,} H} \rVert \geq c) \leq 1 - \delta 
\end{equation}
where $\lambda \rightarrow 0$ and $ \phi_{k}^{\mathcal{O}} \in \mathbb{C}^{n_1} $, $ \phi_{k}^{\mathcal{A}} \in \mathbb{C}^{n_2} $ and $ \phi_{k}^{H} \in \mathbb{C}^{n_3}$ for all $k \in K$, 
\begin{align*}
    & \tilde{\Sigma}_{H, \mathcal{A}}^2 =  \frac{1}{\lvert K \rvert} \sum_{k \in K} [\phi^{H}_k \otimes \phi^{\mathcal{A}}_k] [\phi^{H}_k \otimes \phi^{\mathcal{A}}_k]^* \\ 
    & c \defeq  \frac{2 \log( (n_1 + n_3n_2)/\delta) \Bar{c}_7}{3 \rho_{min}( \tilde{\Sigma}_{\mathcal{H}, \mathcal{A}}^2 )}  + \frac{\sqrt{2 \log ( (n_1 + n_3n_2)/\delta) \overline{Var(\Sigma_{\mathcal{O,A}, H})}}}{2 \rho_{min}( \tilde{\Sigma}_{\mathcal{H}, \mathcal{A}}^2 )} \\ 
    & + 
      \frac{\sqrt{\rho_{max}(\Sigma_{\mathcal{OO}})} + \sqrt{\rho_{max}(\Sigma_{\mathcal{OO}}) } \epsilon_1}{ \sqrt{ \rho_{min}( \tilde{\Sigma}_{\mathcal{H}, \mathcal{A}}^2 )}} \cdot \frac{\epsilon_2 + \lambda}{1 + \epsilon_2 + \lambda }  \\
    & \Bar{c}_7 = \frac{\max_{i,j,z} n_1 n_2 n_3 \sqrt{20 \log(2 \Bar{c}_{i,j,z}/\delta)}}{\sqrt{\lvert K \rvert}} + \frac{\sqrt{200 C n_1 n_2 n_3 \max_{i,j,z} \Bar{c}_{i,j,z}} (\log(2/\delta))^{\frac{3}{4}} }{\lvert K \rvert^{\frac{1}{4}}}  \\
    & \overline{Var(\Sigma_{\mathcal{O, A,} H})} =  \frac{\max_{i,j,z} n_1 n_2 n_3 \sqrt{20 \log(2 \Bar{c}_{i,j,z}/\delta)}}{\sqrt{\lvert K \rvert}}  \\
    & \epsilon_1 \lesssim \frac{ \log (n_1/\delta)}{ \lvert K \rvert \rho_{\max}(\Sigma_{\mathcal{OO}})} \\
    & \epsilon_2 \lesssim \frac{ \log (n_2/\delta)}{  \lvert K \rvert \rho_{\max}(\Sigma_{\mathcal{AA}})}
\end{align*}
The proof is built on the random matrix theory, and the lemmas using the proof have been listed with details in Appendix \ref{Appendix: Main Theorem Proof}.

\textbf{Theorem 2. (Error bound of the shifted forward operator)} Under the same conditions in Theorem 1, the error bound of the shifted forward operator is 
\begin{align*}
    & \mathbb{P}(\lVert  \tilde{\mathcal{P}}_{a, o} \tilde{\Sigma}_{\mathcal{O \mid A}, H_{t}} - \Sigma_{\mathcal{O \mid A}, H_{t+1}} \rVert > c) \leq 1- \delta \\
    & c = \frac{2 \log( (n_1 + n_3n_2)/\delta) \Bar{c}_7}{3 \rho_{min}( \tilde{\Sigma}_{\mathcal{H}, \mathcal{A}}^2 )}  + \frac{\sqrt{2 \log ( (n_1 + n_3n_2)/\delta) \overline{Var(\Sigma_{\mathcal{O,A}, H})}}}{2 \rho_{min}( \tilde{\Sigma}_{\mathcal{H}, \mathcal{A}}^2 )} \\
    & + \sqrt{n_1^2 \frac{\max_{i,j} \lVert c_{i,j} \rVert}{\lvert K \rvert} \log(2 n_1^2/\delta )} \lVert \Sigma_{\mathcal{O \mid A}, H_{t}}  \rVert 
\end{align*}
with probability at least $1 - \delta$, the other symbol definitions are the same with theorem 1. 

\textbf{Theorem 3. (Error bound of the value/risk functions)} For arbitrary value function or risk function parameterized by any policy $\pi \in \Pi$, the error bound $\lVert \Tilde{V}^{\pi} -V^{\pi} \rVert_{\infty}$ scaling polynomially. 

The proof of the result can be directly obtained from Theorem 1 and 2 and Eq. \eqref{Equation: operator-driven value function}, \eqref{Equation: operator-driven risk function}, details are omitted here. 

\textbf{Theorem 4. ($\epsilon$-suboptimal policy with polynomial sample complexity)} When Theorem 1, 2, and 3 holds, the safe policy can converge to a $\epsilon-$suboptimal policy with a polynomial sample complexity. More specifically, it can guarantee the probability with at least $1 - \delta $, $\forall \delta \in (0,1)$ satisfying the condition as
\begin{align*}
    \tilde{V}^{\tilde{\pi}}(h_0) \geq V^{\pi^*}(h_0) - \epsilon \quad \text{and} \quad   \tilde{C}_{i}^{\tilde{\pi}}(h_0) \leq \Bar{C}_{i} + \epsilon, \ \forall h_0 \in [H], \ i \in [N] 
\end{align*}
When the sample size $\lvert K \rvert > \tilde{\mathcal{O}}((n_1n_2)^2 n_3 \Bar{c}_{i,j} \log(2/\delta)^3)$, the error $\epsilon$ can be kept as arbitrarily small with at least $1-\delta$ probablity. \\
\newline 
\textit{Proof. } The estimation of $\tilde{V}^\pi, \tilde{C}^{\pi}$ are all parameterized by $\theta \in \Theta$ and $\pi \in \Pi$. More specifically, $\theta$ is the parameter of various operators and link functional, since both  $V^\pi, C^{\pi}$ are determined by the forward operator and shifted operator and link functional variables $g$ (see Eq. \eqref{Equation: operator-driven value function}). In this situation, we can use $J(\theta, \pi, \eta), \forall \theta \in \Theta; \pi \in \Pi$ to denote the KKT condition of safe PSR problem as 
\begin{align*}
    J(\theta, \pi, \eta) = \tilde{V}^{\tilde{\pi}}(h_0) - \sum_{i} \eta_i (\tilde{C}^{\tilde{\pi}}(h_0) - \Bar{C}_i)^+
\end{align*}
where $\eta = (\eta_{1}, \cdots, \eta_{N})$ is the dual variable,  the solution of the optimization problem can be represented as 
\begin{align*}
    (\tilde{\pi}, \tilde{\theta}, \tilde{\eta}) = \arg \max_{\pi \in \Pi} \text{arg}\min_{\eta, \theta \in \Theta} J(\theta, \pi, \eta)
\end{align*}
By the definition of saddle point, it can be derived that
\begin{align*}
     J(\tilde{\theta}, \pi^{*}, \eta^{*}) \leq J(\tilde{\theta}, \tilde{\pi}, \tilde{\eta}) \leq J(\theta^{*}, \tilde{\pi}, \tilde{\eta})
\end{align*}
\begin{align*}
    \Rightarrow \quad & J(\theta^*, \pi^*, \eta^*) - J(\theta^{*}, \tilde{\pi}, \tilde{\eta}) \\
    & \leq J(\theta^*, \pi^*, \eta^*) -  J(\tilde{\theta}, \pi^{*}, \eta^{*}) 
\end{align*}
To simplify the notions, we denote the $\underline{B}^*  \tilde{V}(h_t) =  \max_{\pi \in \Pi} \min_{\eta, \theta \in \Theta} J(\theta, \pi, \eta)$. Under this definition, the target of the problem becomes to obtain the contraction of the $\lVert \underline{B}^* V - \underline{B}^*  \tilde{V} \rVert$. 
\begin{itemize}
    \item \textbf{Case 1.} If the $[\tilde{C}_{i}^{\pi} (h_t) -\Bar{C}_{i}] > 0 , \forall i \in [N]$ for all action $a_t \in \mathcal{A}$, the $\underline{B}^*  \tilde{V}(h_t) \rightarrow -\infty$, which means the certain risk in the future $(W+1)$ steps. 
    \item  \textbf{Case 2.} If there exist $[\tilde{C}_{t}^{\pi} (h_t)  -  \Bar{C}_{i}] \leq 0, \forall i \in [N]$ for some action $a_t \in \mathcal{A}$. We can assert the contraction of $\lVert \underline{B}^* \tilde{V} - \underline{B}^* V \rVert_{\infty} < \epsilon$ ($\epsilon$ is arbitrary small value) with a polynomial sample complexity, proving it needs a lemma, the details are listed as below. 
\end{itemize}
Consider two arbitrary functions $f$ and $g$, we have
\begin{equation}
    \lvert \max_{x} f(x) - \max_{x} g(x) \rvert  \leq \max_{x} \lvert f(x) - g(x) \rvert 
    \label{Equation: lemma max}
\end{equation}
To see this, we suppose $\max_{x} f(x) > \max_{x} g(x)$ (with respect the symmetric case) and let $x^* \in_x f(x)$, then
\begin{equation}
\begin{split}
    & \lvert \max_x f(x) - \max_{x} g(x) \lvert = f(x^*) - \max_x g(x) \\
    & \leq f(x^*) - g(x^*) \leq \max_{x} \lvert f(x) - g(x) \rvert  
\end{split} 
\end{equation}
Similarly, the symmetric case of Eq. \eqref{Equation: lemma max} can be indicated such that
\begin{equation}
    \lvert \min_{x} f(x) - \min_{x} g(x) \rvert \leq \max_{x} \lvert f(x) - g(x) \lvert
    \label{Equation: min lemma}
\end{equation}
Thus in our case, we have 
\begin{equation}
\begin{split}
    & \lVert \underline{B}^* \tilde{V} - \underline{B}^* V \rVert_{\infty} = \sup_{h_t, a_{t-1}} \lVert \underline{B}^* \tilde{V} - \underline{B}^* V \rVert \\
    & \leq \sup_{h_t} \bigg\lVert \mathbb{E} [ \langle \tilde{g}^{\pi}_t, \tilde{\Sigma}_{{o \mid a, h_t}} \phi^{a} (a_{t-1}) \rangle \mid h_t, a_{t-1} \sim \tilde{\pi} ] \\
       & \qquad + \mathbb{E} [ \mathbb{E}[\langle \tilde{g}^{\pi}_{t+1}, \tilde{\mathcal{P}}_{o_t, a_{t-1}} \tilde{\Sigma}_{\mathcal{O \mid A} , h_{t}} \phi^{\mathcal{O}}(t_{h+1}(o)) \rangle \mid t_{h+1}(a) \sim \tilde{\pi}]\mid h_{t}, o_t, a_{t-1} \sim \tilde{\pi}] \\
       & - \mathbb{E} [ \langle g^{\pi}_t, \Sigma_{{o \mid a, h_t}} \phi^{a} (a_{t-1}) \rangle \mid h_t, a_{t-1} \sim \pi^{*} ] \\
       & \qquad - \mathbb{E} [ \mathbb{E}[\langle g^{\pi}_{t+1}, \mathcal{P}_{o_t, a_{t-1}}\Sigma_{\mathcal{O \mid A} , h_{t}} \phi^{\mathcal{O}}(t_{h+1}(o)) \rangle \mid t_{h+1}(a) \sim \pi]\mid h_{t}, o_t, a_{t-1} \sim \pi^{*}] \bigg\rVert \\ 
    & \leq \mathcal{O}(\epsilon) \\
    & \text{it hold when } \lvert K \rvert\geq \tilde{\mathcal{O}}((n_1n_2)^2 n_3 \Bar{c}_{i,j} \log(2/\delta)^3) \qquad (\text{see Lemma in Appendix \ref{Appendix: Main Theorem Proof}})
\end{split}
\end{equation}
The second line of the equation holds because the available action set satisfies the safety constraint is smaller than the whole action set, the Eq. \eqref{Equation: lemma max} and \eqref{Equation: min lemma} indicate that:
\begin{equation}
\begin{split}
    & \lVert \underline{B}^* \tilde{V} - \underline{B}^* V \rVert_{\infty} \leq \lVert B^* \tilde{V} - B^* V \rVert_{\infty}  \\ 
    & =  \sup_{h_t} \lVert B^* \tilde{V}(h_t)  - B^* V(h_t) \rVert    
\end{split}
\end{equation}
The last line equation is due to Theorem 3, the error bounded can be easy to derive by using triangular inequalities and decomposing the error by parts. Due to Theorem 1 and 2, the error bound shrinks polynomially concerning the data size $\lvert K \rvert$. Also, it can be indicated that error bound $\lVert \underline{B}^* V - \underline{B}^*  \tilde{V} \rVert$ is weaker than the result  $BL(\tilde{\pi}, g, \pi^*)$ Eq. \eqref{Equation: bellman loss}. 

\subsection{Details of Algorithms}
The algorithm should be developed in the following steps to solve the safe PSR problem practically. 
\begin{itemize}
    \item Pre-train the operators of the PSR, ensuring the sufficient accuracy of operators based on the loss function in Section \ref{Section: Estimation of Operators}; 
    \item Follow the Lagrangian relaxation procedure to optimize the corresponding policy as
    \begin{equation}
    J(\pi, \eta) = V^{\pi} - \sum_{i} \eta^i (C_i - \Bar{C}_i)^+
    \end{equation}
    where $\eta = [\eta^1, \eta^2, \cdots, \eta^N]$ are dual variables, the policy is updating as a Gaussian process $\pi(a_t) \sim \mathcal{N}(\mu(t_{h+1}(o)), \sigma)$ (see the description in Section 4.3)\footnote{$\pi(t_{h+1}(o)) \in \Delta(\mathcal{A})$ leverages the information of shifted forward operator, the future shifted observation will depend on $ 
 \phi^{\mathcal{O}}(t_{h_{t+1}}(o))=\mathbb{E}[\Sigma_{\mathcal{O} \mid \mathbb{A}, h_{t+1}} \phi(t_{h+1}(a)) \mid  t_{h+1}(a) \sim \pi]$, the update of $\pi$ can measure the optimal action $a_{t-1}$, since we have the relationship that $\Sigma_{\mathcal{O} \mid \mathbb{A}, h_{t+1}} = \mathcal{P}_{o_t, a_{t-1}} \circ \Sigma_{\mathcal{O} \mid \mathbb{A}, h_{t}}$, the each step policy can be regarded as an instrumental variable to measure the shifted dynamics, then the feedback infomration to help measure the optimal action, the recursive updation of policy and operators, will drive the policy to a fixed point, it has been indicated in Bellman loss. }, the noise $\sigma$ same setting as \cite{schulman2015trust}. When iteratively updating the $\pi_{\theta}$, it can be calculated as 
    \begin{equation}
        \theta_{k+1} \leftarrow \theta_{k} + \alpha_k\restr{(\nabla_{\theta}(J(\pi, \eta)))}{\theta = \theta_k}
    \end{equation}
    where $\alpha_k$ is the step size, the adaptive step size is similar to the \cite{achiam2017constrained}. After the iteration of $\pi = \pi_{k+1}$, fix the policy parameters, update the dual variable as 
    \begin{equation}
        \eta_{k+1} = [\eta_k + \beta_k(C^{\pi_{k+1}} - \Bar{C})]^+
    \end{equation}
    where $\beta_k$ is the step size of dual variables. 
    \item After the rollout of one whole episode, the functional $g$ and $m$ should be updated as 
    \begin{equation}
        \arg \min_{g} \lVert g^{\pi} (\tilde{t}_h(o)) - \sum r_i \rVert
    \end{equation}
    and 
    \begin{equation}
        \arg \min_{m} \lVert m^{\pi} (\tilde{t}_h(o)) - \sum c_i \rVert
    \end{equation}
    where $r_i$ and $c_i$ are the environment's observed reward and risk information. 
\end{itemize}

\section{Conclusion}
We introduce a novel approach, Safe Kernel RL, which combines concepts from Predictive State Representations and Tensor Reproducing Kernel Hilbert Spaces (RKHS). Unlike conventional methods, our approach doesn't require estimating the probability space of the observation and latent spaces. To implement this method, we propose five crucial operators that describe the relationships between forward observations, histories, and policy information. By leveraging these well-defined operators, the value/risk functions defined on finite-horizon latent states can be transformed into value/risk functionals defined on the features of finite-horizon observations using link functions. Once the representation of value/risk functions is established, we can achieve $\epsilon$-sub-optimal solutions with polynomial sample complexity. In contrast to constrained policy optimization methods like \cite{achiam2017constrained} and proximal policy optimization \cite{schulman2017proximal}, we update the policy based on future conditional observations, as our operators can measure future shifted observations under the given policy.

Looking ahead to future research directions, the operators defined in this paper can be linked to current cutting-edge topics, such as neural operators in RKHS \cite{kovachki2021neural}, which provide a more expressive and efficient way to represent stochastic dynamical systems. Additionally, Kernel PSRs can offer greater expressiveness compared to classic Model Predictive Control (MPC) methods due to their non-parametric nature, and constrained PSRs can be reformulated in MPC to achieve more generalized safe control.

\bigskip

\clearpage
\bibliography{ref}
\bibliographystyle{ieeetr}

\clearpage

\appendix

\onecolumn
\section{Important Definitions and Properties of PSR} \label{Appendix: Some Important Definitions and Properties of PSRs}

\textbf{Definition. (Undercompleness assumption) \cite{jin2020sample}} Let $\mathbb{O}: [H] \times L^1(\mathcal{S}) \rightarrow L^1(\mathcal{O})$  and the observable operator defined as defined on $f \in L^1(\mathcal{S})$, then there is an corresponding operator $\mathbb{Q}: [H] \times L^1(\mathcal{O}) \rightarrow L^1(\mathcal{S})$ satisfying
\begin{equation}
    (\mathbb{Q} \circ \mathbb{O}) f = f     \label{Equation: Undercompleness assumption 1}
\end{equation}
where $\mathbb{Q}$ is $\gamma-$regularity. Meanwhile, the operator $\mathbb{Q}$ also satisfying 
\begin{equation}
    (\mathbb{Q} \circ g) (h, \mathcal{O}) = \int_{o \in \mathcal{O}} \xi_{h}(s, o) g(o) do  \label{Equation: Undercompleness assumption 2}
\end{equation}
where $g \in L^{1}(\mathcal{O})$ and $\xi: [H] \times \mathcal{S} \times \mathcal{O} \rightarrow [0,1]$.

The assumption can be originally discovered in \cite{jin2020sample} under the tabular POMDP with undercompleteness settings. By observing the Eq. \eqref{Equation: Undercompleness assumption 1}, it is not hard to see the $\mathbb{Q}$ can be regarded as the left inverse of $\mathbb{O}$. Back to the definition of $\mathbb{O}$, it can be analogous to a Bayesian filter to measure the probability transition from latent state space to observation space.  The left inverse of $\mathbb{O}$ guarantees the observation information can be pulled back to the state space without losing any information. Under the undercompleteness assumption, it allows us to directly estimate any functional or operators on observation space without inferring the corresponding latent distributions. Since it has indicated the operator $\mathbb{Q}$ is the left inverse of $\mathbb{O}$, we will denote the $\mathbb{Q}$ as $\mathbb{O}^{\dag}$ in other parts. 

\textbf{Extension of the undercompleteness assumption in RKHS.} Consider $\Sigma_{\mathcal{O \mid S}, H}:  \mathcal{H}_{H} \otimes \mathcal{H}_{\mathcal{U}_{\mathcal{S}}} \rightarrow \mathcal{H}_{\mathcal{U}_{\mathcal{o}}}$ as the embedding observable operator, then there exist an operator such that $\Sigma_{\mathcal{S \mid O }, H}: \mathcal{H}_{H} \otimes \mathcal{H}_{\mathcal{U}_{\mathcal{O}}} \rightarrow \mathcal{H}_{\mathcal{U}_{\mathcal{S}}}$ satisfying 
\begin{equation}
    \Sigma_{\mathcal{S \mid O }, H} \Sigma_{\mathcal{O \mid S}, H} = \mathbb{I}
\end{equation}
It is a natural result extended from the probabilistic version of the undercompleteness assumption. Here, $ \Sigma_{\mathcal{S \mid O }, H}$ is the left inverse of $\Sigma_{\mathcal{O \mid S}, H}$. 

\textbf{Definition. (Tensor algebra over ring) \cite{matsumura1970commutative}} Let R be a commutative ring and let $A$ and $B$ be R-algebras. Since $A$ and $B$ may both be regarded as R-modules, their tensor product can be written as 
\begin{align*}
    A \otimes_{R} B
\end{align*}
is also an R-module. The tensor product can be given the structure of a ring by defining the product on elements of the form $a \otimes b$ as
\begin{align*}
    (a_1 \otimes b_1) \circ (a_2 \otimes b_2) = (a_1 a_2) \otimes(b_1 b_2)
\end{align*}
and then extending by linearity to all of $A \otimes B$. Furthermore, the property also holds for multiple R-algebras. 

\section{Basics of RKHS} \label{Appendix: Basics of RKHS}

\textbf{Characterization of two important RKHSs}. Without loss of generality, assume $(\mathcal{X}, \mathcal{B}_{\mathcal{X}})$ and $(\mathcal{Y}, \mathcal{B}_{\mathcal{Y}})$ are two measurable spaces, $\mathcal{B}_{\mathcal{X}}, \mathcal{B}_{\mathcal{Y}}$ are the Borel $\sigma-$sets. Assume $(H_{\mathcal{X}}, k_{\mathcal{X}})$ and $(H_{\mathcal{Y}}, k_{\mathcal{Y}})$ are two RKHSs of topological space $\mathcal{X}$ and $\mathcal{Y}$, where $k_{\mathcal{X}}$ and $k_{\mathcal{Y}}$ are two positive definite kernels. Consider a random variable $(X, Y): \Omega  \rightarrow  \mathcal{X \times Y}$ with the joint distribution $\mathbb{P}_{XY}$. Then, we can connect to the sequential problem that $(X, Y)$ forms a cylinder set $(X,Y) \in \mathbb{K}^n$ ($\mathbb{K}$ is a field, either $\mathbb{R}$ or $\mathbb{C}$). The two spaces are essential for constructing the measure of the conditional expectation of a dynamical system. For example, if we treat $X$ as the \textit{history} and $Y$ as the \textit{test}. The $\mathbb{P}_{X}$ and $\mathbb{P}_{Y}$ denoted the marginal distribution histories and tests, respectively. Subsequently,  essential definitions and properties of RKHSs should be introduced. 
\begin{enumerate}
    \item \textit{Reproducing property:} The reproducing property allows decomposing functions in a group of basis such that $f = \sum_{i \in I} w_i k(\cdot, x_i)$, then $f(x)$ can be represented as an integral form as
    \begin{equation}
        f(x) = \langle f, k(\cdot, x) \rangle_{\mathcal{H}_{\mathcal{X}}} = \langle \sum_{i \in I} w_i k(\cdot, x_i),  k(\cdot, x) \rangle_{\mathcal{H}_{\mathcal{X}}}
    \end{equation}
    where $w_i$ is the weight. 
    \item \textit{Mean map:} The mean functional $m_{X}$ and $m_{Y}$ on $H_{\mathcal{X}}$ and $H_{\mathcal{Y}}$ satisfy that
    
    \begin{equation}
    \begin{split}
               & \mathbb{E}_{X}(f(X)) = \langle m_{X}, f \rangle_{\mathcal{H}_{\mathcal{X}}}; \\ 
        &  \mathbb{E}_{Y}(g(Y)) = \langle m_{Y}, g \rangle_{\mathcal{H}_{\mathcal{Y}}} 
        \label{mean embedding}
    \end{split}
    \end{equation}
    \item \textit{Kernel:} the positive definite kernels are bounded as:
    \begin{center}
        $\mathbb{E} [k_{\mathcal{X}}(x,x)] < \infty$ and $\mathbb{E} [k_{\mathcal{Y}}(y,y)] <\infty$
    \end{center}
    where 
    \begin{center}
        $k_{X} (x, x) = \langle k_{X}(\cdot, x), k_{X}(\cdot, x) \rangle_{\mathcal{H}_\mathcal{X}}$\\ 
        $= \langle \phi^{X}(x), \phi^{X}(x) \rangle_{\mathcal{H}_{X}}$
        , for $x \in X$
    \end{center}
    $\phi^{X}$ is a feature map such that $x \mapsto \phi^X(x)$. $\overline{span(\{ \phi^X(x_{i})\})}$ is the complete feature space as well as the induced Hilbert space. 
    \item \textit{Uncentered covariance operator:} By Kernel Baye's Rule (KBR) \cite{fukumizu2011kernel}, the kernel mean of the joint probability on $\mathcal{H}_{\mathcal{X}}$ and $\mathcal{H}_{\mathcal{Y}}$ requires the cross-variance operator:
    \begin{equation}
        \Sigma_{XY} \defeq \mathbb{E} [\phi^{X}(x) \otimes \phi^{Y}(y)]
        \label{Equation: KBR_Cov_XY}
    \end{equation}
    The joint expectation of $\mathbb{P}(X, Y)$ can be represented in the RKHS with an adjoint operator:
    \begin{equation}
        \begin{split}
             \langle f, \Sigma_{XY}g \rangle 
             &= \mathbb{E}_{XY}[\langle f, k_{X}(\cdot, x) \rangle_{\mathcal{H}_{\mathcal{X}}} \langle g, k_{Y}(\cdot, y) \rangle_{\mathcal{H}_{{\mathcal{Y}}}}] \\ 
             &= \mathbb{E}_{XY}[f(x)g(y)]  
        \end{split}
        \label{uncentred covariance}
    \end{equation}
    The first line in Eq.\eqref{uncentred covariance} reveals the adjoint property in Hilbert space of $\Sigma_{XY}: \mathcal{H}_{\mathcal{Y}} \rightarrow \mathcal{H}_{\mathcal{X}}$. Then, the covariance operator can be presented as self-adjoint $\Sigma_{XX} \defeq \mathbb{E} [\phi^{X}(x) \otimes \phi^{X}(x) ] $.
    For practical calculation of $\Sigma_{XY}$, the joint probability, $\mathbb{P}_{XY}$ on the $(\mathcal{X} \times \mathcal{Y}, \mathcal{B}_{\mathcal{X}} \otimes \mathcal{B}_{\mathcal{Y}}, \mathbb{P}_{XY})$, can be estimated via sampling a i.i.d dataset such that $\mathcal{D} \defeq (x_{i}, y_{i})_{i=1}^{N}$. By projecting each point of data to RKHS, the corresponding cross-variance operator can be represented by:
    \begin{equation}
        \hat{\Sigma}_{XX} = \frac{1}{N} \Phi_{X} \Phi_{X}^{*}, \quad \hat{\Sigma}_{XY} = \frac{1}{N} \Phi_{X} \Phi_{Y}^{*},
    \end{equation}
    where $\Phi_{X} = (\phi^{X}(x_{1}), \phi^{X}(x_{2}), \cdots, \phi^{X}(x_{N}))$ and  
    $\Phi_{Y} = (\phi^{Y}(y_{1}), \phi^{Y}(y_{2}), \cdots, \phi^{Y}(y_{N}))$.
    \item \textit{Embedding theorem:} the conditional operator $\Sigma_{X \mid Y}$ can be represented as \cite{song2009hilbert}:
    \begin{equation}
        \Sigma_{X \mid Y} = \Sigma_{XY} \Sigma_{YY}^{\dag}. \label{Equation: conditional operator}
    \end{equation}
    If we use an inner product format to embed $Y$ into $H_{\mathcal{X}}$, we can obtain the following result by applying Eq.\eqref{Equation: KBR_Cov_XY} and Eq.\eqref{uncentred covariance}:
    \begin{equation}
    \begin{split}
       \mathbb{E}_{X\mid Y}[f(x) \mid Y=y ] &= \mathbb{E}_{X\mid Y}[\langle f, \Sigma_{X \mid Y} k_{Y}(\cdot , y) \rangle_{\mathcal{H}_{\mathcal{X}}}] \\ 
        &= \mathbb{E}_{X\mid Y}[\langle f, \Sigma_{X \mid Y} \phi^{Y}(y) \rangle_{\mathcal{H}_{\mathcal{X}}}], 
    \end{split}
    \end{equation}
    for the conditional operator $ \Sigma_{X \mid Y}: \mathcal{H}_{\mathcal{Y}} \rightarrow \mathcal{H}_{\mathcal{X}}$. The existence of $ \Sigma_{X \mid Y}$ allows us to change the measurement from $\mathcal{H}_{\mathcal{Y}}$ to $\mathcal{H}_{\mathcal{X}}$. 
    It should be noted that the measure in $(\restr{\mathcal{X}}{\mathcal{Y}}, \restr{\mathcal{B}_{\mathcal{X}}}{\mathcal{Y}}, \restr{\mathbb{P_{\mathcal{X}}}}{\mathcal{Y}})$ is absolutely continuous to another measure in $(\mathcal{Y}, \mathcal{B}_{\mathcal{Y}}, \mathbb{P}_{\mathcal{Y}})$ ($\restr{\mathbb{P_{\mathcal{X}}}}{\mathcal{Y}} \ll \mathbb{P}_{\mathcal{Y}}$). If the absolutely continuous probability measure fails, the stochastic dynamical system becomes difficult to predict. Here, the original probability space should be separable to guarantee weak convergence and no pathological areas in the case. Similar to the calculation in the covariance operator, the condition variance can be calculated from: 
    \begin{equation}
        \hat{\Sigma}_{X\mid Y} = \frac{1}{N} \Phi_{X} (\frac{1}{N}   \Phi_{Y})^{\dag}.
    \end{equation}
    To get the regularized pseudo-inverse $\Phi_{Y}^{\dag}$, the practical calculation can become:
    \begin{equation}
        \Phi_{Y}^{\dag} = \Phi_{Y}^{*}(\Phi_{Y}\Phi_{Y}^{*} + \lambda I)^{-1}.
    \end{equation}
    Thus the corresponding conditional expectation $\Sigma_{X \mid Y}$ can be calculated from: 
    \begin{equation}
        \hat{\Sigma}_{X \mid Y} =  \Phi_{X} (\Phi_{Y}\Phi_{Y}^{*} + \lambda I)^{-1} \Phi_{Y}^{*} \label{conditional expectation estimation}
    \end{equation}
    The RKHS has included wide variants of models such as the Gaussian Process (GP), so our model will be more generalized than the GP-based method \cite{williams2006gaussian}. 
    
    \vspace{12pt}
    
    \textit{Remark.} We have discussed the basics of RKHS and related properties. Some examples can be partially observable dynamical systems. The optimal estimation problem is usually expressed as a pair $(\theta, \xi)$ \cite{shiryaev2016probability} (true states and observation pairs), where $\theta, \xi$ are assumed Gaussian The problem is to derive $\theta$ from $\xi$. We use the $\Sigma_{\theta\theta}, \Sigma_{\xi\xi}$ and $\Sigma_{\theta\xi}$ to denote the uncentered covariance of $\theta\theta, \xi\xi$ and $\theta\xi$. Then, the optimal estimator under the partial observation $\xi$:
    \begin{equation}
        \mathbb{E} (\theta \mid \xi) = \Sigma_{\theta\xi}\Sigma_{\xi\xi}^{-1}\xi \label{LQG 1}
    \end{equation}
    Correspondingly, the minimization of covariance is
    \begin{equation}
        \epsilon = \Sigma_{\theta\theta} - \Sigma_{\theta\xi}\Sigma_{\xi\xi}^{-1}\Sigma_{\theta\xi}^*  \label{LQG 2}
    \end{equation}
    The second equation can be derived from:
    \begin{equation}
    \begin{split}
        \Sigma_{\theta\theta \mid \xi} &= \mathbb{E}[(\theta-\mathbb{E}(\theta \mid \xi))(\theta-\mathbb{E}(\theta \mid \xi))^*] \\
        &= \Sigma_{\theta\theta} + \Sigma_{\theta\xi}\Sigma_{\xi\xi}^{-1}\Sigma_{\xi\xi}\Sigma_{\xi\xi}^{-1}\Sigma_{\theta\xi}^* - 2 \Sigma_{\theta\xi}\Sigma_{\xi\xi}^{-1}\Sigma_{\xi\xi}\Sigma_{\xi\xi}^{-1}\Sigma_{\theta\xi}^* \\
        &= \Sigma_{\theta\theta} - \Sigma_{\theta\xi}\Sigma_{\xi\xi}^{-1}\Sigma_{\theta\xi}^*
         \label{LQG 3}
    \end{split}
    \end{equation}
    The first equation is an explicit solution of Eq. (\eqref{conditional expectation estimation}) We proved from this solution consistency of in 2nd-order, the minimized covariance. This simple example revealed that the partially observable Gaussian system can be a sub-class of RKHS.    
\end{enumerate}

\vspace{12 pt}

\subsection{Proof of Lemma 1} \label{Appendix: proof of lemma 1}
\textit{Proof.} The three properties
\begin{itemize}
    \item Here, we denote the feature map of $Z$ as $z \mapsto \phi^{Z}(z)$. According to the definition of the uncentered covariance operator in Eq. \eqref{Equation: KBR_Cov_XY}, we can see the joint distribution on $H_{\mathcal{X}}$, $H_{\mathcal{Y}}$ and $H_{\mathcal{Y}}$ as a tensor form, due to the induced product Borel set is $\mathcal{B}_{\mathcal{X, Y, Z}} = \mathcal{B}_{\mathcal{X}} \otimes \mathcal{B}_{\mathcal{Y}} \otimes \mathcal{B}_{\mathcal{Z}}$, then we can obtain $\Sigma_{X,Y,Z}$ as
    \begin{equation}
        \Sigma_{X,Y,Z} = \mathbb{E}[\phi^{X}(x) \otimes \phi^{Y}(y) \otimes \phi^{Z}(z)] \label{Equation: cross-variance three mode}        
    \end{equation}
    \item By the embedding theorem in Eq. \eqref{Equation: conditional operator}, we have a similar representation by combining Eq. \eqref{Equation: cross-variance three mode}, we have 
    \begin{equation}
        \Sigma_{XY\mid Z} = \Sigma_{X,Y,Z} \Sigma_{ZZ}^{-1} \label{Equation: conditional operator 1}        
    \end{equation}
    The conditional operator can be regarded as a linear operator, such that $\Sigma_{XY \mid Z} \in \mathcal{L}(H_{\mathcal{Z}}, H_\mathcal{X} \times H_\mathcal{Y})$. 
    \item Recursively using the properties in Eq. \eqref{Equation: conditional operator}, we have 
    \begin{equation}
    \begin{split}
        \Sigma_{X\mid YZ} & = \Sigma_{XY\mid Z}  \Sigma_{YY\mid Z}^{-1}  \\
        & = \Sigma_{X,Y,Z} [\Sigma_{ZZ} \otimes \Sigma_{YY}]^{\dag} \ \quad \text{(tensor product on RKHS see Appendix \ref{Appendix: Some Important Definitions and Properties of PSRs} and \cite{jakobsen1979tensor})} \\
        & = \Sigma_{X,Y,Z} [\Sigma_{ZY} \otimes \Sigma_{ZY}^{*}] \label{Equation: conditional operator 2}        
    \end{split}
    \end{equation}
    where $\otimes$ is column-wise Kronecker product in the practical calculation. 
\end{itemize}

\vspace{12 pt}

Recall  KBR in Eq. \eqref{Equation: KBR_Cov_XY}, considering a third element $Z$ with  $\mathbb{P}_{X,Y \mid z}$, the conditional expectation under the conditional operator in Eq. \eqref{Equation: conditional operator 1} becomes:
    \begin{equation}
    \begin{split}
        \Sigma_{XY \mid z} &= \mathbb{E}[\phi^X(x) \otimes \phi^Y(y) \mid Z = z] \\
        &= \Sigma_{XYZ} \Sigma_{ZZ}^{-1} \phi^{Z}(z) \in H_{\mathcal{X}} \otimes H_{\mathcal{Y}}  \label{condition on xy|z}
    \end{split}
    \end{equation}    Furthermore, by leveraging $ \Sigma_{X \mid Y, z}$, the conditional mean of $X$ given further knowledge of $Y$ can be represented as:
    \begin{equation}
    \begin{split}
        \mathbb{E}(X \mid Y = y, z ) &= \langle \cdot, \Sigma_{X\mid Y,z} k_Y(\cdot, y) \rangle_{\mathcal{H}_{X}} \\
        &= \langle \cdot, \Sigma_{X\mid Y,z} \phi^{Y}(y) \rangle_{\mathcal{H}_{X}}  \\
        &= \Sigma_{XY \mid z} \Sigma_{YY\mid z}^{-1} \phi^{Y}(y) \label{conditional covariance 1} \\
        &= \Sigma_{X \mid y,z} \in H_\mathcal{X}
    \end{split}
    \end{equation}
    Then, the estimation of $\hat{\Sigma}_{X \mid Y, z}$ and $\hat{\Sigma}_{X \mid y, z}$ can be expressed correspondingly as:
    \begin{equation}
    \begin{split}
        \hat{\Sigma}_{X\mid Y z} &= \hat{\Sigma}_{XY \mid z} \hat{\Sigma}_{YY\mid z}^{-1} \\     
        &= \Phi_{X} (\Psi_{z}\Phi_{Y}\Phi_{Y}^{*} + \lambda I)^{-1}\Psi_{z} \Phi_{Y} \label{estimation matrix on x|yz}
    \end{split}
    \end{equation}
    \begin{equation}
    \begin{split}
            \hat{\Sigma}_{X\mid y z} &= \hat{\Sigma}_{X\mid Y z} \phi^{Y}(y) \\
            &= \Phi_{X} (\Psi_{z}\Phi_{Y}\Phi_{Y}^{*} + \lambda I)^{-1}\Psi_{z} \Phi_{Y} \phi^{Y}(y) 
                \label{estimation mean on x|yz}
    \end{split}
    \end{equation}
    where $\Psi_{z}$ is defined as $diag ((\Phi_{Z}\Phi_{Z}^{*} + \lambda I)^{-1} \Phi_{Z}^{*}\phi^Z(z))$ Since we need to diagonalize the vector $(\Phi_{Z}\Phi_{Z}^{*} + \lambda I)^{-1} \Phi_{Z}^{*}\phi^Z(z)$ as a matrix for calculation, $\Phi_{Z} = (\phi^{Z}(z_{1}), \phi^{Z}(z_{2}), \cdots, \phi^{Z}(z_{N}))$.

\subsection{Generalization of Kernel Mean Embedding PSRs.} \label{Appendix. Generalization of Kernel Mean Embedding PSRs}

\textbf{Remark 1.} The Hilbert Space Embedding can learn a large class of dynamical systems. For example, consider a control problem as:
\begin{equation}
    \begin{cases}
        s_{t+1} = As_{t} + Ba_{t} + \epsilon \\
        o_{t+1} = Cs_{t+1} + b
    \end{cases}
\end{equation}
where $s, a, o$ mean the state, action and observation, respectively; $\epsilon, b \in \mathcal{L}^2$ (Lebesgue space) are orthogonal measures. Our goal is to find a sequence of optimal actions such that action $\{ a_t, t \in [t, t+W-1]\}$ drives the system to a target state. The conditional expectation of future observations can be expressed as:
\begin{equation}
    \mathbb{E}(o_{t:t+W-1} \mid s_t, a_{t:t+k-1}) = \Gamma_{k} s_t + U_k a_{t:t+k-1},
\end{equation}
where $\Gamma_{k}$ and $U_k$ represent the following matrix that is constituted of $\left\{C, A\right\}$ and $\left\{A, B \right\}$:
\begin{center}
    $$\Gamma_{k} =\begin{bmatrix} CA \\CA^{2} \\ \vdots \\ CA^{k}\ \end{bmatrix}$$
        \\ 
    $$U_{k} =\begin{bmatrix} B &0 &\cdots &\quad &\quad &0  
    \\AB &B &0 &\cdots &\quad &0 
    \\A^{2}B &AB &B  &0 &\cdots &0 
    \\ \quad &\quad &\vdots
    \\ A^{k-1}B &\quad &\cdots &\quad &AB &B  \ \end{bmatrix}$$
\end{center}
It is easy to observe that the new state will lose the Markovian property, and historical information will influence the future. The next state action $s_{t+1}$ will rely on estimating latent states from the observation $o_{t+1}$. When it becomes partially observable, the estimation of $s_{t}$ can be formulated as a least square problem in Hilbert space, denoted as the $\sigma-$algebra $\mathcal{F}_T = \sigma(o_t, t \in [t, t+W-1])$. The problem can be expressed as: 
\begin{equation}
    \hat{s}_{t:t+W-1} = \mathbb{E} [s_{t:t+W-1} \mid \mathcal{F}_T] 
\end{equation}
This implicit equation can be regarded as a dual form of least square problem in Hilbert space such that in Eq. \eqref{LQG 1}, \eqref{LQG 2}, \eqref{LQG 3}. A deeper connection between observation and latent states will be revealed in representing value functions. The PSR embedding framework addresses problems beyond the linear dynamic systems, and it is possible to address time-varying systems via recursively updating the future observation matrix.

\textbf{Remark 2.} A class of bounded step POMDP lies in our framework. Consider the standard POMDP $(\mathcal{S}, \mathcal{O}, \mathcal{A},\{\mathbb{T}_i \}_{i\in I}, \mathbb{O}, \mathcal{T})$, W.L.O.G the $\mathcal{S}, \mathcal{O}, \mathcal{A},\mathbb{T}, \mathbb{O}, \mathcal{T}$ are latent state, observation space, action space, transition dynamics, omission matrix, and history distribution respectively. In step $t$, the next step, latent transition dynamics can be represented by:
\begin{equation}
\begin{split}
    \mathbb{E} [ \mathbb{I}_{o_{t+1}} \mid  a_{t+1}] & = \mathbb{P} (o_{t+1} \mid a_{t+1} )
    \\ & = \sum_{s_{t+1} \in \mathcal{S}} \mathbb{P} (o_{t+1} \mid s_{t+1})  \mathbb{P}(s_{t+1} \mid s_{t}, a_{t+1}) \\ &
    = \mathbb{O}_{t}\mathbb{T}_{t} a(t)  \label{projection on sigma algebra}
\end{split}
\end{equation}
By induction, we can assert that the multi-step of POMDP can be decomposed as:
\begin{equation}
\begin{split}
        & \mathbb{E} [ \mathbb{I}_{u_{t:t+W-1} } \mid \mathcal{T}] = \mathbb{P}(u_{t:t+W-1} \mid \mathcal{T})  \\ 
        &= \begin{bmatrix} \mathbb{P}( u_{1} \mid \tau_1) &\cdots &\mathbb{P} (u_{1} \mid \tau_K) &\cdots &\mathbb{P} (u_{1} \mid \tau_{\lvert H \rvert})
    \\ \cdots &\ddots &\vdots &\ddots &\vdots 
    \\ \mathbb{P}( u_{\lvert \mathcal{U}_{h} \lvert} \mid \tau_1) &\cdots &\mathbb{P} (u_{\lvert \mathcal{U}_{h} \lvert} \mid \tau_K) &\cdots &\mathbb{P} (u_{\lvert \mathcal{U}_{h} \lvert} \mid \tau_{\lvert H \rvert}) 
    \ \end{bmatrix}  \\ 
    &= \mathbb{O}_{o_{t+W-1}} \mathbb{T}_{s_{t+W-1}} \times \cdots \times \mathbb{O}_{o_{t+1}} \mathbb{T}_{s_{t+1}} \\&
    = \underbrace{\begin{bmatrix} \mathbb{P}( u_{1} \mid s_1) &\cdots &\mathbb{P} (u_{\lvert H \lvert} \mid s_{\lvert \mathcal{S} \rvert})
    \\ \cdots &\ddots &\vdots 
    \\ \mathbb{P}( u_{\lvert \mathcal{U}_{h} \rvert} \mid s_1) &\cdots &\mathbb{P} ( u_{\lvert \mathcal{U}_{h} \rvert} \mid s_{\lvert \mathcal{S} \rvert})  
    \ \end{bmatrix}}_{Matrix \ A}       \times  
    \\  & 
    \underbrace{\begin{bmatrix} \mathbb{P}( s_1 \mid \tau_1) &\cdots &\mathbb{P} (s_{1} \mid \tau_{\lvert H \rvert})
    \\ \cdots &\ddots &\vdots 
    \\ \mathbb{P}( s_{\lvert \mathcal{S} \rvert} \mid \tau_1) &\cdots &\mathbb{P} (s_{\lvert \mathcal{S} \rvert} \mid \tau_{\lvert H  \rvert})
    \ \end{bmatrix}}_{Matrix \ B}  \label{POMDP matrices}
\end{split}
\end{equation}
Here, $\tau \in \mathcal{T}$ represents the historical distribution, while $\lvert \mathcal{U} \rvert, \lvert \mathcal{S} \rvert,$ and $\lvert H \rvert$ denote the corresponding cardinality. In the case of a continuous system, these matrices become infinitely large. We define $\mathcal{U}_{h} \subset \mathcal{O}^{W} \times \mathcal{A}^{W}, \forall W \in \mathbb{N}^{+}$ to make the concept clearer. To simplify notation, we denote the matrix $A$ and $B$ to denote the two matrices in the last line. The pseudoinverse matrix of $A$, can be interpreted as$\mathbb{O}_{t:t+w-1}^{\dag}$, which is a special case of the operator $\Sigma_{\mathcal{S \mid O}}$, allowing us to change the measure from state to observation. If we further write Eq. \eqref{POMDP matrices} as 
\begin{equation}
\begin{split}
     \mathbb{E}[ \mathbb{I}_{u_{t:t+W-1} } \mid \mathcal{T}] & = \mathbb{E}[ A^{\dag} \mathbb{I}_{u_{t:t+W-1} } \mid \mathcal{T}] \\
     & = \mathbb{E}[ \mathbb{I}_{v_{t:t+W-1} } \mid \mathcal{T}] 
\end{split}
\end{equation}
Where $v_{t:t+W-1} \in \mathcal{V}_h \subset \mathcal{S}^{W} \times \mathcal{A}^{W}, \forall W \in \mathbb{N}^{+}$. It is clear that the equation is a special case of Hilbert space embedding PSRs. This solution also serves as a dual representation of the equation given in Eq. \eqref{projection on sigma algebra}. A deeper understanding of matrices within the Hilbert space can be analogous to the adjoint operator, which becomes crucial when approximating value and risk functions.

\subsection{Bilinear Form and Function Approximation} \label{Appendix: Bilinear Form and Function Approximation}

\textbf{Lemma 3. (the existence of link functions)} For any separable functions $V$ or $C$ lies in Hilbert spaces, link functions always exist for any policy $\pi$. 

\textit{proof.} Consider any function $f: H \times \mathcal{S} \rightarrow \mathbb{R}$. Denote the one-hot encoding of $h$ as $\boldsymbol{1}(h)$ and the one-hot encoding of $s$ as $\boldsymbol{1}(s)$. We have: 
\begin{equation}
    f(h, s) = \langle f, \boldsymbol{1}(h) \otimes \boldsymbol{1}(s) \rangle
\end{equation}
We have known there exists an operator $\mathbb{O}^{\dag}: [H] \otimes [\mathcal{O}] \rightarrow [\mathcal{S}]$ as described in Eq. \eqref{Equation: pull-back value} (the use $[\cdot]$ it is a one-hot encoding), and we have that: 
\begin{equation}
\begin{split}
    f(h,s) &= \langle f, \boldsymbol{1}(h) \otimes \mathbb{O}^{\dag}_{h} \mathbb{E}_{o \sim \mathbb{O}_{h}} [\boldsymbol{1}(o)]  \rangle \\
    &= \mathbb{E}_{o \sim \mathbb{O}} \langle f, \boldsymbol{1}(h) \otimes \mathbb{O}^{\dag}_{h} (\boldsymbol{1}(o))  \rangle \label{Equation: link function one-hot encoding}
\end{split}
\end{equation}
The proof indicates the existence of link functions, and we can assert that the value link function can be expressed as $g(h, o) \defeq  \langle g, \boldsymbol{1}(h) \otimes \boldsymbol{1}(o) \rangle$ and the corresponding one-step value/risk function represented link functions are: 
\begin{equation}
\begin{split}
        V^{\pi} (h_t) &= \mathbb{E}[g^{\pi}(h_t, o_t) \mid h_t, s_t, a_{t-1} \sim \pi] \\
        & = \mathbb{E}\langle \varsigma^{\pi}, \boldsymbol{1}(h_t) \otimes \mathbb{O}^{\dag}_{h_t} (\boldsymbol{1}(o)) \rangle \\ 
        & = \mathbb{E}\underbrace{ \langle \underbrace{(\mathbb{I} \otimes \mathbb{O}^{\dag}_{h_t} )^* \varsigma^{\pi}}_{ = g^{\pi}},   \boldsymbol{1}(h_t) \otimes (\boldsymbol{1}(o)) \rangle }_{ = g^{\pi}(h_t, o_t)}
        \label{Equation: value function one-hot}
\end{split}
\end{equation}
where $\varsigma$ is related to the intrinsic reward function of the environment. 
Similarly,
\begin{equation}
\begin{split}
        C^{\pi} (h_t) &= \mathbb{E}[m^{\pi}(h_t, o_t) \mid h_t, s_t, a_{t-1} \sim \pi] \\
        & = \mathbb{E}\langle \tau^{\pi}, \boldsymbol{1}(h_t) \otimes \mathbb{O}^{\dag}_{h_t} (\boldsymbol{1}(o)) \rangle \\
        & = \mathbb{E} \underbrace{  \langle \underbrace{(\mathbb{I} \otimes \mathbb{O}^{\dag}_{h_t} )^* \tau^{\pi}}_{ = m^{\pi}} 
 , \boldsymbol{1}(h_t) \otimes (\boldsymbol{1}(o)) \rangle  }_{ = m^{\pi}(h_t, o_t)}
        \label{Equation: risk function one-hot}
\end{split}
\end{equation}
where $\tau$ is related to the setting of risks.

\textbf{The generalized version of the Lemma 2 in RKHS.}
The equation strongly connects observation space and value/risk function approximation. The Eq. \eqref{Equation: link function one-hot encoding} has a natural connection with the reproducing property in RKHS. RKHS can give a symmetric formulation since the one-hot encoding is one of the feature representations in RKHS. Please note we will provide a one-step version, and a multi-step version can be easy to derive following the same idea.

\textit{Proof.} Consider an arbitrary functional $f \in \mathcal{H}_{H} \otimes \mathcal{H}_{\mathcal{S}} \rightarrow \mathbb{R}$, we have the following form according to the reproducing property
\begin{align*}
    f(h, s) = \langle f, \phi^{H}(h) \otimes \phi^{\mathcal{S}}(s)\rangle
\end{align*}
Here, instead of using the observable operator $\mathbb{O}$, the embedding operator  $\Sigma_{\mathcal{S \mid O }, H} : \mathcal{H}_{H} \otimes \mathcal{H}_{\mathcal{U}_{\mathcal{O}}} \rightarrow  \mathcal{H}_{\mathcal{U}_{\mathcal{S}}} $, thus the generalization of Eq. \eqref{Equation: link function one-hot encoding} in RKHS can be 
\begin{equation}
    f(h,s) = \langle f, \phi^{H}(h) \otimes \Sigma_{\mathcal{S \mid O }, h} \phi^{o}(o)  \rangle \quad (\text{reproducing property})
\end{equation}
Where $\Sigma_{\mathcal{S \mid O }, h_t}$ has been an embedding operator, it is not necessary to calculate an integral form. Then for arbitrary $g(h, o) \defeq \langle g, \phi^{H}(h) \otimes \phi^{o}(o) \rangle$ and $m(h, o) \defeq \langle m, \phi^{H}(h) \otimes \phi^{o}(o) \rangle$, we have the corresponding one-step value/risk function in RKHS as 
\begin{equation}
\begin{split}
        V^{\pi} (h_t) &= \mathbb{E}[g^{\pi}(h_t, o_t) \mid h_t, s_t, a_{t-1} \sim \pi] \\
        & = \langle \varsigma^{\pi}, \phi^{H}(h_t) \otimes \Sigma_{\mathcal{S \mid O }, h_t} \phi^{o}(o) \rangle \\
        & = \langle \underbrace{(\mathbb{I} \otimes \Sigma_{\mathcal{S \mid O }, h_t})^* \varsigma^{\pi}}_{\mathbb{E}[g^{\pi}]} , \phi^{H}(h_t) \otimes \phi^{o}(o) \rangle
        \label{Equation: ones-step value function in RKHS}
\end{split}
\end{equation}
The $\varsigma^{\pi} \in \mathcal{H}_{H} \otimes \mathcal{H}_{\mathcal{O}}$, and the $(\mathbb{I} \otimes \Sigma_{\mathcal{S \mid O }, h_t})^* \varsigma^{\pi}$ is equivalent to the mean embedding of $g^{\pi}$.  
Similarly,
\begin{equation}
\begin{split}
        C^{\pi} (h_t) &= \mathbb{E}[m^{\pi}(h_t, o_t) \mid h_t, s_t, a_{t-1} \sim \pi] \\
        & = \mathbb{E}\langle \tau^{\pi}, \phi^{H}(h_t) \otimes \Sigma_{\mathcal{S \mid O }, h_t } \phi^{o}(o) \rangle  \\
        & = \langle \underbrace{(\mathbb{I} \otimes \Sigma_{\mathcal{S \mid O }, h_t})^* \tau^{\pi}}_{\mathbb{E}[m^{\pi}]} , \phi^{H}(h_t) \otimes \phi^{o}(o) \rangle
        \label{Equation: ones-step risk function in RKHS}
\end{split}
\end{equation}

\section{Main Proof}\label{Appendix: Main Theorem Proof}   
\vspace{12 pt}

\subsection{Main Theorem Proofs}
\textbf{Theorem 1. (Error bound of the forward operator)} 
Under the regularity assumption, the error bound of the forward operator $\Sigma_{\mathcal{O \mid A,} H}$ has the empirical estimation as:
\begin{equation}
\begin{split}
    \tilde{\Sigma}_{\mathcal{O \mid A,} H} & = \tilde{\Sigma}_{\mathcal{O,A,}H} (\tilde{\Sigma}_{HH} \otimes  \tilde{\Sigma}_{\mathcal{AA}} + \lambda I )^{-1} \\
    & = \tilde{\Sigma}_{\mathcal{O,A,}H} (\tilde{\Sigma}_{H,\mathcal{A}} ^2  + \lambda I)^{-1} 
\end{split}
\end{equation}
we have the probability with at least $1 - \delta $, $\forall \delta \in (0,1)$ satisfying the  
\begin{equation}
    \mathbb{P}(\lVert \Sigma_{\mathcal{O \mid A,} H} -  \tilde{\Sigma}_{\mathcal{O \mid A,} H} \rVert \geq c) \leq 1 - \delta 
\end{equation}
where $\lambda \rightarrow 0$ and $ \phi_{k}^{\mathcal{O}} \in \mathbb{C}^{n_1} $, $ \phi_{k}^{\mathcal{A}} \in \mathbb{C}^{n_2} $ and $ \phi_{k}^{H} \in \mathbb{C}^{n_3}$ for all $k \in K$, 
\begin{align*}
    & \tilde{\Sigma}_{H, \mathcal{A}}^2 =  \frac{1}{\lvert K \rvert} \sum_{k \in K} [\phi^{H}_k \otimes \phi^{\mathcal{A}}_k] [\phi^{H}_k \otimes \phi^{\mathcal{A}}_k]^* \\ 
    & c \defeq  \frac{2 \log( (n_1 + n_3n_2)/\delta) \Bar{c}_7}{3 \rho_{min}( \tilde{\Sigma}_{\mathcal{H}, \mathcal{A}}^2 )}  + \frac{\sqrt{2 \log ( (n_1 + n_3n_2)/\delta) \overline{Var(\Sigma_{\mathcal{O,A}, H})}}}{2 \rho_{min}( \tilde{\Sigma}_{\mathcal{H}, \mathcal{A}}^2 )} \\ 
    & + 
      \frac{\sqrt{\rho_{max}(\Sigma_{\mathcal{OO}})} + \sqrt{\rho_{max}(\Sigma_{\mathcal{OO}}) } \epsilon_1}{ \sqrt{ \rho_{min}( \tilde{\Sigma}_{\mathcal{H}, \mathcal{A}}^2 )}} \cdot \frac{\epsilon_2 + \lambda}{1 + \epsilon_2 + \lambda }  \\
    & \Bar{c}_7 = \frac{\max_{i,j,z} n_1 n_2 n_3 \sqrt{20 \log(2 \Bar{c}_{i,j,z}/\delta)}}{\sqrt{\lvert K \rvert}} + \frac{\sqrt{200 C n_1 n_2 n_3 \max_{i,j,z} \Bar{c}_{i,j,z}} (\log(2/\delta))^{\frac{3}{4}} }{\lvert K \rvert^{\frac{1}{4}}}  \\
    & \overline{Var(\Sigma_{\mathcal{O, A,} H})} =  \frac{\max_{i,j,z} n_1 n_2 n_3 \sqrt{20 \log(2 \Bar{c}_{i,j,z}/\delta)}}{\sqrt{\lvert K \rvert}}  \\
    & \epsilon_1 \lesssim \frac{ \log (n_1/\delta)}{ \rho_{\max}(\Sigma_{\mathcal{OO}})} \\
    & \epsilon_2 \lesssim \frac{ \log (n_2/\delta)}{ \rho_{\max}(\Sigma_{\mathcal{AA}})}
\end{align*}
\textit{Proof.} According to the definition of linear PSR, we have 
\begin{align*}
    \Sigma_{\mathcal{O, A}, H} \defeq \mathbb{E}[\phi^{\mathcal{O}}(t_h(o)) \otimes \phi^{\mathcal{A}}(t_h(a))  \otimes \phi^{H}(h)] \in \mathbb{C}^{n_1 \times n_2 \times n_3}
\end{align*}
Where $"\otimes"$ indicates the Khatri–Rao product (as known column-wise Kronecker product). In this situation, the embedding of $\phi^{\mathcal{O}}$ is under the condition of $\phi^{\mathcal{A}} \otimes \phi^{H}$ such that 
\begin{align*}
    \mathbb{E}[\phi^{\mathcal{O}}(o) \mid \phi^{\mathcal{A}}(a),  \phi^{H}(h)] = \Sigma_{\mathcal{O \mid A}, H} \times (\phi^{H}(h) \otimes \phi^{\mathcal{O}}(o))
\end{align*}
Then we can infer that the $3-mode$ tensor property of forward operator $\Sigma_{\mathcal{O \mid A}, H} $. By the empirical estimation in Proposition 1, the estimation of the forward operator $ \tilde{\Sigma}_{\mathcal{O \mid A,} H} $ can be written as
\begin{align*}
    & \tilde{\Sigma}_{H, \mathcal{A}}^2 =  \frac{1}{\lvert K \rvert} \sum_{k \in K} [\phi^{H}_k \otimes \phi^{\mathcal{A}}_k] [\phi^{H}_k \otimes \phi^{\mathcal{A}}_k]^*  
\end{align*}
where $\tilde{\Sigma}_{H, \mathcal{A}}^2 \in \mathbb{C}^{n_3n_2 \times n_3n_2}$ is a Hermitian matrix. On the other hand, empirical estimation of $\Sigma_{\mathcal{O,A}, H}$ can be written as
\begin{align*}
    \tilde{\Sigma}_{\mathcal{O,A}, H} = \frac{1}{\lvert K \rvert} \sum_{k \in K} \phi^{\mathcal{O}}_{k}  [\phi^{H}_k \otimes \phi^{\mathcal{A}}_k]^*  
\end{align*}
where $\Sigma_{\mathcal{O,A}, H} \in \mathbb{C}^{n_1 \times n_3n_2} $. Under the assistance of Corollary 3, the $\tilde{\Sigma}_{\mathcal{O \mid A,} H}$ can be estimated as
\begin{align*}
    \tilde{\Sigma}_{\mathcal{O \mid A,} H} = \tilde{\Sigma}_{\mathcal{O,A}, H} (\tilde{\Sigma}_{H, \mathcal{A}}^2 + \lambda I)^{-1}
\end{align*}
The proof of this theorem is highly similar to Corollary 3, we have
\begin{equation}
    \mathbb{P}(\lVert \Sigma_{\mathcal{O \mid A,} H} -  \tilde{\Sigma}_{\mathcal{O \mid A,} H} \rVert \geq c) \leq 1 - \delta 
\end{equation}
The error of the forward operator can be expressed as: 
\begin{align*}
    & \ \quad \lVert \Sigma_{\mathcal{O \mid A,} H} -  \tilde{\Sigma}_{\mathcal{O \mid A,} H} \rVert \\
    & = \lVert \Sigma_{\mathcal{O,A}, H} (\Sigma_{H, \mathcal{A}}^2)^{-1} - \Tilde{\Sigma}_{\mathcal{O A}} (\tilde{\Sigma}_{H, \mathcal{A}}^2 + \lambda I)^{-1}  \rVert \\
    & = \lVert  \Sigma_{\mathcal{O,A}, H}  (\Sigma_{H, \mathcal{A}}^2)^{-1} - \Tilde{\Sigma}_{\mathcal{O A}} (\Sigma_{H, \mathcal{A}}^2)^{-1} + \Tilde{\Sigma}_{\mathcal{O A}} (\Sigma_{H, \mathcal{A}}^2)^{-1} - \Tilde{\Sigma}_{\mathcal{O A}} (\tilde{\Sigma}_{H, \mathcal{A}}^2 + \lambda I)^{-1}  \rVert \\
    & \leq \underbrace{\lVert  \Sigma_{\mathcal{O,A}, H}  (\Sigma_{H, \mathcal{A}}^2)^{-1} - \Tilde{\Sigma}_{\mathcal{O A}} (\Sigma_{H, \mathcal{A}}^2)^{-1} \rVert}_{\text{P1}} + \underbrace{\lVert \Tilde{\Sigma}_{\mathcal{O A}} (\Sigma_{H, \mathcal{A}}^2)^{-1} - \Tilde{\Sigma}_{\mathcal{O A}} (\tilde{\Sigma}_{H, \mathcal{A}}^2 + \lambda I)^{-1}  \rVert}_{\text{P2}} 
\end{align*}
Using the result in the Proposition 1, 2 and Corollary 1, we have
\begin{equation}
\begin{split}
    c \defeq  & \underbrace{\frac{2 \log( (n_1 + n_3n_2)/\delta) \Bar{c}_7}{3 \rho_{min}( \tilde{\Sigma}_{\mathcal{H}, \mathcal{A}}^2 )}  + \frac{\sqrt{2 \log ( (n_1 + n_3n_2)/\delta) \overline{Var(\Sigma_{\mathcal{O,A}, H})}}}{2 \rho_{min}( \tilde{\Sigma}_{\mathcal{H}, \mathcal{A}}^2 )}}_{\text{P1 error bound}} \\ 
    & + 
     \underbrace{ \frac{\sqrt{\rho_{max}(\Sigma_{\mathcal{OO}})} + \sqrt{\rho_{max}(\Sigma_{\mathcal{OO}}) } (\epsilon_1)}{ \sqrt{ \rho_{min}( \tilde{\Sigma}_{\mathcal{H}, \mathcal{A}}^2 )}} \cdot \frac{\epsilon_2 + \lambda}{1 + \epsilon_2 + \lambda }}_{\text{P2 error bound}}    
\end{split}
\end{equation}
\begin{align*}
    & \Bar{c}_7 = \frac{\max_{i,j,z} n_1 n_2 n_3 \sqrt{20 \log(2 \Bar{c}_{i,j,z}/\delta)}}{\sqrt{\lvert K \rvert}} + \frac{\sqrt{200 C n_1 n_2 n_3 \max_{i,j,z} \Bar{c}_{i,j,z}} (\log(2/\delta))^{\frac{3}{4}} }{\lvert K \rvert^{\frac{1}{4}}}  \\
    & \sqrt{\overline{Var(\Sigma_{\mathcal{O, A,} H})}} =  \frac{\max_{i,j,z} n_1 n_2 n_3 \sqrt{20 \log(2 \Bar{c}_{i,j,z}/\delta)}}{\sqrt{\lvert K \rvert}}  \\ 
    & \epsilon_1 \lesssim \frac{ \log (n_1/\delta)}{ \lvert K \rvert \rho_{\max}(\Sigma_{\mathcal{OO}})} \\
    & \epsilon_2 \lesssim \frac{ \log (n_2/\delta)}{  \lvert K \rvert \rho_{\max}(\Sigma_{\mathcal{AA}})}
\end{align*}
with at least $1 - 3 \delta $ probability due to the union bound.

\vspace{12 pt}

\textbf{Theorem 2. (Error bound of the shifted forward operator)} Under the same conditions in Theorem n, the error bound of the shifted forward operator is 
\begin{align*}
    & \lVert  \tilde{\mathcal{P}}_{a, o} \tilde{\Sigma}_{\mathcal{O \mid A}, H_{t}} - \Sigma_{\mathcal{O \mid A}, H_{t+1}} \rVert = \lVert  \Sigma_{\mathcal{O \mid A}, H_{t+1}} - \Sigma_{\mathcal{O \mid A}, H_{t+1}} \rVert \lesssim c \\
    & c = \frac{2 \log( (n_1 + n_3n_2)/\delta) \Bar{c}_7}{3 \rho_{min}( \tilde{\Sigma}_{\mathcal{H}, \mathcal{A}}^2 )}  + \frac{\sqrt{2 \log ( (n_1 + n_3n_2)/\delta) \overline{Var(\Sigma_{\mathcal{O,A}, H})}}}{2 \rho_{min}( \tilde{\Sigma}_{\mathcal{H}, \mathcal{A}}^2 )} \\
    & + \sqrt{n_1^2 \frac{\max_{i,j} \lVert c_{i,j} \rVert}{\lvert K \rvert} \log(2 n_1^2/\delta )} \lVert \Sigma_{\mathcal{O \mid A}, H_{t}}  \rVert 
\end{align*}
with probability at least $1 - \delta$, the other symbol definitions are the same with theorem n.

\vspace{12 pt}

\textit{Proof.}
According to the definition of shifted forward operator, we have that
\begin{align*}
    \Sigma_{\mathcal{O \mid A}, H_{t+1}} = \mathcal{P}_{a, o} \Sigma_{\mathcal{O \mid A}, H_{t}}
\end{align*}
\begin{align*}
     \mathcal{P}_{a, o} = \Sigma_{\mathcal{O \mid A}, H_{t+1}} (\Sigma_{\mathcal{O \mid A}, H_{t}})^{-1} 
\end{align*}
\begin{align*}
    \Rightarrow \mathcal{P}_{a, o}\mathcal{P}_{a, o}^{*} =\Sigma_{\mathcal{O \mid A}, H_{t+1}} (\Sigma_{\mathcal{O \mid A}, H_{t}})^{-1}  [(\Sigma_{\mathcal{O \mid A}, H_{t}})^{-1}]^* \Sigma_{\mathcal{O \mid A}, H_{t+1}}^*
\end{align*}
The shifted operator is bounded by 
\begin{align*}
     \lVert \mathcal{P}_{a, o} \rVert_{\infty} \leq \frac{\rho_{max}(\sqrt{ \Sigma_{\mathcal{O \mid A}, H_{t+1}}\Sigma_{\mathcal{O \mid A}, H_{t+1}}^*})}{\rho_{min}(\sqrt{ \Sigma_{\mathcal{O \mid A}, H_{t}}\Sigma_{\mathcal{O \mid A}, H_{t}}^*}))} \leq c_8    
\end{align*}
The error bound of $\tilde{\Sigma}_{\mathcal{O \mid A}, H_{t+1}}$ can be 
\begin{align*}
    & \ \quad\lVert  \tilde{\mathcal{P}}_{a, o} \tilde{\Sigma}_{\mathcal{O \mid A}, H_{t}} - \Sigma_{\mathcal{O \mid A}, H_{t+1}} \rVert \\
    & = \lVert  \tilde{\mathcal{P}}_{a, o} \tilde{\Sigma}_{\mathcal{O \mid A}, H_{t}} - 
    \tilde{\mathcal{P}}_{a, o} \Sigma_{\mathcal{O \mid A}, H_{t}} + \tilde{\mathcal{P}}_{a, o} \Sigma_{\mathcal{O \mid A}, H_{t}} 
    - \Sigma_{\mathcal{O \mid A}, H_{t+1}} \rVert \\
    & = \lVert  \tilde{\mathcal{P}}_{a, o} \tilde{\Sigma}_{\mathcal{O \mid A}, H_{t}} - 
    \tilde{\mathcal{P}}_{a, o} \Sigma_{\mathcal{O \mid A}, H_{t}} + \tilde{\mathcal{P}}_{a, o} \Sigma_{\mathcal{O \mid A}, H_{t}} 
    - \mathcal{P}_{a, o} \Sigma_{\mathcal{O \mid A}, H_{t}} \rVert \\
    & = \underbrace{\lVert  \tilde{\mathcal{P}}_{a, o} \tilde{\Sigma}_{\mathcal{O \mid A}, H_{t}} - 
    \tilde{\mathcal{P}}_{a, o} \Sigma_{\mathcal{O \mid A}, H_{t}} \rVert }_{\text{P1}} + \underbrace{\lVert \tilde{\mathcal{P}}_{a, o} \Sigma_{\mathcal{O \mid A}, H_{t}} 
    - \mathcal{P}_{a, o} \Sigma_{\mathcal{O \mid A}, H_{t}} \rVert }_{\text{P2}}
\end{align*}
According to the Theorem n, the Part 1 error bound satisfies the following condition with at least $1-\delta$ probability
\begin{equation}
\begin{split}
    & \ \quad \lVert  \tilde{\mathcal{P}}_{a, o} \tilde{\Sigma}_{\mathcal{O \mid A}, H_{t}} - 
    \tilde{\mathcal{P}}_{a, o} \Sigma_{\mathcal{O \mid A}, H_{t}} \rVert \\
    & \leq \lVert \tilde{\mathcal{P}}_{a, o}  \rVert_{\infty} \lVert  \tilde{\Sigma}_{\mathcal{O \mid A}, H_{t}} - \Sigma_{\mathcal{O \mid A}, H_{t}} \rVert \\
    & \leq c_8 c \label{error 1: shifted forward}
\end{split}
\end{equation}
where $c$ has the same definition in the Theorem n. 

\vspace{12 pt}

For the error bound of Part 2, we have 
\begin{align*}
    &\ \quad \lVert \tilde{\mathcal{P}}_{a, o} \Sigma_{\mathcal{O \mid A}, H_{t}} 
    - \mathcal{P}_{a, o} \Sigma_{\mathcal{O \mid A}, H_{t}} \rVert  \\
    & \leq \lVert \tilde{\mathcal{P}}_{a, o} -  \mathcal{P}_{a, o}  \rVert \lVert \Sigma_{\mathcal{O \mid A}, H_{t}}  \rVert
\end{align*}
For the random matrix $\tilde{\mathcal{P}}_{a, o} \in BH(\mathbb{C}^{n_1 \times n_3n_2}, \mathbb{C}^{n_1 \times n_3n_2})$ we  have the following property 
\begin{align*}
    \mathbb{E}_{\tilde{\mathcal{P}} \sim \pi(\mathcal{P})} [\tilde{\mathcal{P}}_{a, o} - \mathcal{P}_{a,o} ] \rightarrow 0
\end{align*}
$Err_{\mathcal{P}} \defeq \tilde{\mathcal{P}}_{a, o} - \mathcal{P}_{a,o} $ and for all $[Err_{\mathcal{P}} ]_{i \in [n_1], j \in [n_1]}$ we have
\begin{align*}
    \lVert [Err_{\mathcal{P}} ]_{i,j} \rVert \leq \frac{c_{i,j}}{\lvert K \rvert}
\end{align*}
Where $\lvert K\rvert$ is the sample size. In such a situation, we have
\begin{align*}
    & \ \quad \lVert Err_{\mathcal{P}}  \rVert_F  \\
    & = \sqrt{\sum_{i \in [n_1]} \sum_{j \in [n_2]} [Err_{\mathcal{P}}]_{i, j} } \\
    & \leq n_1^2 \frac{\max_{i,j} \lVert c_{i,j} \rVert}{\lvert K \rvert}
\end{align*}
By the McDiarmid's Inequality in Lemma 3, we have 
\begin{align*}
    & \ \quad \mathbb{P}(\lVert \tilde{\mathcal{P}}_{a, o} - \mathcal{P}_{a,o}  \rVert_{\infty} \geq  n_1 \epsilon ) \\
    & \leq \mathbb{P}(\lVert \tilde{\mathcal{P}}_{a, o} - \mathcal{P}_{a,o}  \rVert_{F} \geq  n_1^2 \epsilon ) \\ 
    & \leq 2 \exp( - \frac{2 \epsilon^2}{\sum_{ i \in [n_1], j \in [n_1]} \frac{c_{i,j}^2}{\lvert K \rvert} }) \\
    & \leq  2 \exp( - \frac{2 \epsilon^2}{n_1^2 \frac{\max_{i,j} \lVert c_{i,j} \rVert}{\lvert K \rvert}})
\end{align*}
set $2 \exp( - \frac{2 \epsilon^2}{n_1^2 \frac{\max_{i,j} \lVert c_{i,j} \rVert}{\lvert K \rvert}}) \leq \frac{\delta}{n_1^2} $, obtaining 
\begin{equation}
    \mathbb{P} \biggl( \lVert \tilde{\mathcal{P}}_{a, o} - \mathcal{P}_{a,o}  \rVert_{F} \leq \sqrt{n_1^2 \frac{\max_{i,j} \lVert c_{i,j} \rVert}{\lvert K \rvert} \log(2 n_1^2/\delta )} \biggr) \geq 1 - \delta 
\end{equation}
\begin{equation}
\begin{split}
    \Rightarrow & \ \quad \lVert \tilde{\mathcal{P}}_{a, o} -  \mathcal{P}_{a, o}  \rVert \lVert \Sigma_{\mathcal{O \mid A}, H_{t}}  \rVert \\
    & \leq \lVert \tilde{\mathcal{P}}_{a, o} -  \mathcal{P}_{a, o}  \rVert_F \lVert \Sigma_{\mathcal{O \mid A}, H_{t}}  \rVert \\
    & \leq  \sqrt{n_1^2 \frac{\max_{i,j} \lVert c_{i,j} \rVert}{\lvert K \rvert} \log(2 n_1^2/\delta )} \lVert \Sigma_{\mathcal{O \mid A}, H_{t}}  \rVert \label{error 2: shifted forward}
\end{split}
\end{equation}
Combine the Eq. \eqref{error 1: shifted forward} and \eqref{error 2: shifted forward}, we have the following the error bound
\begin{align*}
    & \ \quad \lVert  \tilde{\mathcal{P}}_{a, o} \tilde{\Sigma}_{\mathcal{O \mid A}, H_{t}} - 
    \tilde{\mathcal{P}}_{a, o} \Sigma_{\mathcal{O \mid A}, H_{t}} \rVert  + \lVert \tilde{\mathcal{P}}_{a, o} \Sigma_{\mathcal{O \mid A}, H_{t}} 
    - \mathcal{P}_{a, o} \Sigma_{\mathcal{O \mid A}, H_{t}} \rVert \\
    & \leq \lVert \tilde{\mathcal{P}}_{a, o}  \rVert_{\infty} \lVert  \tilde{\Sigma}_{\mathcal{O \mid A}, H_{t}} - \Sigma_{\mathcal{O \mid A}, H_{t}} \rVert + \lVert \tilde{\mathcal{P}}_{a, o} -  \mathcal{P}_{a, o}  \rVert_F \lVert \Sigma_{\mathcal{O \mid A}, H_{t}}  \rVert \\
    & \leq c_8 \biggl( \frac{2 \log( (n_1 + n_3n_2)/\delta) \Bar{c}_7}{3 \rho_{min}( \tilde{\Sigma}_{\mathcal{H}, \mathcal{A}}^2 )}  + \frac{\sqrt{2 \log ( (n_1 + n_3n_2)/\delta) \overline{Var(\Sigma_{\mathcal{O,A}, H})}}}{2 \rho_{min}( \tilde{\Sigma}_{\mathcal{H}, \mathcal{A}}^2 )} \biggr) \\
    & + \sqrt{n_1^2 \frac{\max_{i,j} \lVert c_{i,j} \rVert}{\lvert K \rvert} \log(2 n_1^2/\delta )} \lVert \Sigma_{\mathcal{O \mid A}, H_{t}}  \rVert 
\end{align*}

\vspace{12 pt}

\textbf{Theorem 4. ($\epsilon$-suboptimal policy with polynomial sample complexity)} When Theorems 1, 2 and 3 hold, the safe policy can converge to a $\epsilon-$suboptimal policy with a polynomial sample complexity. More specifically, it can guarantee the probability with at least $1 - \delta $, $\forall \delta \in (0,1)$ satisfying the condition as
\begin{align*}
    \tilde{V}^{\tilde{\pi}}(h_0) \geq V^{\pi^*}(h_0) - \epsilon \quad and \quad   \tilde{C}_{i}^{\tilde{\pi}}(h_0) \leq \Bar{C}_{i} + \epsilon, \ \forall h_o \in H, \ i \in [N] 
\end{align*}
\newline
\textit{Proof. } The estimation of $\tilde{V}^\pi, \tilde{C}^{\pi}$ are all parameterized by $\theta \in \Theta$ and $\pi \in \Pi$. More specifically, $\theta$ is the parameter of various operators, since both  $V^\pi, C^{\pi}$ are determined by the forward operator and shifted operator. In this situation, we can use $J(\theta, \pi, \eta), \forall \theta \in \Theta; \pi \in \Pi$ to denote the KKT condition of safe PSR problem as 
\begin{align*}
    J(\theta, \pi, \eta) = \tilde{V}^{\tilde{\pi}}(h_0) - \sum_{i} \eta_i (\tilde{C}^{\tilde{\pi}}(h_0) - \Bar{C}_i)
\end{align*}
where the dual variable $\eta$ the solution of the optimization problem is on the saddle point as 
\begin{align*}
    (\tilde{\pi}, \tilde{\theta}, \tilde{\eta}) = \arg \max_{\pi \in \Pi} \text{arg}\min_{\eta, \theta \in \Theta} J(\theta, \pi, \eta)
\end{align*}
By the definition of saddle point, it can be derived that
\begin{align*}
     J(\tilde{\theta}, \pi^{*}, \eta^{*}) \leq J(\tilde{\theta}, \tilde{\pi}, \tilde{\eta}) \leq J(\theta^{*}, \tilde{\pi}, \tilde{\eta})
\end{align*}
\begin{align*}
    \Rightarrow \quad & J(\theta^*, \pi^*, \eta^*) - J(\theta^{*}, \tilde{\pi}, \tilde{\eta}) \\
    & \leq J(\theta^*, \pi^*, \eta^*) -  J(\tilde{\theta}, \pi^{*}, \eta^{*}) 
\end{align*}
To simplify the notions, we denote the $\underline{B}^*  \tilde{V}(h_t) =  \max_{\pi \in \Pi} \min_{\eta, \theta \in \Theta} J(\theta, \pi, \eta)$. Under this definition, the target of the problem becomes to obtain the contraction of the $\lVert \underline{B}^* V - \underline{B}^*  \tilde{V} \rVert$. 
\begin{itemize}
    \item \textbf{Case 1.} If the $[\tilde{C}_{i}^{\pi} (h_t) -\Bar{C}_{i}] > 0 , \forall i \in [N]$ for all action $a_t \in \mathcal{A}$, the $\underline{B}^*  \tilde{V}(h_t) \rightarrow -\infty$, which means the certain risk in the future $(W+1)$ steps. 
    \item  \textbf{Case 2.} If there exist $[\tilde{C}_{t}^{\pi} (h_t)  -  \Bar{C}_{i}] \leq 0, \forall i \in [N]$ for some action $a_t \in \mathcal{A}$. We can assert the contraction of $\lVert \underline{B}^* \tilde{V} - \underline{B}^* V \rVert_{\infty} < \epsilon$ ($\epsilon$ is an arbitrarily small value) with a polynomial sample complexity, proving it needs a lemma, the details are listed below. 
\end{itemize}
Consider two arbitrary functions $f$ and $g$, we have
\begin{equation}
    \lvert \max_{x} f(x) - \max_{x} g(x) \rvert  \leq \max_{x} \lvert f(x) - g(x) \rvert
\end{equation}
To see this, we suppose $\max_{x} f(x) > \max_{x} g(x)$ (with respect the symmetric case) and let $x^* \in_x f(x)$, then
\begin{equation}
\begin{split}
    & \lvert \max_x f(x) - \max_{x} g(x) \lvert = f(x^*) - \max_x g(x) \\
    & \leq f(x^*) - g(x^*) \leq \max_{x} \lvert f(x) - g(x) \rvert  
\end{split} 
\end{equation}
Similarly, the symmetric case of Eq. \eqref{Equation: lemma max} can be indicated such that
\begin{equation}
    \lvert \min_{x} f(x) - \min_{x} g(x) \rvert \leq \max_{x} \lvert f(x) - g(x) \lvert
\end{equation}
Thus in our case, we have 
\begin{equation}
\begin{split}
    & \lVert \underline{B}^* \tilde{V} - \underline{B}^* V \rVert_{\infty} = \sup_{h_t, a_{t-1}} \lVert \underline{B}^* \tilde{V} - \underline{B}^* V \rVert \\
    & \leq \sup_{h_t} \bigg\lVert \mathbb{E} [ \langle \tilde{g}^{\pi}_t, \tilde{\Sigma}_{{o \mid a, h_t}} \phi^{a} (a_{t-1}) \rangle \mid h_t, a_{t-1} \sim \tilde{\pi} ] \\
       & \qquad + \mathbb{E} [ \mathbb{E}[\langle \tilde{g}^{\pi}_{t+1}, \tilde{\mathcal{P}}_{o_t, a_{t-1}} \tilde{\Sigma}_{\mathcal{O \mid A} , h_{t}} \phi^{\mathcal{O}}(t_{h+1}(o)) \rangle \mid t_{h+1}(a) \sim \tilde{\pi}]\mid h_{t}, o_t, a_{t-1} \sim \tilde{\pi}] \\
       & - \mathbb{E} [ \langle g^{\pi}_t, \Sigma_{{o \mid a, h_t}} \phi^{a} (a_{t-1}) \rangle \mid h_t, a_{t-1} \sim \pi^{*} ] \\
       & \qquad - \mathbb{E} [ \mathbb{E}[\langle g^{\pi}_{t+1}, \mathcal{P}_{o_t, a_{t-1}}\Sigma_{\mathcal{O \mid A} , h_{t}} \phi^{\mathcal{O}}(t_{h+1}(o)) \rangle \mid t_{h+1}(a) \sim \pi]\mid h_{t}, o_t, a_{t-1} \sim \pi^{*}] \bigg\rVert \\ 
    & \leq \mathcal{O}(\epsilon) \\
    & 
\end{split}
\end{equation}
The second line of the equation holds because the available action set satisfies the safety constraint is smaller than the whole action set, the Eq. \eqref{Equation: lemma max} and \eqref{Equation: min lemma} indicates that:
\begin{equation}
\begin{split}
    & \lVert \underline{B}^* \tilde{V} - \underline{B}^* V \rVert_{\infty} \leq \lVert B^* \tilde{V} - B^* V \rVert_{\infty}  \\ 
    & =  \sup_{h_t} \lVert B^* \tilde{V}(h_t)  - B^* V(h_t) \rVert    
\end{split}
\end{equation}
The last line equation is due to Theorem 3, the error bounded can be easy to derive by using triangular inequalities and decomposing the error by parts. Due to Theorem 1 and 2, the error bound shrinks with the polynomially with respect to the data size $\lvert K \rvert$. Also, it can be indicated that error bound $\lVert \underline{B}^* V - \underline{B}^*  \tilde{V} \rVert$ is weaker than the result  $BL(\tilde{\pi}, g, \pi^*)$ Eq. \eqref{Equation: bellman loss}. 

\subsection{Technical Details}
\textbf{Lemma 4. (McDiarmid's Inequality)}. Let $X = (X_{1}, \cdots, X_{n})$ be independent random variables with ranges $X_{i} \in \mathcal{X} \subset \mathbb{C}$ for all $i \in [n]$. Let $y: \mathcal{X}^n \rightarrow \mathbb{C}$ be any function. If there exists constant $c_i$ for any $i \in [n]$,
\begin{align*}
    \lvert y(X_{1}, \cdots, X_{n}) - y(X^{'}_1, \cdots, X^{'}_{n}) \rvert \leq c_i
\end{align*}
for any $ (X^{(1)}, \cdots, X^{(n)}), (X^{(1), '}, \cdots, X^{(n), '}) \in \mathcal{X}$ that differ only in $i-$th coordinate, then it holds for any $\epsilon > 0$ that 
\begin{align*}
    \mathbb{P}(\lVert y(X_{1}, \cdots, X_{n}) - \mathbb{E}[y(X_{1}, \cdots, X_{n})] \rVert \geq \epsilon ) \leq 2 \exp( - \frac{2 \epsilon^2}{\sum_{ i \in [n]} c_i^2})
\end{align*}

\vspace{12 pt}

\textbf{Lemma 5. } Let $\mu_1, \cdots \mu_k$ be $\lvert K \rvert$ distributions over $\mathcal{X}$ and $k(\cdot, \cdot)$ be kernel function over $\mathcal{X \times X}$ that satisfies $k(x_1, x_2) \leq 1$ $\forall x_1, x_2 \in \mathcal{X}$. Suppose $X_t$ is i.i.d sample from distribution $\mu_t$ for any $k \in \lvert K \rvert$. Define $y: \mathcal{X}^n \rightarrow \mathbb{C}$
\begin{align*}
    y(X_{1}, \cdots, X_{n}) \defeq \lVert \frac{1}{\lvert K \rvert} \sum_{k \in K} k_{\mathbb{I}}(X_k, X) -\frac{1}{\lvert K \rvert} \sum_{k \in K} m_{\mu_k} \rVert
\end{align*}
where $\mathbb{I}$ is the Dirac measure function and $m$ is the mean embedding function under distribution $\mu_t$. Then it has the probability with at least $1 - \delta$, $\forall \delta \in (0,1)$ satisfying that 
\begin{align*}
     y(X_{1}, \cdots, X_{n}) \leq \frac{\sqrt{20 \log(2/\delta)}}{\sqrt{\lvert K \rvert}} 
\end{align*}

\vspace{12 pt}

\textit{Proof.} By Jensen's inequality, obtaining
\begin{align*}
    & \ \quad \mathbb{E}[ y(X_{1}, \cdots, X_{n}) ] \\
    & \leq \sqrt{\mathbb{E}[\lVert \frac{1}{\lvert K \rvert} \sum_{k \in K} k_{\mathbb{I}}(X_k, X) -\frac{1}{\lvert K \rvert} \sum_{k \in K} m_{\mu_k} \rVert^2]} \\
    & \leq  \frac{1}{\lvert K \rvert} \sqrt{\mathbb{E}[\lVert \sum_{k \in K} k_{\mathbb{I}}(X_k, X)\rVert^2] - 2\mathbb{E}[\lVert \sum_{k \in K} k_{\mathbb{I}}(X_k, X) \rvert \lVert \sum_{k \in K} m_{\mu_k} \rVert]+ \mathbb{E}[\lVert \sum_{k \in K} m_{\mu_k}  \rVert^2]}
\end{align*}
By the property of kernel mean embedding, we have
\begin{align*}
    \mathbb{E}[\lVert \sum_{k \in K} k_{\mathbb{I}}(X_k, X) \rvert ] = \lVert \sum_{k \in K} m_{\mu_k}  \rVert
\end{align*}
Then,
\begin{equation}
\begin{split}
    & \ \quad \mathbb{E}[ y(X_{1}, \cdots, X_{n}) ] \\
    & \leq  \frac{\sqrt{2}}{\lvert K \rvert} \sqrt{\mathbb{E}[\lVert \sum_{k \in K} k_{\mathbb{I}}(X_k, X)\rVert^2] -  \mathbb{E}[\lVert \sum_{k \in K} m_{\mu_k} \rVert^2} \\
    & =  \frac{\sqrt{2}}{\lvert K \rvert} \sqrt{ \sum_{k \in K}  \sum_{k^{'} \in K} \mathbb{E}[k_{\mathbb{I}}(X_k, X_k^{'})] - \mathbb{E}_{X_k\sim \mu_t, X_k^{'} \sim \mu_t}[k(X_k, X_k^{'})]} \label{dirac measure minus mean embedding}
\end{split}    
\end{equation}

Since $k(x_1, x_2) \leq 1$, for all $x_1, x_2 \in \mathcal{X}$, and 
\begin{align*}
    & \Rightarrow \frac{\sqrt{2}}{\lvert K \rvert} \sqrt{ \sum_{k \in K}  \sum_{k^{'} \in K} \mathbb{E}[k_{\mathbb{I}}(X_k, X_k^{'})] - \mathbb{E}_{X_k\sim \mu_t, X_k^{'} \sim \mu_t}[k(X_k, X_k^{'})]} \\
    & \leq \frac{2}{\sqrt{\lvert K \rvert}}
\end{align*}
Also, for any $k \in K$, we have 
\begin{align*}
    & \ \quad \lVert y(X_1, \cdots, X_t,\cdots, X_{k}) -  y(X_1, \cdots, X_t^{'},\cdots, X_{k}) \rvert \\
    & \leq \frac{1}{\lvert K \rvert} \lVert k_{\mathcal{I}}(X_t, X) - k_{\mathcal{I}}(X_t^{'}, X)  \rVert
\end{align*}
For any $X_t, X_t^{'} $, for have 
\begin{align*}
    & \ \quad \lVert k_{\mathcal{I}}(X_t, X) - k_{\mathcal{I}}(X_t^{'}, X)  \rVert^2 \\
    & \leq k(X_t, X_t) - 2k(X_t, X_t^{'}) + k(X_t^{'}, X_t^{'})   \quad   \\
    & \leq 4
\end{align*}
By the McDiarmid's inequality in Lemma 3, we have with probability at least $1 -\delta, \forall \delta\in(0,1)$ that 
\begin{align*}
    \lvert y(X_1, \cdots, X_{k}) - \mathbb{E}[y(X_1, \cdots, X_{k})] \rvert \leq \frac{2\sqrt{2 \log(2/\delta)}}{\sqrt{\lvert K \rvert}}
\end{align*}
By the triangle inequality, we have  with probability at least $1 -\delta, \forall \delta\in(0,1)$ that 
\begin{equation}
\begin{split}
    \lvert  y(X_1, \cdots, X_{k})  \rvert \leq & \mathbb{E}[y(X_1, \cdots, X_{k})] + \lvert y(X_1, \cdots, X_{k}) - \mathbb{E}[y(X_1, \cdots, X_{k})] \rvert \\
    &\leq \frac{2\sqrt{2 \log(2/\delta)} + 2}{\sqrt{\lvert K \rvert}}  \leq \frac{\sqrt{20 \log(2/\delta)}}{\sqrt{\lvert K \rvert}} 
\end{split}
\end{equation}

\vspace{12pt}

\textbf{Lemma 6 (Isotropic Random Matrices) \cite{vershynin2018high}.} Let $B \in \mathbb{C}^{n \times m}$, and $X$ is an isotropic random vector in $\mathbb{R}^{m}$, we have the 
\begin{align*}
    \mathbb{E}[\lVert BX \rVert_2] = \lVert B \rVert_{F}
\end{align*}

\vspace{12 pt}

\textbf{Lemma 7 (Norm Concentration of Isotropic Random Matrices) \cite{vershynin2018high}.}   Let $B \in \mathbb{C}^{n \times m}$,  and $X$ is an isotropic random vector in $\mathbb{R}^{m}$, we have the sub-gaussian distribution as
\begin{align*}
    \mathbb{P} (\lvert \lVert BX \rVert - \lVert B \rVert_F \rvert \leq \epsilon \lVert B \rVert_F ) \leq 2\exp(-c\epsilon^2 \frac{\lVert B \rVert_F^2}{C^4 \lVert B \rVert^2})
\end{align*}
Changing variables to $t = \epsilon \lVert B \rVert_F$, we obtain
\begin{align*}
    \mathbb{P} (\lvert \lVert BX \rVert - \lVert B \rVert_F \rvert \leq t) \leq 2\exp(- \frac{ct^2}{C^4 \lVert B \rVert^2})    
\end{align*}

In this context, we will use $c_1, c_2, \cdots$ to represent the constant in various inequalities. We give a mild $\gamma-$regularity assumption of operators as $\rho_{min}(\Sigma) > \gamma$. 
\\ 
\newline
\textbf{Lemma 8. (Matrix Chernoff Inequality) \cite{tropp2015introduction}}. Consider a finite sequence $\{R_{k}\}, \forall R_{k} \in \mathbb{C}^{n \times n} $ of independent, random, Hermitian matrices. Assume that
\begin{center}
    $\gamma \leq \rho_{min}(R_{k})$ and $\rho_{max}(R_{k}) \leq L$ for each index $k$. 
\end{center}
The approximation of matrix $M$
\begin{equation}
    \tilde{M} = \sum_{k} R_k
\end{equation}

We use $\rho$ to denote the spectrum of the random matrix. Then the random matrix $M = \sum_{k} R_{k}$ has the following property
\begin{equation}
\begin{split}
    \mathbb{P}\{\mathbb{E}(\rho_{min}(\tilde{M})) \leq (1 - \epsilon)\rho_{min}(M) \}  & \leq n[\frac{\exp(-\epsilon)}{(1-\epsilon)^{1-\epsilon}}]^{\rho_{min}(M)/L} 
    \\& \leq c_1ne^{-\epsilon \rho_{min}(M)/L}
\end{split}
\end{equation}
and 
\begin{equation}
\begin{split}
    \mathbb{P}\{\mathbb{E}(\rho_{max}(\tilde{M})) \geq (1 + \epsilon)\rho_{max}(M) \}  & \leq n[\frac{\exp(\epsilon)}{(1+\epsilon)^{1+\epsilon}}]^{\rho_{max}(M)/L} \\
    & \leq c_2 n e^{-\epsilon \rho_{max}(M)/L}
\end{split}
\end{equation}

where $\epsilon \in [0, 1)$. 

\vspace{12pt}

\textbf{Lemma 9. (Matrix Bernstein Inequality) \cite{tropp2015introduction}}. Consider a finite sequence $\{ R_k \}$, $\forall R_k \in \mathbb{C}^{n_1 \times n_2}$ of independent, random matrices. Assume that 
\begin{equation}
    \mathbb{E}[R_k] = 0 \ \text{and} \ \lVert R_k \rVert \leq c_4 \quad  \text{for each index $k$.}
\end{equation}
Following the above symbol, the matrix is defined as 
\begin{equation}
    M = \sum_{k} R_k
\end{equation}
Let $Var(M)$ be the variance of the random matrix $M$: 
\begin{equation}
    Var(M) = \max \{ \lVert \mathbb{E}[MM^*] \rVert, \lVert \mathbb{E}[M^*M] \rVert \} 
\end{equation}
and 
\begin{equation}
    \mathbb{P}(\lVert M \rVert \leq c_5) \leq (n_1 + n_2) \exp (\frac{-c_5^2/2}{Var(M)+c_4c_5/3})
\end{equation}

\vspace{12 pt}

\textbf{Proposition 1. (Concentration of minimum eigenvalue of Hermitian matrices)} Consider an arbitrary matrix of dimensionality $n$ such that $\Sigma$ and $\lVert \Sigma \rVert_{F} \leq c_2$. Let $\{R_k\}_{k \in K}$ be $\lvert K \rvert$ i.i.d samples of the distribution of $\Sigma$. The $\Sigma$ can be represented as:
\begin{equation}
    \Sigma = \mathbb{E}[\Phi \Phi^*] 
\end{equation}
and the empirical estimation is 
\begin{equation}
    \Tilde{\Sigma} = \frac{1}{\lvert K \rvert} \sum_{k \in K} \phi_{k} \phi_{k}^*
\end{equation}
\\
To guarantee the sufficiently large of probability $1 - \delta$ $\forall \delta \in (0,1)$ with the sample complexity is $\mathcal{O}(\frac{\log(c_2 n /\delta)}{\rho_{min}(\Sigma)})$, satisfying 
\begin{equation}
    \rho_{min}(\Sigma) - \rho_{min}(\tilde{\Sigma}) \lesssim - \frac{\log(c_2 n / \delta )}{\rho_{min}(\Sigma)}
\end{equation}

\textit{Proof.} Following the definition in Lemma 8, the tail probability is bounded as $\mathbb{P}\{\mathbb{E}(\rho_{min}(M)) \leq (1 - \epsilon)\rho_{min}(M) \} \leq c_2de^{-\epsilon \rho_{min}(M)/L}$. We can set
\begin{equation}
    \delta \leq c_2de^{-\epsilon \rho_{min}(\Sigma)/L}
\end{equation}
since $\lVert \Sigma \rVert_{F} \leq c_2$, $\Rightarrow \rho_{max} (\Sigma) \leq c_2 \leq c_3^2/ \lvert K \rvert$ then 
\begin{equation}
    \epsilon \geq c_2de^{-\epsilon \rho_{min}(\Sigma) \lvert K \rvert/c_3^2}
\end{equation}
\begin{equation}
    \Rightarrow \epsilon \leq \frac{c_3^2 \log (c_2 n/\delta)}{\lvert K \rvert \rho_{\min}(\Sigma)}
\end{equation}
Then the error bound is convergent with the sample complexity as $\lvert K \rvert \sim \mathcal{O}(\frac{\log(c_2 n /\delta)}{\rho_{min}(\Sigma)})$.

\vspace{12 pt}

\textbf{Proposition 2. (Error bound of the empirical asymmetric matrices)} Consider the asymmetric uncentred random matrix $\Sigma \in \mathbb{C}^{n_1 \times n_2}$ such that in Eq.\eqref{uncentred covariance}. The empirical estimation is 
\begin{equation}
    \Tilde{\Sigma} = \frac{1}{\lvert K \rvert} \sum_{k \in K} \phi_{k} \varphi_{k}^*
\end{equation} 
we have the probability with at least $1 - \delta $, $\forall \delta \in (0,1)$ satisfying the 
\begin{equation}
    \mathbb{P}(\lVert \Tilde{\Sigma} - \Sigma \rVert \leq \varepsilon)  \geq 1 - \delta
\end{equation}
where $\phi_k \in \mathbb{C}^{n_1}, \varphi_k \in \mathbb{C}^{n_2}$ for all $k \in K$ and 
\begin{equation}
\begin{split}
    & \varepsilon \defeq \frac{2 \log( (n_1 + n_2)/\delta) \Bar{c}_4}{3} + \frac{\sqrt{2 \log ( (n_1 + n_2)/\delta) \overline{Var(\Sigma)}}}{2} 
\end{split}
\end{equation}

\begin{align*}
    \sqrt{\overline{Var(\Sigma)}} \defeq \frac{\max_{i,j} n_1 n_2 \sqrt{20 \log(2 \Bar{c}_{i,j}/\delta)}}{\sqrt{\lvert K \rvert}}     
\end{align*}
\begin{align*}
    \Bar{c}_4 \defeq \frac{\max_{i,j} n_1 n_2 \sqrt{20 \log(2 \Bar{c}_{i,j}/\delta)}}{\sqrt{\lvert K \rvert}} + \frac{\sqrt{200 C n_1 n_2 \max_{i,j} \Bar{c}_{i,j}} (\log(2/\delta))^{\frac{3}{4}} }{\lvert K \rvert^{\frac{1}{4}}}      
\end{align*}

\vspace{12 pt}

\textit{Proof.}
For the uncentred random matrix $\Sigma$, when the empirical sample size $\lvert K \rvert$ is sufficiently large, we have the result such that
\begin{center}
    $ \limsup_{\lvert K \rvert \rightarrow \infty} \mathbb{E}[\Sigma - \tilde{\Sigma}] \rightarrow 0$
\end{center}
Following this property, we can observe that
\begin{equation}
\begin{split}
    & \ \quad \lVert \Sigma  - \tilde{\Sigma} \rVert_F \\
    & = \lVert \Sigma  - \frac{1}{\lvert K \rvert} \sum_{k \in K} \phi_{k}^* \varphi_{k} \rVert_F 
\end{split}
\end{equation}
Here we denote the $Err_{\Sigma} = \Sigma  - \frac{1}{\lvert K \rvert} \sum_{k \in K} \phi_{k}^* \varphi_{k}$, since $\forall k \in K$, the $\phi_k, \varphi \in \mathcal{H}$ are kernel functions. The matrix $\Sigma$ can be regarded as the mean embedding of a random matrix under one specific distribution. In such a situation, we have the result for the arbitrary element of $[Err_{\Sigma}]_{i \in [n_1], j \in [n_2]}$ is bounded since $2\Bar{c}_{i,j}/\lvert K \rvert$, see Eq. \eqref{dirac measure minus mean embedding}, we have the following property as
\begin{align*}
    \lvert \mathbb{E}[Err_{\Sigma}]_{i \in [n_1], j \in [n_2]} \rvert  \leq \frac{2\Bar{x}_{i,j}}{\sqrt{\lvert K \rvert}}
\end{align*}
and invoking by the Lemma 4, we have the probability with at least $1 - \delta$ 
\begin{align*}
    \lvert [Err_{\Sigma}]_{i \in [n_1], j \in [n_2]} \rvert \leq \frac{\sqrt{20 \log(2 \Bar{c}_{i,j}/\delta)}}{\sqrt{\lvert K \rvert}} 
\end{align*}
After estimating the bound for each element, the Frobenius norm of $Err_{\Sigma}$ we have 
\begin{align*}
    & \ \quad \lVert Err_{\Sigma} \rVert_F \\
    & \leq \sum_{i \in [n_1], j \in [n_2]} \lvert [Err_{\Sigma}]_{i \in [n_1], j \in [n_2]} \rvert \\
    & \leq \frac{\max_{i,j} n_1 n_2 \sqrt{20 \log(2 \Bar{c}_{i,j}/\delta)}}{\sqrt{\lvert K \rvert}}
\end{align*}

\vspace{12 pt}

To obtain the variance, we have
\begin{equation}
\begin{split}
    &  \ \quad Var(\Sigma  - \tilde{\Sigma} ) \\ 
    & \defeq \max\{ \lVert \mathbb{E}[(\Sigma  - \tilde{\Sigma} )^* (\Sigma  - \tilde{\Sigma} )]  \rVert, \lVert  \mathbb{E}[(\Sigma  - \tilde{\Sigma} ) (\Sigma  - \tilde{\Sigma} )^*] \rVert \} 
\end{split}
\end{equation}
By the Lemma 6, we have the following property 
\begin{align*}
    & \Rightarrow \max\{ \lVert \mathbb{E}[(\Sigma  - \tilde{\Sigma} )^* (\Sigma  - \tilde{\Sigma} )]  \rVert, \lVert  \mathbb{E}[(\Sigma  - \tilde{\Sigma} ) (\Sigma  - \tilde{\Sigma} )^*] \rVert \} \\
    & \leq  \max\{ \mathbb{E}[\lVert (\Sigma  - \tilde{\Sigma} )^* \rVert  \lVert (\Sigma  - \tilde{\Sigma} )\rVert]  ,  \mathbb{E}[\lVert (\Sigma  - \tilde{\Sigma} ) \rVert \lVert  (\Sigma  - \tilde{\Sigma} )^* \rVert] \} \\
    & \leq \lVert Err_{\Sigma} \rVert_F^2 \\
    & \defeq \overline{Var(\Sigma)}
\end{align*}
According to the Lemma 9, we have 
\begin{equation}
    \mathbb{P}(\lVert \Sigma - \tilde{\Sigma} \rVert \leq c_5 ) \leq (n_1 + n_2) \exp (\frac{-c_5^2/2}{\overline{Var(\Sigma)}+\Bar{c}_4 c_5/3})
\end{equation}
\begin{equation}
   \Rightarrow  (n_1 + n_2) \exp (\frac{-c_5^2/2}{\overline{Var(\Sigma)}+\Bar{c}_4 c_5/3}) \leq \delta 
\end{equation}

\begin{equation}
    \Rightarrow c_5^2 - \frac{2 \log( (n_1 + n_2)/\delta) \Bar{c}_4}{3} c_5 - 2 \log ( (n_1 + n_2)/\delta) \overline{Var(\Sigma)} \leq 0 
\end{equation}
Due to the quadratic root formula, we have 
\begin{equation}
    \Rightarrow c_5 \leq \frac{-[-\frac{2 \log( (n_1 + n_2)/\delta) \Bar{c}_4}{3}] + \sqrt{[-\frac{2 \log( (n_1 + n_2)/\delta) \Bar{c}_4}{3}]^2 + 2 \log ( (n_1 + n_2)/\delta) \overline{Var(\Sigma)} } }{2}
\end{equation}
\begin{equation}
     \Rightarrow c_5 \leq \frac{2 \log( (n_1 + n_2)/\delta) \Bar{c}_4}{3} + \frac{\sqrt{2 \log ( (n_1 + n_2)/\delta) \overline{Var(\Sigma)}}}{2} \label{Berstein}
\end{equation}
According to the Lemma 9, $\Bar{c}_4 \defeq \lVert Err_{\Sigma} \rVert$, by the Lemma 4 and 6, we have a probability of at least $1 - \delta$
\begin{equation}
\begin{split}
    & \ \quad \lVert Err_{\Sigma} \rVert \\
    & = \lVert \mathbb{E}[ Err_{\Sigma} ] \rVert + \lVert Err_{\Sigma} -  \mathbb{E}[ Err_{\Sigma}  ] \rVert \\
    & \leq \lVert  \mathbb{E}[ Err_{\Sigma} \rVert_F + \frac{\sqrt{10 C\lVert Err_{\Sigma} \rVert_F \log(2/\delta)}}{\sqrt{\lVert K \rVert}} \\
    & \leq \frac{\max_{i,j} n_1 n_2 \sqrt{20 \log(2 \Bar{c}_{i,j}/\delta)}}{\sqrt{\lvert K \rvert}} + \frac{\sqrt{200 C n_1 n_2 \max_{i,j} \Bar{c}_{i,j}} (\log(2/\delta))^{\frac{3}{4}} }{\lvert K \rvert^{\frac{1}{4}}}
\end{split}
\end{equation}

Plug the $\Bar{c}_4, \overline{Var(\Sigma)}$ into the Formula \eqref{Berstein} we have the sample size 
\begin{equation}
    \lvert K \rvert \geq  \tilde{\mathcal{O}}((n_1n_2)^2 \Bar{c}_{i,j} \log(2/\delta)^3)
\end{equation}
we can derive the polynomial sample complexity as $\tilde{\mathcal{O}}((n_1n_2)^2 \Bar{c}_{i,j} \log(2/\delta)^3)$.

\vspace{12pt}

\textbf{Corollary 1.} The error of bound of operator $\Sigma_{\mathcal{O \mid A}} \defeq \Sigma_{\mathcal{OA}} \Sigma_{\mathcal{AA}}^{-1}$ has the empirical estimation as:
\begin{equation}
    \Tilde{\Sigma}_{\mathcal{O \mid A}} = \Tilde{\Sigma}_{\mathcal{O A}} ( \tilde{\Sigma}_{\mathcal{AA}} + \lambda I)^{-1}
\end{equation}
where $\lambda \rightarrow 0$ and $ \phi_{k}^{\mathcal{O}} \in \mathbb{C}^{n_1} $ and $ \phi_{k}^{\mathcal{A}} \in \mathbb{C}^{n_2} $ for all $k \in K$, 
\begin{equation}
\begin{split}
    & \Tilde{\Sigma}_{\mathcal{O A}} \defeq \frac{1}{\lvert K \rvert} \sum_{k \in K} \phi_{k}^{\mathcal{O}}  \phi_{k}^{\mathcal{O} *} \\
    & \Tilde{\Sigma}_{\mathcal{A A}} \defeq \frac{1}{\lvert K \rvert} \sum_{k \in K} \phi_{k}^{\mathcal{A}}  \phi_{k}^{\mathcal{A} *}
\end{split}
\end{equation}
we have $\forall \delta \in (0,1)$ satisfying the property as
\begin{equation}
    \mathbb{P}(\lVert \Sigma_{\mathcal{O \mid A}} -\Tilde{\Sigma}_{\mathcal{O \mid A}} \rVert \leq c_6) \geq 1 - 3 \delta 
\end{equation}
where 
\begin{equation}
\begin{split}
    c_6 \defeq  & \frac{2 \log( (n_1 + n_2)/\delta) \Bar{c}_7}{3 \rho_{min}(\Sigma_{\mathcal{AA}})} + \frac{\sqrt{2 \log ( (n_1 + n_2)/\delta) \overline{Var(\Sigma_{\mathcal{OA}})}}}{2 \rho_{min}(\Sigma_{\mathcal{AA}})} \\ 
    & + 
      \frac{\sqrt{\rho_{max}(\Sigma_{\mathcal{OO}})} + \sqrt{\rho_{max}(\Sigma_{\mathcal{OO}}) }  \epsilon_1}{ \sqrt{ \rho_{min}(\Sigma_{\mathcal{AA}})}} \cdot \frac{\epsilon_2 + \lambda}{1 + \epsilon_2 + \lambda } 
\end{split}
\end{equation}
\begin{align*}
    & \Bar{c}_7 = \frac{\max_{i,j} n_1 n_2 \sqrt{20 \log(2 \Bar{c}_{i,j}/\delta)}}{\sqrt{\lvert K \rvert}} + \frac{\sqrt{200 C n_1 n_2 \max_{i,j} \Bar{c}_{i,j}} (\log(2/\delta))^{\frac{3}{4}} }{\lvert K \rvert^{\frac{1}{4}}}   \\
    & \sqrt{\overline{Var(\Sigma_{\mathcal{OA}})}} =  \frac{\max_{i,j} n_1 n_2 \sqrt{20 \log(2 \Bar{c}_{i,j}/\delta)}}{\sqrt{\lvert K \rvert}}  \\
    & \epsilon_1 \lesssim \frac{ \log (n_1/\delta)}{ \lvert K \rvert \rho_{\max}(\Sigma_{\mathcal{OO}})} \\
    & \epsilon_2 \lesssim \frac{ \log (n_1/\delta)}{ \lvert K \rvert  \rho_{\max}(\Sigma_{\mathcal{AA}})}
\end{align*}

\vspace{12pt}

\textit{Proof.} 
\begin{equation}
\begin{split}
    &\ \quad \Sigma_{\mathcal{O \mid A}} -  \Tilde{\Sigma}_{\mathcal{O \mid A}} \\
    & = \Sigma_{\mathcal{OA}} \Sigma_{\mathcal{AA}}^{-1} - \Tilde{\Sigma}_{\mathcal{O A}} ( \tilde{\Sigma}_{\mathcal{AA}} + \lambda I)^{-1} \\
    & = \underbrace{\Sigma_{\mathcal{OA}} \Sigma_{\mathcal{AA}}^{-1} - \Tilde{\Sigma}_{\mathcal{O A}} \Sigma_{\mathcal{AA}}^{-1}}_{\text{P1}} + \underbrace{\Tilde{\Sigma}_{\mathcal{O A}} \Sigma_{\mathcal{AA}}^{-1}   - \Tilde{\Sigma}_{\mathcal{O A}} ( \tilde{\Sigma}_{\mathcal{AA}} + \lambda I)^{-1}}_{\text{P2}}
\end{split}
\end{equation}
To analyze the error bound, we can decompose the error into two parts as
\begin{center}
    $\text{P1} = \Sigma_{\mathcal{OA}} \Sigma_{\mathcal{AA}}^{-1} - \Tilde{\Sigma}_{\mathcal{O A}} \Sigma_{\mathcal{AA}}^{-1} $ 
\end{center}
\begin{center}
    $\text{P2} = \Tilde{\Sigma}_{\mathcal{O A}} \Sigma_{\mathcal{AA}}^{-1}   - \Tilde{\Sigma}_{\mathcal{O A}} ( \tilde{\Sigma}_{\mathcal{AA}} + \lambda I)^{-1} $
\end{center}
For the part 1, we have 
\begin{equation}
\begin{split}
    &\ \quad  \Sigma_{\mathcal{OA}} \Sigma_{\mathcal{AA}}^{-1} - \Tilde{\Sigma}_{\mathcal{O A}} \Sigma_{\mathcal{AA}}^{-1} \\
    & = ( \Sigma_{\mathcal{OA}} -  \Tilde{\Sigma}_{\mathcal{O A}}) \Sigma_{\mathcal{AA}}^{-1}  \\
    & \leq \lVert \Sigma_{\mathcal{OA}} -  \Tilde{\Sigma}_{\mathcal{O A}} \rVert \lVert \Sigma_{\mathcal{AA}}^{-1} \rVert 
\end{split}
\end{equation}
By Proposition 2, we have 
\begin{equation}
    \mathbb{P}(\lVert \Sigma_{\mathcal{OA}} -  \Tilde{\Sigma}_{\mathcal{O A}} \rVert \geq\varepsilon_{\mathcal{OA}}) \leq 1 - \delta
\end{equation}

\begin{equation}
\begin{split}
    & \varepsilon_{\mathcal{OA}} \defeq \frac{2 \log( (n_1 + n_2)/\delta) \Bar{c}_7}{3} + \frac{\sqrt{2 \log ( (n_1 + n_2)/\delta) \overline{Var(\Sigma_{\mathcal{OA}})}}}{2} 
\end{split}
\end{equation}
\begin{center}
    $\Bar{c}_7 = \frac{\max_{i,j} n_1 n_2 \sqrt{20 \log(2 \Bar{c}_{i,j}/\delta)}}{\sqrt{\lvert K \rvert}} + \frac{\sqrt{200 C n_1 n_2 \max_{i,j} \Bar{c}_{i,j}} (\log(2/\delta))^{\frac{3}{4}} }{\lvert K \rvert^{\frac{1}{4}}}   $
\end{center}
\begin{center}
    $\overline{Var(\Sigma_{\mathcal{OA}})} = \frac{\max_{i,j} n_1 n_2 \sqrt{20 \log(2 \Bar{c}_{i,j}/\delta)}}{\sqrt{\lvert K \rvert}} $
\end{center}

\vspace{12 pt}

Since the operator $\Sigma_{\mathcal{AA}}$ is self-adjoint, exiting a unitary decomposition as $\Sigma_{\mathcal{AA}} = U_{\mathcal{AA}}^* \Lambda U_{\mathcal{AA}}$, $U_{\mathcal{AA}}$ is unitary matrix and $\Lambda$ is a diagonal matrix with $\gamma-$regularity assumption, then we have
\begin{equation}
\begin{split}
    & \Rightarrow \lVert \Sigma_{\mathcal{OA}} -  \Tilde{\Sigma}_{\mathcal{O A}} \rVert \lVert \Sigma_{\mathcal{AA}}^{-1} \rVert \\
    & \leq \lVert \Sigma_{\mathcal{OA}} -  \Tilde{\Sigma}_{\mathcal{O A}} \rVert \lVert U_{\mathcal{AA}} \Lambda^{-1} U_{\mathcal{AA}}^*   \rVert  \\ 
    & \leq \lVert \Sigma_{\mathcal{OA}} -  \Tilde{\Sigma}_{\mathcal{O A}} \rVert \lVert U_{\mathcal{AA}} \Lambda^{-1} U_{\mathcal{AA}}^*   \rVert \\
    & \leq  \lVert \Sigma_{\mathcal{OA}} -  \Tilde{\Sigma}_{\mathcal{O A}} \rVert \lVert U_{\mathcal{AA}} \rVert \lVert \Lambda \rVert^{-1} \lVert U_{\mathcal{AA}}^* \rVert \\
    & \leq \frac{2 \log( (n_1 + n_2)/\delta) \Bar{c}_7}{3 \rho_{min}(\Lambda)} + \frac{\sqrt{2 \log ( (n_1 + n_2)/\delta) \overline{Var(\Sigma_{\mathcal{OA}})}}}{2 \rho_{min}(\Lambda)} 
    \label{Part 1 error }
\end{split}    
\end{equation}

with at least $1 - \delta$ probability.

According to the matrix inversion lemma such 
\begin{equation}
    M^{-1} - N^{-1} = M^{-1}(N - M) N^{-1}
\end{equation}
\newline 
The part 2 error can be represented as
\begin{equation}
\begin{split}
    & \ \quad \Tilde{\Sigma}_{\mathcal{O A}} \Sigma_{\mathcal{AA}}^{-1}   - \Tilde{\Sigma}_{\mathcal{O A}} ( \tilde{\Sigma}_{\mathcal{AA}} + \lambda I)^{-1} \\
    & = \Tilde{\Sigma}_{\mathcal{O A}} \Sigma_{\mathcal{AA}}^{-1} [\Sigma_{\mathcal{AA}} -  (\tilde{\Sigma}_{\mathcal{AA}} + \lambda I) ]( \tilde{\Sigma}_{\mathcal{AA}} + \lambda I)^{-1} \\
    & = \Tilde{\Sigma}_{\mathcal{O A}} \Sigma_{\mathcal{AA}}^{-1} (Err_{\mathcal{AA}} + \lambda I )( \tilde{\Sigma}_{\mathcal{AA}} + \lambda I)^{-1} \\
    & = (\Sigma_{\mathcal{OA}} + Err_{\mathcal{OA}} ) \Sigma_{\mathcal{AA}}^{-1} (Err_{\mathcal{AA}} + \lambda I )( \Sigma_{\mathcal{AA}} + Err_{\mathcal{AA}} + \lambda I)^{-1}
    \label{Collary: Part 2 }
\end{split}
\end{equation}
where 
\begin{align*}
    & Err_{\mathcal{OA}} = \Sigma_{\mathcal{OA}} - \tilde{\Sigma}_{\mathcal{OA}} \\
    & Err_{\mathcal{AA}} = \Sigma_{\mathcal{AA}} - \tilde{\Sigma}_{\mathcal{AA}}
\end{align*}
In this situation, we can decompose the matrix $\Sigma_{\mathcal{OA}}$, $ Err_{\mathcal{OA}}$ and $Err_{\mathcal{AA}}$ as subspace projection as
\begin{align*}
    & \Sigma_{\mathcal{OA}} = \Sigma_{\mathcal{OO}}^{\frac{1}{2}} \Sigma_{\mathcal{AA}}^{\frac{1}{2}} \\
    & Err_{\mathcal{OA}} = \Sigma_{\mathcal{OO}}^{\frac{1}{2}} \Lambda_{1} \Sigma_{\mathcal{AA}}^{\frac{1}{2}} \\
    & Err_{\mathcal{AA}} = \Sigma_{\mathcal{AA}}^{\frac{1}{2}} \Lambda_{2} \Sigma_{\mathcal{AA}}^{\frac{1}{2}}
\end{align*}
Plug into the Eq. \eqref{Collary: Part 2 }, we have
\begin{align*}
    & \ \quad (\Sigma_{\mathcal{OA}} + Err_{\mathcal{OA}} ) \Sigma_{\mathcal{AA}}^{-1} [Err_{\mathcal{AA}} + \lambda I) ]( \Sigma_{\mathcal{AA}} + Err_{\mathcal{AA}} + \lambda I)^{-1} \\
    & = \Sigma_{\mathcal{OO}}^{\frac{1}{2}} (I  + \Lambda_{1})  \Sigma_{\mathcal{AA}}^{\frac{1}{2}} \Sigma_{\mathcal{AA}}^{-1} (Err_{\mathcal{AA}} + \lambda I) ( \Sigma_{\mathcal{AA}} + Err_{\mathcal{AA}} + \lambda I)^{-1} \\
    & = \Sigma_{\mathcal{OO}}^{\frac{1}{2}} (I  + \Lambda_{1})  \Sigma_{\mathcal{AA}}^{-\frac{1}{2}} (Err_{\mathcal{AA}} + \lambda I) ( \Sigma_{\mathcal{AA}} + Err_{\mathcal{AA}} + \lambda I)^{-1}
\end{align*}
In this situation, the norm of error bound is 
\begin{equation}
\begin{split}
    & \ \quad \lVert \Sigma_{\mathcal{OO}}^{\frac{1}{2}} (I  + \Lambda_{1})  \Sigma_{\mathcal{AA}}^{-\frac{1}{2}} (Err_{\mathcal{AA}} + \lambda I) ( \Sigma_{\mathcal{AA}} + Err_{\mathcal{AA}} + \lambda I)^{-1} \rVert \\
    & \leq \lVert \Sigma_{\mathcal{OO}}^{\frac{1}{2}} \rVert \lVert (I  + \Lambda_{1}) \rVert  \lVert \Sigma_{\mathcal{AA}}^{-\frac{1}{2}} \rVert \lVert( Err_{\mathcal{AA}} + \lambda I) ( \Sigma_{\mathcal{AA}} + Err_{\mathcal{AA}} + \lambda I)^{-1} \rVert \\ 
    & \leq \frac{\rho_{max}(\Sigma_{\mathcal{OO}}^{\frac{1}{2}}) + \rho_{max}(\Sigma_{\mathcal{OO}}^{\frac{1}{2}}) \rho_{max}(\Lambda_1) }{\rho_{min}(\Sigma_{\mathcal{AA}}^{\frac{1}{2}})}  \underbrace{\lVert( Err_{\mathcal{AA}} + \lambda I)  \rVert \lVert \Sigma_{\mathcal{AA}} + Err_{\mathcal{AA}} + \lambda I)^{-1} \rVert}_{\text{fractional convex problem}}
    \\
    & \leq \frac{\rho_{max}(\Sigma_{\mathcal{OO}}^{\frac{1}{2}}) + \rho_{max}(\Sigma_{\mathcal{OO}}^{\frac{1}{2}}) \rho_{max}(\Lambda_1) }{\rho_{min}(\Sigma_{\mathcal{AA}}^{\frac{1}{2}})}  \cdot \max_{\lVert Err_{\mathcal{AA}} \rVert} \{ \lVert( Err_{\mathcal{AA}} + \lambda I)  \rVert \lVert \Sigma_{\mathcal{AA}} + Err_{\mathcal{AA}} + \lambda I) \rVert^{-1} \} \\
    & = \frac{\rho_{max}(\Sigma_{\mathcal{OO}}^{\frac{1}{2}}) + \rho_{max}(\Sigma_{\mathcal{OO}}^{\frac{1}{2}}) \rho_{max}(\Lambda_1) }{\rho_{min}(\Sigma_{\mathcal{AA}}^{\frac{1}{2}})}  \cdot \underbrace{ \max_{\rho_{max}(\Lambda_2)} \frac{\lVert \Lambda_{2}\rVert + \lambda}{\lVert I + \Lambda_2 + \lambda \rVert} }_{\text{concave function}} \\
    & = \frac{\sqrt{\rho_{max}(\Sigma_{\mathcal{OO}})} + \sqrt{\rho_{max}(\Sigma_{\mathcal{OO}}) } \rho_{max}(\Lambda_1) }{ \sqrt{ \rho_{min}(\Sigma_{\mathcal{AA}})}} \cdot \frac{\rho_{max}(\Lambda_2) + \lambda}{1 + \rho_{max}(\Lambda_2) + \lambda }    
    \label{Part 2 error}
\end{split}
\end{equation}
By Lemma 1, we can know that 
\begin{equation}
\begin{split}
     \mathbb{P}\{\mathbb{E}(\rho_{max}(\tilde{\Sigma})) \geq (1 + \epsilon)\rho_{max}(M) \} 
    & \lesssim n e^{-\epsilon \rho_{max}(M)/L}
\end{split}
\end{equation}
Using the inequality, $\delta \leq n e^{-\epsilon \rho_{max}(M)/L}$ we have the result as $\epsilon \lesssim n e^{-\epsilon \rho_{max}(M)/L}$. Therefore, we can drive the bound of $\Lambda_1$ and $\Lambda_2$ such that
\begin{align*}
    & \rho_{max}(\Lambda_1) \leq \epsilon_1 \\
    & \epsilon_1 \lesssim \frac{ \log (n_1/\delta)}{ \lvert K \rvert \rho_{\max}(\Sigma_{\mathcal{OO}})}
\end{align*}
With at least $1- \delta$ probability, and similarly
\begin{align*}
    & \rho_{max}(\Lambda) \leq 1 + \epsilon_2 \\
    & \epsilon_2 \lesssim \frac{ \log (n_1/\delta)}{ \lvert K \rvert \rho_{\max}(\Sigma_{\mathcal{AA}})}
\end{align*}
Combine the Eq. \eqref{Part 1 error } and \eqref{Part 2 error}, we have the error bound 
\begin{equation}
\begin{split}
    & \lVert \Sigma_{\mathcal{O \mid A}} -  \Tilde{\Sigma}_{\mathcal{O \mid A}} \rVert \\
    & \leq \frac{2 \log( (n_1 + n_2)/\delta) \Bar{c}_7}{3 \rho_{min}(\Sigma_{\mathcal{AA}})} + \frac{\sqrt{2 \log ( (n_1 + n_2)/\delta) \overline{Var(\Sigma_{\mathcal{OA}})}}}{2 \rho_{min}(\Sigma_{\mathcal{AA}})} + \\
    & + \frac{\sqrt{\rho_{max}(\Sigma_{\mathcal{OO}})} + \sqrt{\rho_{max}(\Sigma_{\mathcal{OO}}) }  \epsilon_1}{ \sqrt{ \rho_{min}(\Sigma_{\mathcal{AA}})}} \cdot \frac{\epsilon_2 + \lambda}{1 + \epsilon_2 + \lambda }
\end{split}
\end{equation}
with at least $1 - 3 \delta $ probability due to the union bound.


\end{document}